\documentclass{article} % For LaTeX2e
\usepackage[preprint]{colm2026_conference}
\usepackage{wrapfig}
\usepackage{microtype}
\usepackage{hyperref}
\usepackage{url}
\usepackage{booktabs}
\usepackage{amsmath,amssymb}
\usepackage{hyperref}
\usepackage{url}
\usepackage{amsthm}
\theoremstyle{definition}

\usepackage{algorithmic}
\usepackage[ruled,vlined,linesnumbered]{algorithm2e}
\usepackage{adjustbox}
\usepackage[misc]{ifsym}
\usepackage{xcolor}
\usepackage{colortbl}
\usepackage{microtype}
\usepackage{mathtools}
\usepackage{graphicx}
\usepackage{xspace}
\newcommand{\trans}{^{\intercal}} % Standard transpose notation
\newcommand{\inner}[2]{\langle #1, #2 \rangle}
\newcommand{\sketch}{\Pi}
\newcommand{\proxySketch}{\psi}
\IfFileExists{enumitem.sty}{
  \usepackage{enumitem}
}{
  \makeatletter
  \let\origitemize\itemize
  \let\endorigitemize\enditemize
  \renewenvironment{itemize}[1][]{\origitemize}{\endorigitemize}
  \makeatother
}

\usepackage{lineno}

\definecolor{darkblue}{rgb}{0, 0, 0.5}
\hypersetup{colorlinks=true, citecolor=darkblue, linkcolor=darkblue, urlcolor=darkblue}

\title{OPUS: Towards Efficient and Principled Data Selection in Large Language Model Pre-training in \textit{Every} Iteration}

\author{{
  \bf Shaobo Wang$^{1,2\dagger}$\thanks{Equal contribution. $\dagger$Work done while Shaobo Wang (\href{mailto:shaobowang1009@sjtu.edu.cn}{shaobowang1009@sjtu.edu.cn}) was an intern at the Qwen Team, Alibaba. $^{\text{\Letter}}$ Corresponding authors: Xingzhang Ren (\href{mailto:xingzhang.rxz@alibaba-inc.com}{xingzhang.rxz@alibaba-inc.com}), Dayiheng Liu (\href{mailto:liudayiheng.ldyh@alibaba-inc.com}{liudayiheng.ldyh@alibaba-inc.com}), and Linfeng Zhang (\href{mailto:zhanglinfeng@sjtu.edu.cn}{zhanglinfeng@sjtu.edu.cn})}  \quad 
    Xuan Ouyang$^{1,3}$$^*$ \quad
    Tianyi Xu$^{1,3}$$^*$ \quad
    Yuzheng Hu$^{4}$ \quad
    Jialin Liu$^{1}$ \quad
    \vspace{3pt}
  } \\
  {
  \bf
    Guo Chen$^{1}$  \quad
    Tianyu Zhang$^{5}$ \quad 
    Junhao Zheng$^{2}$ \quad 
    Kexin Yang$^{2}$ \quad 
    Xingzhang Ren$^{2}$$^{\text{\Letter}}$ 
    \vspace{3pt}
  } \\
    {
  \bf
    Dayiheng Liu$^{2}$$^{\text{\Letter}}$ \quad 
    Linfeng Zhang$^{1}$$^{\text{\Letter}}$ \quad 
    \vspace{5pt}
  } \\
}

\begin{document}

\ifcolmsubmission
\linenumbers
\fi

\maketitle
\vspace{-30pt}
\begin{center}
\small
% -------- first row --------
$^{1}$ EPIC Lab, SJTU
\raisebox{-0.2em}{\includegraphics[height=1.6em]{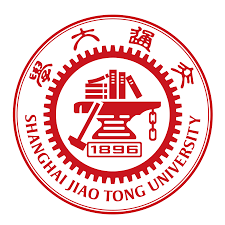}}
\hspace{1.8em}
$^{2}$ Qwen Team, Alibaba Group
\raisebox{-0.2em}{\includegraphics[height=1.6em]{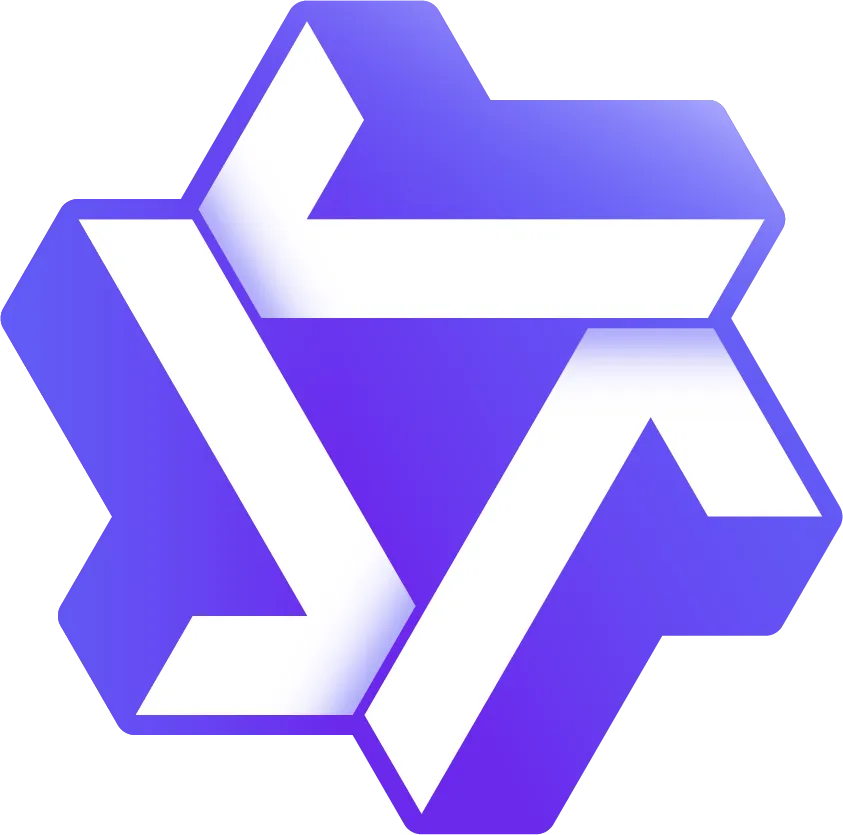}}
\hspace{1.8em}
$^{3}$ UW--Madison
\raisebox{-0.2em}{\includegraphics[height=1.6em]{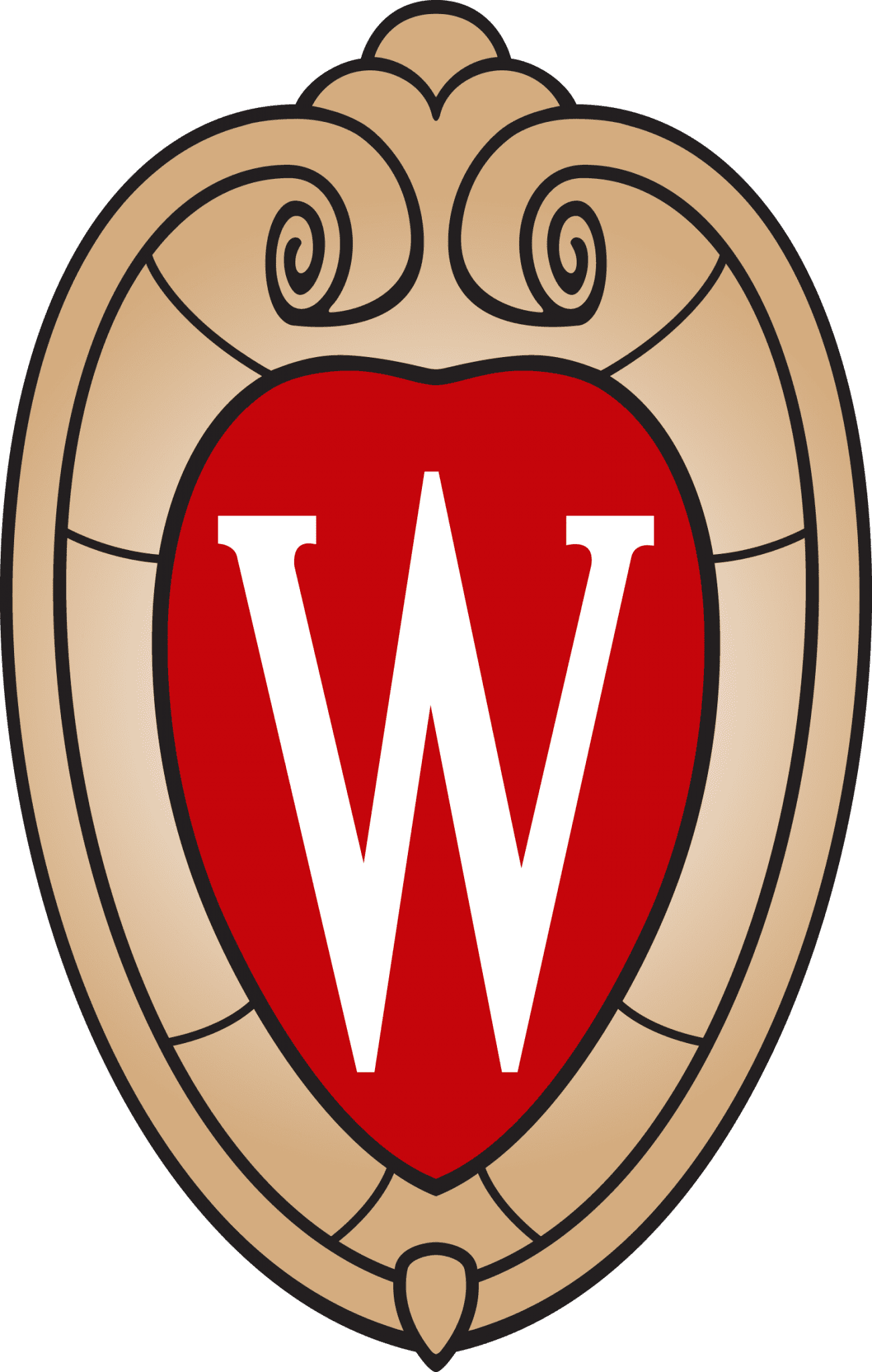}}
\\[2pt]
$^{4}$ UIUC
\raisebox{-0.3em}{\includegraphics[height=1.6em]{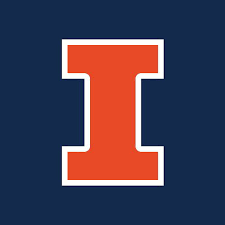}}
\hspace{3em}
$^{5}$ Mila - Quebec AI Institute
\raisebox{-0.3em}{\includegraphics[height=2em]{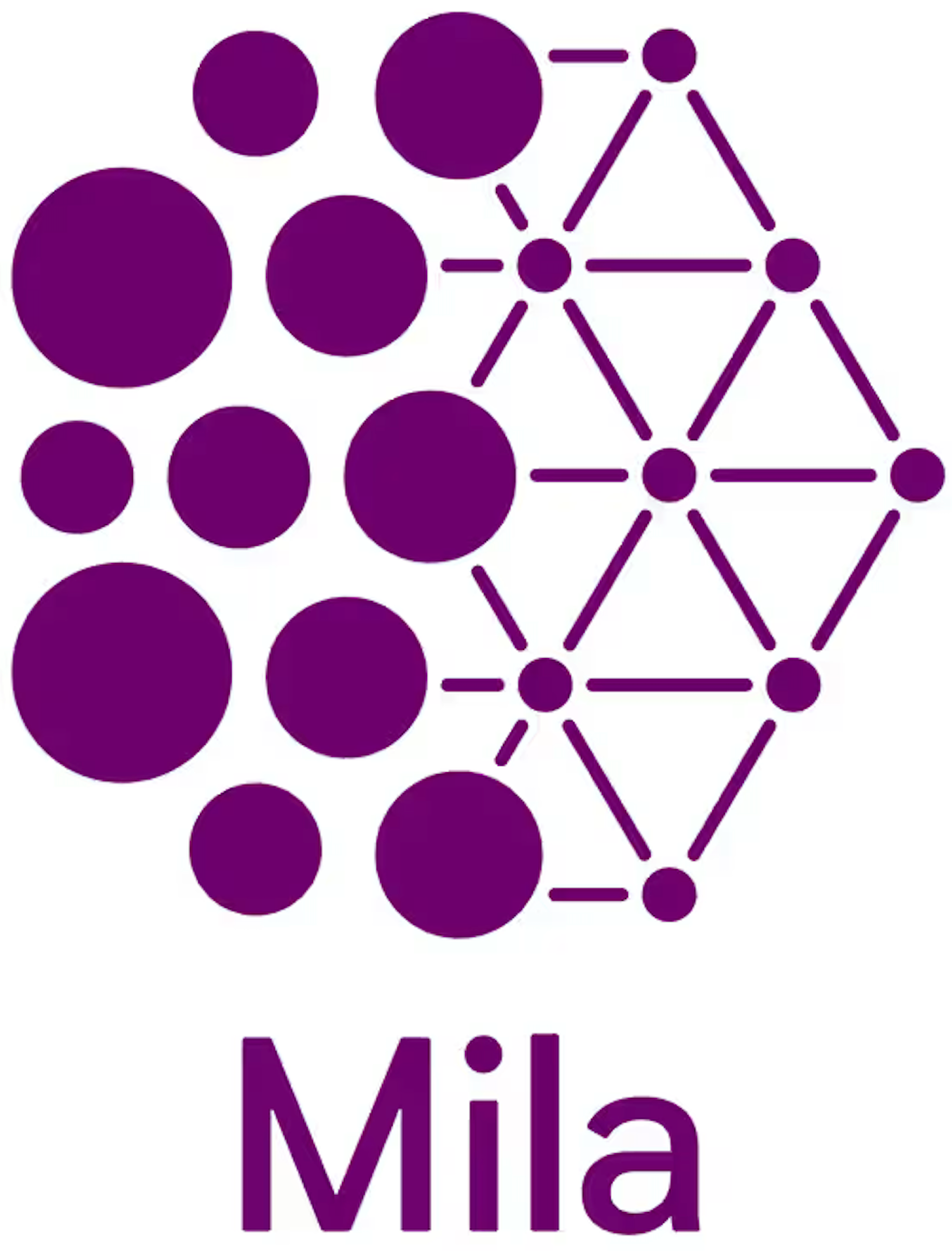}}
\end{center}

\vspace{15pt}

% ==================================================================
% ABSTRACT
% ==================================================================

\begin{figure}[htbp]
  \centering
  \vspace{-10pt}
    \includegraphics[width=0.99\linewidth]{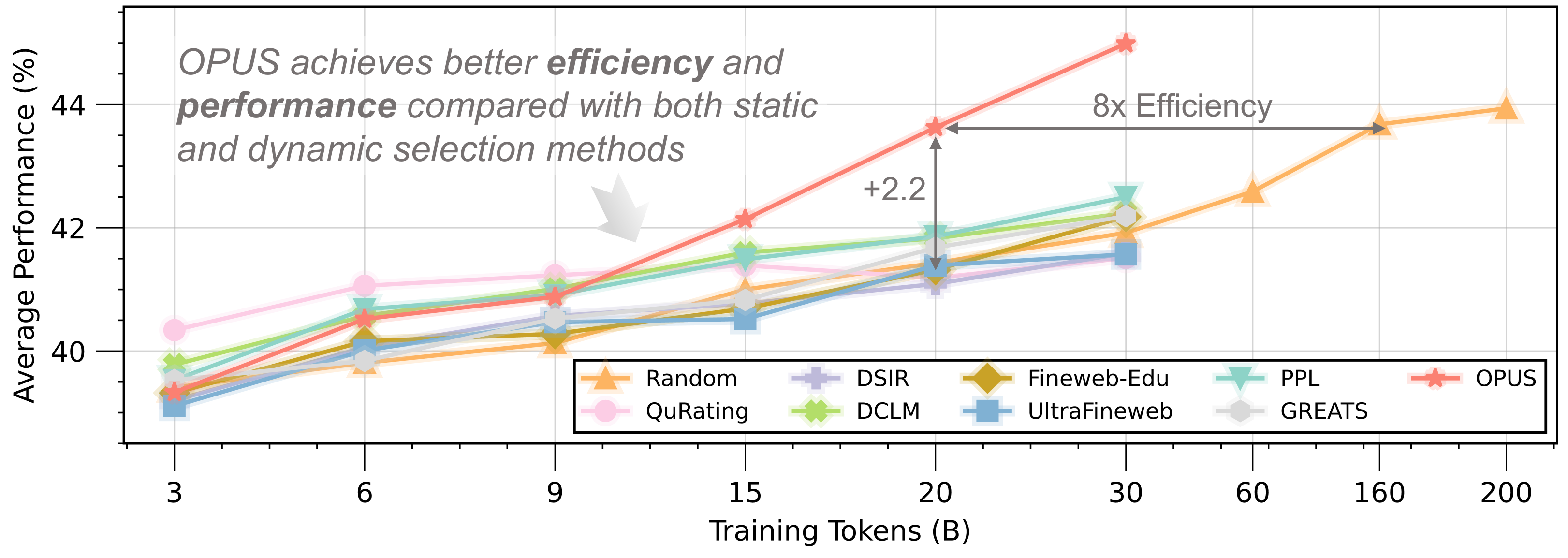}
  \vspace{-5pt}
  \caption{OPUS outperforms random selection by an average of 2.2\% accuracy across 10  benchmarks and achieves 8$\times$ reduction in computation on GPT-XL using FineWeb dataset.}
  % \vspace{-5pt}
  \label{fig:perf}
\end{figure}

\begin{abstract}
As high-quality public text approaches exhaustion, a phenomenon known as the \textit{Data Wall}~\citep{villalobos2022will}, pre-training is shifting from \textit{more tokens} to \textit{better tokens}. However, existing methods either rely on heuristic static filters that ignore training dynamics, or use dynamic yet optimizer-agnostic criteria based on raw gradients. We propose \textbf{OPUS} (\textbf{O}ptimizer-induced \textbf{P}rojected \textbf{U}tility \textbf{S}election), a dynamic data selection framework that defines utility in the optimizer-induced update space. OPUS scores candidates by projecting their effective updates, shaped by modern optimizers, onto a target direction derived from a stable, in-distribution proxy. To ensure scalability, we employ Ghost technique with CountSketch for computational efficiency, and Boltzmann sampling for data diversity, incurring only 4.7\% additional compute overhead. OPUS achieves remarkable results across diverse corpora, quality tiers, optimizers, and model scales. In pre-training of GPT-2 Large/XL on FineWeb and FineWeb-Edu with 30B tokens, OPUS outperforms industrial-level baselines and even full 200B-token training. Moreover, when combined with industrial-level static filters, OPUS further improves pre-training efficiency, even with lower-quality data. Furthermore, in continued pre-training of Qwen3-8B-Base on SciencePedia, OPUS achieves superior performance using only 0.5B tokens compared to full training with 3B tokens, demonstrating significant data efficiency gains in specialized domains.
\end{abstract}

% ==================================================================
% INTRODUCTION (REFINED FOR ICML SUBMISSION)
% ==================================================================
\section{Introduction}
\label{sec:intro}

Large language model (LLM) pre-training has entered a critical phase, transitioning from an era of unconstrained data scaling to a regime where the efficiency and quality of every training token are paramount. For the past decade, progress in language modeling has been driven by scaling two primary factors: model size and data volume~\citep{gpt2,gpt3,gpt4,qwen2,qwen25,qwen3,deepseekr1,deepseekv3,claude3}. Scaling laws emphasize that performance is tightly coupled with the efficiency of converting compute into effective training signals~\citep{chinchilla}. Yet the data factor is now saturating: projections suggest that readily available high-quality public text may be exhausted by 2026--2028~\citep{villalobos2022will}. In this data-wall regime, pre-training must shift from a problem of ingestion capacity to one of control: \textit{which tokens should shape the model at this specific optimizer step?} When every update consumes scarce tokens, data selection is no longer a pure preprocessing choice but an integral component of the optimization process.

% \begin{figure}[tb!]
%   \centering
%     \includegraphics[width=0.8\linewidth]{figs/opus_perf.pdf}
%   \caption{OPUS outperforms random selection by an average of 2.2\% accuracy across 10 downstream benchmarks and achieves 8$\times$ reduction in computation on GPT-XL using FineWeb dataset.}
%   \label{fig:perf}
% \end{figure}

\begin{figure}[tb!]
  \centering
    \includegraphics[width=0.8\linewidth]{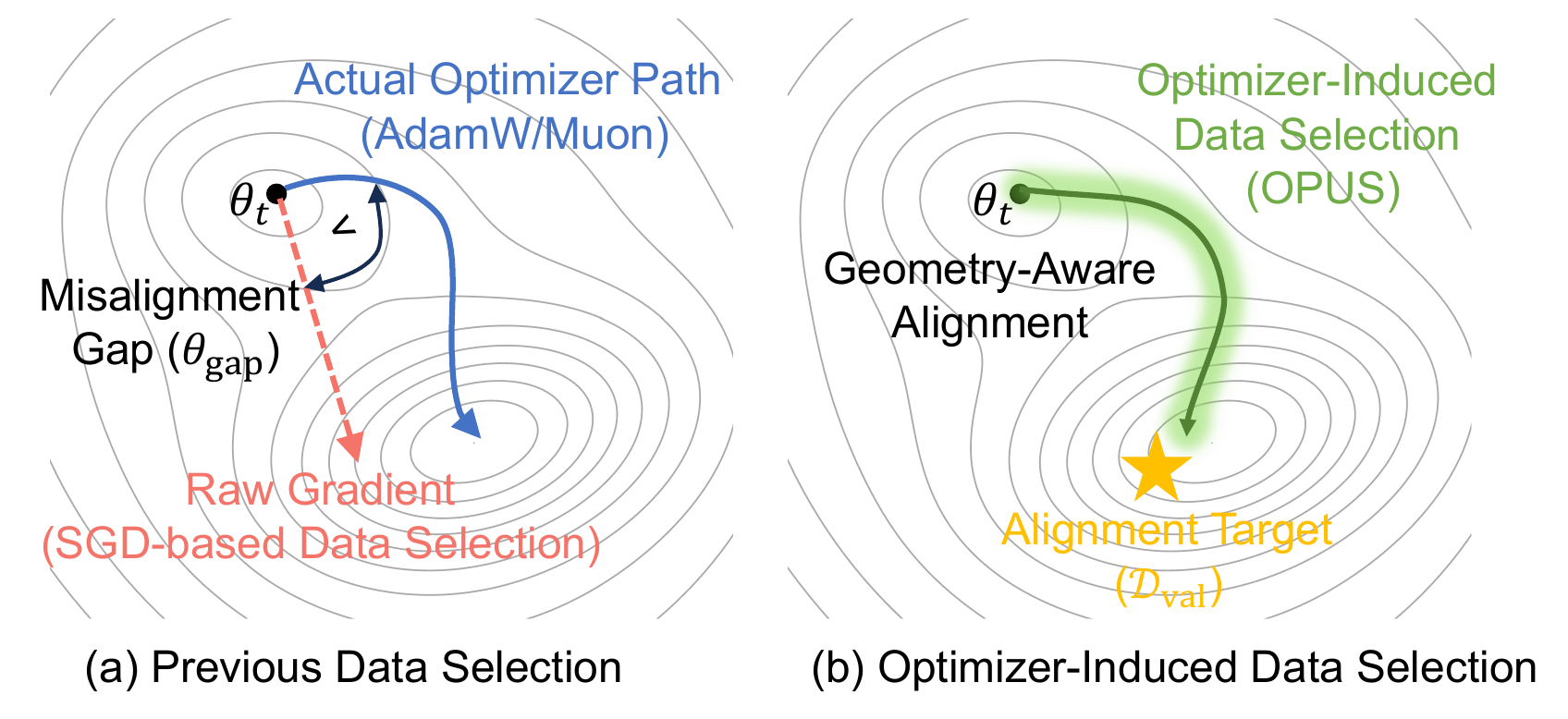}
  \caption{Comparison of different data selection methods. }
  \label{fig:comparison}
\end{figure}

Existing approaches to this problem present distinct limitations. Static curation methods, such as FineWeb-Edu classifiers~\citep{finewebedu} and the DCLM quality classifier~\citep{dclm}, rely on fixed, training-agnostic heuristics that assume a sample’s utility remains constant as the model evolves. In contrast, prior dynamic selection methods~\citep{Greats,wang2025data,wang2025capturing} score candidates in raw gradient space, implicitly assuming Stochastic Gradient Descent (SGD) dynamics. This induces a fundamental misalignment with modern LLM training, which relies on adaptive optimizers such as AdamW~\citep{adamw} and Muon~\citep{muon} that precondition and reshape the effective update direction. As shown in Figure~\ref{fig:comparison},  existing approaches depart from the optimizer's actual update geometry, causing unsatisfied optimization trajectory.

To bridge this gap, we introduce \textbf{OPUS} (\textbf{O}ptimizer-induced \textbf{P}rojected \textbf{U}tility \textbf{S}election), a framework designed to make data selection in pre-training both principled and scalable. OPUS achieves a principled objective by adapting during training to the model's evolving needs, unlike static filters, and by defining utility in the optimizer-induced update space. The core insight is that a batch is valuable only insofar as it moves parameters in a direction that improves the model's performance on a high-quality target distribution, referred to as the proxy, under the optimizer's specific geometry. OPUS scores each candidate by projecting its optimizer-induced effective update onto the descent direction of this proxy set, eliminating the discrepancy between scoring and training that arises when Adam or Muon training is treated as if it were SGD. To ensure scalability, OPUS estimates these utilities via lightweight projections, avoiding the prohibitive cost of materializing full gradients.

\begin{figure}[tb!]
  \centering
  \includegraphics[width=0.99\linewidth]{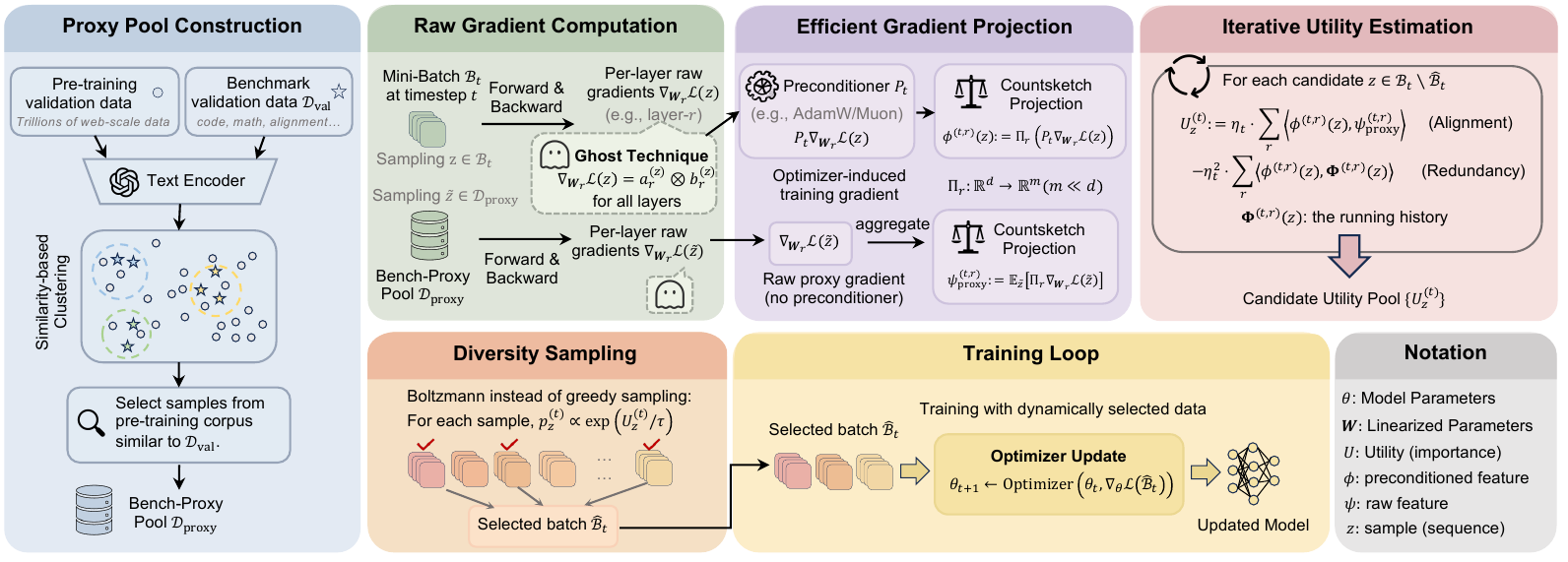}
  \caption{Overview of OPUS pipeline.}
  \label{fig:pipeline}
\end{figure}

OPUS operationalizes this principle through an objective, an estimator, and a selection rule. First, we formalize utility as the expected one-step improvement on a held-out proxy distribution, measured in the optimizer-induced update geometry, so that scoring aligns with the trajectory induced by AdamW or Muon. Second, we make this objective practical at LLM scale by (i) constructing a stable, in-distribution target direction for the proxy signal and (ii) estimating the required inner products efficiently without materializing per-sample gradients. Third, we use Boltzmann sampling to preserve data diversity. Figure~\ref{fig:pipeline} summarizes the end-to-end workflow. Our contributions are as follows:
\begin{itemize}[leftmargin=*,topsep=0pt,itemsep=0ex]
    \item \textbf{A principled, optimizer-aware utility for dynamic selection:} We introduce optimizer-induced utility as a theoretically grounded objective for dynamic data selection. By deriving closed-form approximations for the effective update directions of AdamW~\citep{adamw} and Muon~\citep{muon}, OPUS scores data in the actual optimizer-induced geometry, yielding a model- and optimizer-aware alternative to heuristic  filters.
    \item \textbf{Stable in-distribution proxy construction:} We propose \textsc{Bench-Proxy}, a procedure for constructing a proxy pool by retrieving benchmark-aligned samples directly from the pre-training corpus. This yields a reliable, in-distribution proxy direction that stabilizes utility estimation compared to using raw benchmark validation data.
    \item \textbf{Scalable utility estimation via ghost and CountSketch:} To make scoring efficient at LLM scale, we avoid per-sample gradient materialization by combining the ghost technique~\citep{Greats} with CountSketch projections~\citep{cormode2005improved}, reducing inner products to computations in a low-dimensional space.
    \item \textbf{Boltzmann sampling to prevent diversity collapse:} To avoid biased or redundant selection induced by greedy top-$k$ under non-stationary streams, OPUS uses Boltzmann soft sampling with an in-step redundancy penalty.
    \item \textbf{OPUS achieves strong empirical gains over industrial baselines:} Across from-scratch pre-training of GPT-2 Large/XL on FineWeb and FineWeb-Edu~\citep{finewebedu} and continued pre-training of Qwen3-8B-Base~\citep{qwen3} on SciencePedia~\citep{sciencepedia}, OPUS outperforms prior industrial static filters and dynamic selectors with better efficiency.
\end{itemize}

% ==================================================================
% PRELIMINARIES
% ==================================================================

\section{Related Work}
\noindent \textbf{Static pre-training data selection.}
Most large-scale LLM pre-training pipelines rely on static corpus filtering, where documents are filtered or reweighted once before training. Representative approaches include classifier- or rule-based filtering over web corpora, exemplified by FineWeb and its educational subset FineWeb-Edu~\citep{finewebedu}, which document large-scale deduplication and quality filtering choices for Common Crawl derived data. Recent work has also studied more targeted quality signals: QuRating~\citep{qurating} learns scalar quality ratings from pairwise preferences and shows that balancing quality and diversity improves downstream performance, while DSIR~\citep{dsir} formalizes dataset matching via importance resampling in a reduced feature space, enabling scalable selection without human curation. Complementary benchmark and pipeline efforts such as DataComp-LM (DCLM)~\citep{dclm} provide standardized corpora and evaluation suites to compare filtering strategies, and UltraFineweb~\citep{ultrafineweb} proposes efficient filtering and verification mechanisms (including lightweight classifier-based pipelines) to further improve web-scale data quality. While effective at removing low-quality noise, these static approaches are inherently training-agnostic: they assume sample utility is time-invariant and do not adapt to the model’s evolving needs across optimization.

\noindent \textbf{Dynamic data selection during pre-training.}
To move beyond fixed corpora, dynamic selection chooses samples on-the-fly based on an estimated training utility. Early and widely-used heuristics prioritize samples with large loss or high perplexity, and several works formalize this intuition via online batch selection and importance sampling \citep{loshchilov2016onlinebatchselectionfaster, katharopoulos2019samplescreatedequaldeep}. A more rigorous approach uses influence functions (IF) to estimate the impact of training points on validation loss~\citep{koh2017understanding}. While classic IF methods are computationally intensive and require Hessian inversion, recent approximations have made them more feasible for deep learning. In LLM pre-training, GREATS proposes a principled objective by approximating per-sample validation loss reduction via a Taylor expansion, and then selects a subset each step, typically greedily. It can incur substantial scoring overhead due to per-sample gradient and influence approximations \citep{Greats}. More recently, MATES~\citep{mates} learns a lightweight influence model to track evolving data preferences during pre-training, and Group-MATES~\citep{yu2025groupleveldataselectionefficient} emphasizes that utility is not additive and that group-level interactions matter, mitigating redundancy induced by greedy top-$k$ selection. In parallel, perplexity-based pruning remains a competitive, simple signal for data selection and pruning, including settings where a small reference model computes PPL to prune large-scale corpora \citep{ankner2025perplexed}. OPUS fits this dynamic-selection family, but differs by aligning utility with the optimizer-induced update and by using efficient projected scoring with soft sampling.

\noindent \textbf{Influence-function scores and data-attribution.}
A large line of work studies training-data influence and attribution~\citep{hammoudeh2024training,deng2025survey}---estimating how individual samples affect model behavior or validation loss. Classical influence functions approximate the effect of upweighting a training point via Hessian-based sensitivity analysis, enabling fine-grained data attribution without retraining \citep{koh2017understanding}. To make influence estimation practical in deep, non-convex settings, some works replace exact second-order IF computation with scalable surrogates \citep{pruthi2020estimating, guo2021fastif, yeh2018representer}. Related directions also develop first-order or early-training proxies for data importance, such as selecting informative subsets early in training \citep{paul2021deep}, leveraging forgetting events to identify noisy or hard-to-learn samples \citep{toneva2018an}, and optimizing subset selection via gradient-matching \citep{killamsetty2021grad} or influence functions~\citep{hu2024most}. Another line of research explores Shapley value, a concept from cooperative game theory, to quantify the value of data~\citep{ghorbani2019data,jia2021scalability,wang2025data}. Recently, influence and data-attribution signals have been adapted from classical IF literature to practical data selection for large language models, including LoRA-aware influence approximations and gradient-datastore based retrieval \citep{less}, as well as more structured selection pipelines that optimize selection objectives for instruction tuning \citep{du2023mods, liu2024what}. Moreover, many approaches implicitly operate in raw-gradient geometry and/or employ deterministic top-$k$ retrieval, which can become brittle under rapidly changing training dynamics and optimizer-induced transformations. These limitations motivate online selection objectives that remain faithful to the effective optimizer update while preserving scalability and diversity.

\section{Background}
\label{sec:preliminaries}

\subsection{LLM Pre-training}
We consider an autoregressive language model $f_\theta$ parameterized by $\theta \in \mathbb{R}^d$. A training sample is a token sequence $z=(x_1,\dots,x_L)$ with $x_i\in\mathcal{V}$, where $\mathcal{V}$ is the vocabulary and $L$ is the sequence length. The model defines the next-token distribution $p_\theta(x_i\mid x_{<i})$, and the per-sequence loss is the negative log-likelihood: $\mathcal{L}(z; \theta) = - \frac{1}{L} \sum_{i=1}^L \log p_\theta(x_i \mid x_{<i}).$ For any distribution (or finite set) $\mathcal{Q}$ over sequences, we define the expected loss
$\mathcal{L}(\mathcal{Q};\theta):=\mathbb{E}_{z\sim \mathcal{Q}}[\mathcal{L}(z;\theta)]$ (or its empirical average for a finite $\mathcal{Q}$).
Let $\mathcal{D}$ denote the full pre-training corpus. We partition it into (i) a training set $\mathcal{D}_{\text{tr}}$ used for parameter updates and (ii) a  held-out validation set $\mathcal{D}_{\text{val}}$ used only to guide selection. Importantly, $\mathcal{D}_{\text{val}} \cap \mathcal{D}_{\text{tr}}=\emptyset$, so validation samples never appear in training updates.

\subsection{Data Selection in Pre-training}
Data selection in pre-training aims to choose samples that compress knowledge both efficiently and effectively, which can be categorized into two domains.

\noindent \textbf{Static Data Selection.}
Static methods operate offline, filtering the entire candidate pool $\mathcal{D}_{\text{tr}}$ before training begins. A scoring function $S(z)$ assigns a quality score to each sample $z \in \mathcal{D}_{\text{tr}}$. A subset $\mathcal{D}_{\text{selected}} \subset \mathcal{D}_{\text{tr}}$ is retained by thresholding or top-$k$ selection: $\mathcal{D}_{\text{selected}} = \{z \in \mathcal{D}_{\text{tr}} \mid S(z) \ge \text{threshold} \}.$ The model is then trained on $\mathcal{D}_{\text{selected}}$ using a standard optimizer. While scalable, static selection ignores the model's evolving state $\theta_t$ during training.

\noindent \textbf{Dynamic Data Selection.}
Dynamic methods select data during training at each step $t$, adapting to the current model parameter $\theta_t$ and optimizer state. At step $t$, the algorithm receives a candidate buffer $\mathcal{B}_t=\{z_{1},\dots,z_{N}\}$ of $N$ sequences from the update stream $\mathcal{D}_{\text{tr}}$. It selects a subset $\widehat{\mathcal{B}}_t\subset\mathcal{B}_t$ of size $K=\lfloor\rho N\rfloor$ (selection ratio $\rho\in(0,1]$) to update the model, \emph{i.e.}, $\widehat{\mathcal{B}}_t = \textsc{Select}\big(\mathcal{B}_t; s_t(\cdot),K\big),$ where $s_t(z)$ is a step-dependent score (or sampling distribution) computed from the current model and proxy signal. 
\subsection{Modern Optimizers in Large-Scale Pre-training}
Many dynamic selection methods score candidates using the raw gradient $\nabla \mathcal{L}(z;\theta_t)$, implicitly assuming SGD-like geometry. Modern LLM training instead uses optimizers that transform gradients using state, such as momentum and adaptive preconditioning, changing the effective update direction.
We write the optimizer-induced effective update at step $t$ using an optimizer-induced preconditioner (operator) $\mathbf{P}_t$ applied to per-sample gradients:
\begin{equation}
    \Delta \theta_t(\widehat{\mathcal{B}}_t) = - \eta_t \sum_{z \in \widehat{\mathcal{B}}_t} \mathbf{P}_t\nabla \mathcal{L}(z; \theta_t).
\end{equation}
Here, $\mathbf{P}_t$ encapsulates the optimizer state at step $t$ and induces the geometry that the training trajectory actually follows. When the optimizer's transformation is not strictly linear, $\mathbf{P}_t$ should be read as a state-dependent operator acting on the gradient. This motivates defining selection scores in the optimizer-induced geometry rather than raw-gradient space. The details of common optimizers (SGD, AdamW, and Muon) are attached in Section~\ref{sec:optimizer}.

\section{Optimizer-induced Preconditioners}\label{sec:optimizer}
\subsection{Stochastic gradient descent}
We include SGD as a minimal reference point, since many prior dynamic selection methods implicitly assume an SGD-like update geometry and score candidates directly using raw gradients. In SGD, the optimizer applies a uniform scalar learning rate (and optional weight decay) without stateful preconditioning, so the effective update direction is aligned with the mini-batch gradient. Consequently, at a fixed step $t$, SGD induces an (approximately) identity update geometry, $\mathbf{P}_t \approx \mathbf{I}$, making raw-gradient similarity a natural scoring signal.

\begin{figure}[htbp]
    \centering
    \setlength{\fboxsep}{6pt}
    \fcolorbox{gray!30}{red!3}{
        \begin{minipage}{0.92\linewidth}
            \textbf{\textsc{SGD}}
            \vspace{1mm} 
            \hrule height 0.5pt 
            \vspace{1mm}
            Stochastic gradient descent updates parameters by moving along the negative mini-batch gradient:
            \[
            \mathbf g_t = \nabla_\theta \mathcal{L}(\mathcal{B}_t;\theta_t),
            \qquad
            \Delta\theta_t = -\eta_t\mathbf g_t.
            \]
            With optional weight decay, the one-step update becomes
            \[
            \Delta\theta_t
            =
            -\eta_t\big(\mathbf g_t + \lambda\theta_t\big).
            \]
            For online scoring at a fixed step $t$, SGD induces an identity update geometry
            $\mathbf P_t \approx \mathbf I$, so utility is naturally measured in raw-gradient space.
        \end{minipage}
    }
    % \caption{SGD induces an update geometry that is (approximately) raw-gradient space.}
    \label{fig:optbox_sgd}
\end{figure}

\subsection{Muon preconditioner}
We derive the Muon-instantiated preconditioner by linearizing Muon’s one-step \emph{lookahead} update
at a fixed training step $t$ (the regime used for online selection). Consider a linear weight matrix
$W_\mathcal{L}\in\mathbb{R}^{o\times i}$ updated by Muon. Ignoring bias-corrections for exposition, Muon maintains
an EMA momentum on the (mini-batch) gradient
$\mathbf{g}_{t,\mathcal{L}}(S):=\frac{1}{|S|}\sum_{z\in S}\nabla_{W_\mathcal{L}}\mathcal{L}(z;\theta_t)$:
\begin{equation}
\mathbf{m}_{t+1,\mathcal{L}}(S)=\mu\mathbf{m}_{t,\mathcal{L}}+(1-\mu)\mathbf{g}_{t,\mathcal{L}}(S).
\end{equation}
In practice, Muon forms a ``double-smoothed'' direction fed to the orthogonalizer,

\begin{equation}
\mathbf{q}_{t+1,\mathcal{L}}(S):=(1-\mu)\mathbf{g}_{t,\mathcal{L}}(S)+\mu\mathbf{m}_{t+1,\mathcal{L}}(S)
=\mu^2\mathbf{m}_{t,\mathcal{L}}(S)+(1-\mu^2)\mathbf{g}_{t,\mathcal{L}}(S).
\label{eq:muon_q_def}
\end{equation}

and takes the parameter step
\begin{equation}
\Delta W_{t,\mathcal{L}}(S)
:= W_{t+1,\mathcal{L}}(S)-W_{t,\mathcal{L}}
= -\eta_t\mathcal{O}_{t,\mathcal{L}}\!\big(\mathbf{q}_{t+1,\mathcal{L}}(S)\big).
\label{eq:muon_step}
\end{equation}

\textbf{Online-selection view.}
For scoring at fixed step $t$, we hold Muon’s state fixed (learning rate $\eta_t$, momentum coefficient $\mu$,
and the history buffer $\mathbf{m}_{t,\mathcal{L}}$). Moreover, we \emph{freeze} the Newton--Schulz (NS) operator
during selection by constructing it from a reference direction $\bar{\mathbf q}_{t,\mathcal{L}}$ available at the
start of step $t$ (e.g., from the current optimizer buffer / proxy batch), and reuse it for all candidates.
Under this approximation, NS induces an approximately linear left-multiplication map

\begin{equation}
\mathcal{O}_{t,\mathcal{L}}(Z)\approx \mathbf{S}_{t,\mathcal{L}}Z,
\;\mathbf{S}_{t,\mathcal{L}}=a\mathbf{I}+b\mathbf{A}_{t,\mathcal{L}}+c\mathbf{A}_{t,\mathcal{L}}^2,
\;\mathbf{A}_{t,\mathcal{L}}:=\tilde{\bar{\mathbf q}}_{t,\mathcal{L}}\tilde{\bar{\mathbf q}}_{t,\mathcal{L}}^{\top}.
\label{eq:muon_linear_map}
\end{equation}

where $\widetilde{\bar{\mathbf q}}_{t,\mathcal{L}}:=\bar{\mathbf q}_{t,\mathcal{L}}/\|\bar{\mathbf q}_{t,\mathcal{L}}\|_F$
(and $a,b,c$ are fixed NS polynomial coefficients). Substituting~\eqref{eq:muon_q_def} into~\eqref{eq:muon_step}
and using~\eqref{eq:muon_linear_map} yields the linearized lookahead update

\begin{equation}
\Delta W_{t,\mathcal{L}}(S)\approx \mathbf b_{t,\mathcal{L}}
-\kappa_t\mathbf S_{t,\mathcal{L}}\mathbf g_{t,\mathcal{L}}(S),
\;\mathbf b_{t,\mathcal{L}}:=-\eta_t\mu^2\mathbf S_{t,\mathcal{L}}\mathbf m_{t,\mathcal{L}},
\;\kappa_t:=\eta_t(1-\mu^2).
\label{eq:muon_linearized}
\end{equation}

Since OPUS ranks candidates/subsets by \emph{relative} utility at fixed $t$, the $S$-independent shift can be
dropped for scoring purposes, and the effective data-dependent update is captured by a layerwise preconditioner
\begin{equation}
\Delta W_{t,\mathcal{L}}(S) \approx -\mathbf{P}^{\mathrm{Muon}}_{t,\mathcal{L}}\mathbf g_{t,\mathcal{L}}(S)+\mathrm{const},
\qquad
\mathbf{P}^{\mathrm{Muon}}_{t,\mathcal{L}}:=\kappa_t\mathbf S_{t,\mathcal{L}}.
\label{eq:Pt_muon}
\end{equation}
Thus, Muon induces a \emph{dense, sample-independent} (at fixed $t$ under frozen $\mathbf S_{t,\mathcal{L}}$)
left-preconditioner that reshapes gradient directions before scoring; OPUS remains optimizer-agnostic by
plugging $\mathbf{P}^{\mathrm{Muon}}_{t,\mathcal{L}}$ into the same utility machinery used for AdamW.

\begin{figure}[htbp]
    \centering
    \setlength{\fboxsep}{6pt}
    \fcolorbox{gray!30}{green!3}{
        \begin{minipage}{0.92\linewidth}
            \textbf{\textsc{Muon}}            
            \vspace{1mm} 
            \hrule height 0.5pt 
            \vspace{1mm}
            Muon targets matrix-shaped parameters $W\in\mathbb{R}^{o\times i}$ by maintaining an accumulated
            matrix direction and applying a Newton--Schulz orthogonalization (matrix-sign style) transform:
            \begin{align*}
            \mathbf M_t &= \mu\mathbf M_{t-1} + (1-\mu)\mathbf g_t,\\
            \mathbf Q_t &:= \mathrm{NewtonSchulz}(\mathbf M_t),\\
            \Delta W_t &\propto -\mathbf Q_t.
            \end{align*}
            For online selection at fixed step $t$, we hold the optimizer state and freeze the Newton--Schulz
            operator across candidates, yielding an approximately linear map
            $\mathrm{NewtonSchulz}(Z)\approx \mathbf S_t Z$.
            This induces a dense, layerwise preconditioner $\mathbf P_t$ that reshapes update geometry
            beyond raw-gradient space.
        \end{minipage}
    }
    % \caption{Muon reshapes update geometry by orthogonalizing a momentum direction (for matrix-shaped parameters).}
    \label{fig:optbox_muon}
\end{figure}

\subsection{AdamW preconditioner}
We derive the AdamW-instantiated preconditioner by linearizing the one-step \emph{lookahead} update
that OPUS uses to score candidate subsets.
Consider the (decoupled) AdamW update applied to a subset $S$ at iteration $t$:

\begin{align}
\mathbf{m}_t(S) &= \beta_1 \mathbf{m}_{t-1} + (1-\beta_1)\mathbf{g}_t(S),
\qquad
\mathbf{v}_t(S) = \beta_2 \mathbf{v}_{t-1} + (1-\beta_2)\mathbf{g}_t(S)^{\odot 2},
\label{eq:adam_mv}\\[2pt]
\widehat{\mathbf{m}}_t(S) &= \frac{\mathbf{m}_t(S)}{1-\beta_1^t},
\qquad
\widehat{\mathbf{v}}_t(S) = \frac{\mathbf{v}_t(S)}{1-\beta_2^t},
\qquad
\theta_{t+1}(S) = \theta_t
- \alpha_t \frac{\widehat{\mathbf{m}}_t(S)}{\sqrt{\widehat{\mathbf{v}}_t(S)}+\epsilon}
- \alpha_t \lambda \theta_t.
\label{eq:adam_biascorr_step}
\end{align}

where $\mathbf{g}_t(S):=\frac{1}{|S|}\sum_{z\in S}\nabla_\theta \mathcal{L}(z;\theta_t)$ and $\odot$ denotes elementwise operations.

\textbf{Online-selection view.}
At a fixed training step $t$, OPUS compares subsets $S$ via their \emph{relative} utility under a one-step
lookahead while \emph{holding the optimizer state fixed at the start of step $t$}.
Concretely, we treat $\alpha_t,\beta_1,\beta_2,\epsilon,\lambda$ and the history buffers
$(\mathbf{m}_{t-1},\mathbf{v}_{t-1})$ as constants with respect to $S$.

\textbf{Affine dependence on the batch gradient.}
Under this view, the bias-corrected first moment is affine in $\mathbf{g}_t(S)$:
\begin{equation}
\widehat{\mathbf{m}}_t(S)
=
\frac{\beta_1}{1-\beta_1^t}\mathbf{m}_{t-1}
+
\frac{1-\beta_1}{1-\beta_1^t}\mathbf{g}_t(S).
\label{eq:mhat_affine}
\end{equation}

\textbf{Frozen preconditioner approximation.}
To keep scoring tractable, we freeze the RMS geometry during selection by
dropping the $S$-dependence in the second moment update. Using
$\widehat{\mathbf v}_t(S)=\mathbf v_t(S)/(1-\beta_2^t)$ with
$\mathbf v_t(S)=\beta_2\mathbf v_{t-1}+(1-\beta_2)\mathbf g_t(S)^{\odot 2}$, we approximate
\begin{equation}
\sqrt{\widehat{\mathbf v}_t(S)}+\epsilon
=
\sqrt{\frac{\beta_2\mathbf v_{t-1}+(1-\beta_2)\mathbf g_t(S)^{\odot 2}}{1-\beta_2^t}}+\epsilon
\approx
\sqrt{\overline{\mathbf v}_t}+\epsilon,
\qquad
\overline{\mathbf v}_t := \frac{\beta_2\mathbf v_{t-1}}{1-\beta_2^t}.
\label{eq:frozen_v}
\end{equation}

Substituting~\eqref{eq:mhat_affine} and~\eqref{eq:frozen_v} into~\eqref{eq:adam_biascorr_step} yields the linearized form.
Let $\mathbf{D}_t := \operatorname{Diag}\!\left(\frac{1}{\sqrt{\widehat{\mathbf{v}}_{t-1}}+\epsilon}\right)$,
$A_t := \alpha_t\frac{\beta_1}{1-\beta_1^t}$, and
$C_t := \alpha_t\frac{1-\beta_1}{1-\beta_1^t}$.
Then we have:
\begin{equation}
\Delta\theta_t(S) := \theta_{t+1}(S)-\theta_t
\approx
\underbrace{-A_t\mathbf{D}_t\mathbf{m}_{t-1}-\alpha_t\lambda\theta_t}_{\text{independent of }S}
- C_t\mathbf{D}_t\mathbf{g}_t(S).
\label{eq:adam_linearized}
\end{equation}

Since OPUS ranks subsets by \emph{relative} utility at fixed step $t$, the $S$-independent shift contributes
an additive constant to the (first-order) utility term and does not affect ranking.
Therefore, the effective \emph{data-dependent} update can be written as
{\setlength{\abovedisplayskip}{4pt}
 \setlength{\belowdisplayskip}{4pt}
 \setlength{\abovedisplayshortskip}{2pt}
 \setlength{\belowdisplayshortskip}{2pt}
\begin{equation}
\Delta\theta_t(S)\approx-\mathbf{P}_t^{\mathrm{AdamW}}\mathbf{g}_t(S)+\mathrm{const},\quad
\mathbf{P}_t^{\mathrm{AdamW}}:=C_t\operatorname{Diag}\!\Big(\tfrac{1}{\sqrt{\widehat{\mathbf{v}}_{t-1}}+\epsilon}\Big),\quad
C_t:=\alpha_t\tfrac{1-\beta_1}{1-\beta_1^t}.
\label{eq:Pt_adamw}
\end{equation}
}

\begin{figure}[htbp]
    \centering
    \setlength{\fboxsep}{5pt}
    \fcolorbox{gray!30}{cyan!3}{
        \begin{minipage}{0.92\linewidth}
            \textbf{\textsc{AdamW}}
            \vspace{1mm} 
            \hrule height 0.5pt 
            \vspace{1mm}
            AdamW maintains exponential moving averages of the gradient and its elementwise square:
            \begin{align*}
            \mathbf m_t &= \beta_1 \mathbf m_{t-1} + (1-\beta_1)\mathbf g_t, \qquad
            \widehat{\mathbf m}_t = \mathbf m_t/(1-\beta_1^t),\\
            \mathbf v_t &= \beta_2 \mathbf v_{t-1} + (1-\beta_2)\mathbf g_t^{\odot 2}, \qquad
            \widehat{\mathbf v}_t = \mathbf v_t/(1-\beta_2^t).
            \end{align*}
            With decoupled weight decay, the one-step update is
            \[
            \Delta\theta_t
            = -\alpha_t\frac{\widehat{\mathbf m}_t}{\sqrt{\widehat{\mathbf v}_t}+\epsilon}
            -\alpha_t\lambda\theta_t.
            \]
            For online scoring at a fixed step $t$, we freeze the RMS geometry and obtain an approximate
            diagonal preconditioner $\mathbf P_t \approx \alpha_t\mathrm{Diag}\!\big((\sqrt{\widehat{\mathbf v}_{t-1}}+\epsilon)^{-1}\big)$
            that rescales coordinates before measuring utility.
        \end{minipage}
    }
    % \caption{AdamW induces a state-dependent update geometry via adaptive diagonal preconditioning.}
    \label{fig:optbox_adamw}
\end{figure}

% ==================================================================
% METHODOLOGY
% ==================================================================
\section{Methodology: OPUS}
\label{sec:method}

\noindent We now describe OPUS and organize the section around the requirements that dynamic selection must satisfy in large-scale pre-training. Ideally, dynamic selection in large-scale pre-training should satisfy three desiderata:
\begin{itemize}[leftmargin=*,topsep=0pt,itemsep=0ex]
    \item \emph{Principled:} scores are derived from an explicit objective that measures improvement on a held-out proxy distribution under the optimizer-induced update geometry.
    \item \emph{Efficient:} scoring avoids materializing per-sample gradients in high-dimensional space.
    \item \emph{Scalable:} overhead remains modest as model dimension $m$ grows, enabling selection at every step.
\end{itemize}
\noindent Guided by these desiderata, we introduce \textbf{OPUS}, a dynamic data selection framework for LLM pre-training. At each step $t$, OPUS receives a candidate buffer $\mathcal{B}_t=\{z_1,\dots,z_N\}\subset \mathcal{D}_{\text{tr}}$ and selects $K=\lfloor \rho N\rfloor$ sequences to form the update batch. OPUS also draws a proxy mini-batch of size $K_{\text{proxy}}$ from a proxy pool $\mathcal{D}_{\text{proxy}}$, a finite surrogate for the held-out proxy set $\mathcal{D}_{\text{val}}$. Let $\mathbf{P}_t$ denote the optimizer-induced preconditioner at step $t$. We use sketch dimension $m$ for scoring in a projected space and temperature $\tau>0$ for stochastic sampling. \textbf{For details, please refer Algorithm~\ref{alg:opus} for the iterative OPUS algorithm.}

\label{app:opus_algorithm}
\begin{algorithm}[t]
  \caption{OPUS: Optimizer-induced Projected Utility Selection}
  \label{alg:opus}
\begin{algorithmic}[1]
  \STATE \textbf{Input:} Model $f_\theta$; Training Data stream $\mathcal{D}_{\text{tr}}$; Proxy pool $\mathcal{D}_{\text{proxy}}$; Optimizer $\mathcal{O}$; Selection ratio $\rho$; Projection dim $m$.
  \STATE \textbf{Initialize:} Implicit sketch operator $\Pi$ using CountSketch with hash $h:[d]\to[m]$ and sign $s:[d]\to\{-1,+1\}$.
  \FOR{$t=0, 1, \dots$}
    \STATE \textbf{1. Batch Sampling:} Read candidate buffer $\mathcal{B}_t=\{z_1,\dots,z_N\}$ from $\mathcal{D}_{\text{tr}}$.
    \STATE \textbf{2. Preconditioner Computation:} Construct optimizer-induced preconditioner $\mathbf{P}_t = \mathbf{P}(\mathcal{O}_t)$ from $\mathcal{O}$'s state at step $t$.
    \STATE \textbf{3. Proxy Feature Generation:} Sample $K_{\text{proxy}}$  samples $\{\tilde{z}_k\}$ from $\mathcal{D}_{\text{proxy}}$, obtain ghost factors $\{\mathbf{a}^{(\tilde{z}_k)}_r,\mathbf{b}^{(\tilde{z}_k)}_r\}$, and compute
    per-layer proxy sketches $\proxySketch_{\text{proxy}}^{(t,r)} \leftarrow 
\Pi_r\Big(\frac{1}{K_{\text{proxy}}}\sum_{k=1}^{K_{\text{proxy}}} \mathbf{a}^{(\tilde{z}_k)}_r\otimes\mathbf{b}^{(\tilde{z}_k)}_r\Big)$
for all $r\in\mathcal{R}$.
    \STATE \textbf{4. Candidate Feature Generation:}
    Compute per-layer sketches $\phi^{(t, r)}(z)\in\mathbb{R}^m$ implicitly from ghost factors $\{\mathbf{a}^{(z)}_r,\mathbf{b}^{(z)}_r\}_{r\in\mathcal{R}}$:
    \[
    \phi^{(t, r)}(z) \leftarrow \Pi_r\Big(\mathbf{P}_{t,r}\big(\mathbf{a}^{(z)}_r\otimes\mathbf{b}^{(z)}_r\big)\Big),\quad \forall r\in\mathcal{R}.
    \]
    \STATE \textbf{5. Soft Sampling Loop:}
    \STATE Let target batch size $K = \lfloor \rho N \rfloor$, Selected set $\widehat{\mathcal{B}}_t \leftarrow \emptyset$, and per-layer history $\Phi^{(t,r)} \leftarrow \mathbf{0}$ for all $r\in\mathcal{R}$.
    \FOR{$j=1$ to $K$}
        \STATE For each $z \in \mathcal{B}_t \setminus \widehat{\mathcal{B}}_t$, compute $U_{z}^{(t)}$:
        \[
        \begin{aligned}
        U_{z}^{(t)} \leftarrow \eta_t \sum_{r\in\mathcal{R}} \inner{\phi^{(t, r)}(z)}{\proxySketch_{\text{proxy}}^{(t, r)}} - \eta_t^2 \sum_{r\in\mathcal{R}} \inner{\phi^{(t, r)}(z)}{\Phi^{(t,r)}}
        \end{aligned}
        \]
        \STATE Sample index $z^*$ via Softmax: $p_t({z^*}) \propto \exp(U_{z}^{(t)} / \tau)$.
        \STATE Add to batch: $\widehat{\mathcal{B}}_t \leftarrow \widehat{\mathcal{B}}_t \cup \{z^*\}$.
        \STATE Update history (redundancy): $\Phi^{(t,r)} \leftarrow \Phi^{(t,r)} + \phi^{(t, r)}(z^*)$ for all $r\in\mathcal{R}$.
    \ENDFOR
    \STATE \textbf{6. Update:} Train $\theta_{t+1}$ using batch $\widehat{\mathcal{B}}_t$ with optimizer $\mathcal{O}$.
  \ENDFOR
\end{algorithmic}
\end{algorithm}

\subsection{Optimizer-Induced Utility Objective}
\label{sec:utility}
To obtain a principled scoring signal for selection, we define the utility of a candidate batch $\mathcal{S}$ as the reduction in loss on validation set $\mathcal{D}_{\text{val}}$ after one optimization step. Following~\citep{Greats}, we define utility at step $t$ as:
\begin{equation}
    U^{(t)}(\mathcal{S}) := \mathcal{L}(\mathcal{D}_{\text{val}}; \theta_t) - \mathcal{L}(\mathcal{D}_{\text{val}}; \theta_{t+1}(\mathcal{S})).
    \label{eq:utility_def}
\end{equation}

\textbf{Marginal gain.}
At each training step $t$, we are given a candidate buffer $\mathcal{B}_t$
and aim to construct an update subset $\widehat{\mathcal{B}}_t \subseteq \mathcal{B}_t$.
Let $z \in \mathcal{B}_t \setminus \widehat{\mathcal{B}}_t$ be a remaining candidate.
We define the marginal utility of adding $z$ as
\begin{equation}
U_z^{(t)}
:=
U^{(t)}(\widehat{\mathcal{B}}_t \cup \{z\})
-
U^{(t)}(\widehat{\mathcal{B}}_t).
\label{eq:marginal_gain_def}
\end{equation}

Let $\tilde{\theta}_t(\widehat{\mathcal{B}}_t)$ denote the \emph{virtual parameters} obtained by applying
one descent step on the selected subset $\widehat{\mathcal{B}}_t$: $\tilde{\theta}_t(\widehat{\mathcal{B}}_t)
=
\theta_t
+
\Delta\theta_t(\widehat{\mathcal{B}}_t).$ Adding $z$ induces an additional update $\Delta\theta_t(\{z\})$, so the marginal gain can be written as:
\begin{equation}
U_z^{(t)}
=
\mathcal{L}(\mathcal{D}_{\text{val}};\tilde{\theta}_t(\widehat{\mathcal{B}}_t))
-
\mathcal{L}(\mathcal{D}_{\text{val}};\tilde{\theta}_t(\widehat{\mathcal{B}}_t)+\Delta\theta_t(\{z\})).
\label{eq:marginal_virtual}
\end{equation}
Using a first-order Taylor approximation of the validation loss at
$\tilde{\theta}_t(\widehat{\mathcal{B}}_t)$, we have
\begin{equation}
\begin{split}
\mathcal{L}\!\left(\mathcal{D}_{\text{val}};
\tilde{\theta}_t(\widehat{\mathcal{B}}_t)+\Delta\theta_t(\{z\})\right)
\approx
\mathcal{L}\!\left(\mathcal{D}_{\text{val}};
\tilde{\theta}_t(\widehat{\mathcal{B}}_t)\right)
\\
+
\nabla_\theta \mathcal{L}\!\left(\mathcal{D}_{\text{val}};
\tilde{\theta}_t(\widehat{\mathcal{B}}_t)\right)\trans
\Delta\theta_t(\{z\}).
\end{split}
\label{eq:taylor_virtual_firstorder}
\end{equation}

Substituting Eq.~\eqref{eq:taylor_virtual_firstorder} into Eq.~\eqref{eq:marginal_virtual} yields
\begin{equation}
U_z^{(t)}
\approx
-
\nabla_{\theta}\mathcal{L}\!\left(\mathcal{D}_{\text{val}};\tilde{\theta}_t(\widehat{\mathcal{B}}_t)\right)\trans
\Delta\theta_t(\{z\}).
\label{eq:marginal_firstorder}
\end{equation}

\textbf{Optimizer-induced geometry.}
Unlike vanilla SGD, modern LLM training relies on adaptive optimizers that reshape gradients through a
state-dependent preconditioner. We denote the optimizer state operator at step $t$ as $\mathbf{P}_t$ and define
the \emph{optimizer-induced effective update direction} as:
\begin{equation}
{\mathbf u}_z^{(t)} \;:=\; \mathbf{P}_t \nabla_\theta \mathcal{L}(z;\theta_t).
\label{eq:effective_update_dir}
\end{equation}
Accordingly, the optimizer update induced by a subset $\mathcal{S}$ can be written as
$\Delta \theta_t(\mathcal{S}) = -\eta_t \sum_{z\in\mathcal{S}} {\mathbf u}_z^{(t)}$.
In particular, adding a single candidate $z$ contributes an additional update
$\Delta \theta_t(\{z\}) = -\eta_t\,{\mathbf u}_z^{(t)}$.
Substituting $\Delta \theta_t(\{z\})$  into the marginal approximation
in Eq.~\eqref{eq:marginal_firstorder} gives
\begin{equation}
U_z^{(t)}
\approx
\eta_t\,
\Big\langle
{\mathbf u}_z^{(t)},\,
\nabla_{\theta}\mathcal{L}\!\left(\mathcal{D}_{\text{val}};
\tilde{\theta}_t(\widehat{\mathcal{B}}_t)\right)
\Big\rangle.
\label{eq:marginal_precond_virtual_grad}
\end{equation}

\textbf{Approximating the virtual validation gradient.}
The marginal gain of adding a candidate $z$ to the current subset $\widehat{\mathcal{B}}_t$, denoted as $U_z^{(t)}$, depends on the validation gradient evaluated at the \emph{virtual parameters} $\tilde{\theta}_t(\widehat{\mathcal{B}}_t)$. Specifically, the first-order approximation of the utility is given by the inner product between the optimizer-induced update and the gradient at the virtual point:
\begin{equation}
U_z^{(t)} \approx \eta_t \Big\langle \mathbf{u}_z^{(t)}, \nabla_{\theta}\mathcal{L}(\mathcal{D}_{\text{val}}; \tilde{\theta}_t(\widehat{\mathcal{B}}_t)) \Big\rangle.
\label{eq:marginal_definition_virtual}
\end{equation}
Computing this virtual gradient exactly would require an additional backward pass on $\mathcal{D}_{\text{val}}$ after every selection step, which is prohibitively expensive. To avoid this cost, we linearize the gradient function $\mathbf{g}_{\text{val}}(\theta):=\nabla_{\theta}\mathcal{L}(\mathcal{D}_{\text{val}};\theta)$ around the current parameters $\theta_t$. Let $\Delta\theta_t(\widehat{\mathcal{B}}_t) := \tilde{\theta}_t(\widehat{\mathcal{B}}_t)-\theta_t$ be the accumulated update from the currently selected subset. A first-order Taylor expansion gives:
\begin{equation}
\nabla_{\theta}\mathcal{L}\!\left(\mathcal{D}_{\text{val}};
\tilde{\theta}_t(\widehat{\mathcal{B}}_t)\right)
\approx
\mathbf{g}_{\text{val}}(\theta_t)
+
\nabla_{\theta} \mathbf{g}_{\text{val}}(\theta_t) \, \Delta\theta_t(\widehat{\mathcal{B}}_t) \nonumber =
\mathbf{g}_{\text{val}}^{(t)}
+
\mathbf{H}_{\text{val}}^{(t)}\,\Delta\theta_t(\widehat{\mathcal{B}}_t),
\label{eq:virtual_val_grad_taylor}
\end{equation}
where $\mathbf{g}_{\text{val}}^{(t)}$ is the validation gradient at $\theta_t$ and $\mathbf{H}_{\text{val}}^{(t)}$ is the Hessian.
Using the update rule, the accumulated update is $\Delta\theta_t(\widehat{\mathcal{B}}_t) = -\eta_t\sum_{z_j\in\widehat{\mathcal{B}}_t}{\mathbf u}_{z_j}^{(t)}$. Substituting the gradient approximation (Eq.~\eqref{eq:virtual_val_grad_taylor}) and the explicit update form into Eq.~\eqref{eq:marginal_definition_virtual}, we obtain the final tractable scoring function:
\begin{equation}
U_z^{(t)}\approx
\eta_t \left\langle {\mathbf u}_z^{(t)}, \; \mathbf{g}_{\text{val}}^{(t)} - \eta_t \mathbf{H}_{\text{val}}^{(t)} \sum_{z_j\in\widehat{\mathcal{B}}_t}{\mathbf u}_{z_j}^{(t)} \right\rangle \nonumber =
\underbrace{\eta_t\Big\langle {\mathbf u}_z^{(t)}, \mathbf{g}_{\text{val}}^{(t)} \Big\rangle}_{\text{Alignment}}
-
\underbrace{\eta_t^2
\Big\langle
{\mathbf u}_z^{(t)},\,
\mathbf{H}_{\text{val}}^{(t)}
\sum_{z_j\in\widehat{\mathcal{B}}_t}{\mathbf u}_{z_j}^{(t)}
\Big\rangle}_{\text{Redundancy Penalty}}.
\label{eq:marginal_final_hessian}
\end{equation}

\textbf{Handling the Hessian complexity.}
Materializing $\mathbf H_{\text{val}}$ is intractable at LLM scale. Following~\citep{Greats}, we adopt an isotropic
approximation for this interaction term, $\mathbf H_{\text{val}} \approx \mathbf I$.
Defining the accumulated effective direction
${\mathbf G}^{(t)} := \sum_{z_j\in\widehat{\mathcal{B}}_t}{\mathbf u}_{z_j}^{(t)}$,
we obtain the practical redundancy-adjusted score:
\begin{equation}
U_z^{(t)}
\approx
\eta_t\Big\langle {\mathbf u}_z^{(t)}, \mathbf g_{\text{val}}^{(t)} \Big\rangle
-
\eta_t^2\Big\langle {\mathbf u}_z^{(t)}, {\mathbf G}^{(t)} \Big\rangle.
\label{eq:score_with_proxy_precond}
\end{equation}

\noindent \textbf{Stable proxy construction via \textsc{Bench-Proxy}.}
The quality of the proxy direction $\mathbf g_{\text{val}}^{(t)}$ is critical for principled selection. While a random hold-out set provides a low-variance signal, it often fails to capture the specific distribution of downstream tasks. Conversely, using raw benchmark samples directly as the proxy introduces severe distribution shift and gradient noise, destabilizing the ranking.
To bridge this gap, we introduce \textbf{\textsc{Bench-Proxy}}, a retrieval-based construction shown in Fig.~\ref{fig:pipeline}(a).
We embed both (i) the target benchmark validation set and (ii) candidate documents from the pre-training corpus using a frozen text encoder, and retrieve the top-$M$ most similar pre-training documents to form an \emph{in-distribution} proxy pool $\mathcal{D}_{\text{proxy}}$.
This approach yields a proxy that is aligned with the target tasks yet remains within the pre-training manifold, ensuring valid gradient estimation.
Concretely, at step $t$ we draw a proxy mini-batch $\{\tilde{z}_{k}\}_{k=1}^{K_{\text{proxy}}}\subset\mathcal{D}_{\text{proxy}}$ and estimate the direction via $\mathbf g_{\text{proxy}}^{(t)} = \frac{1}{K_{}}\sum_{k=1}^{K_{\text{proxy}}}\nabla_\theta \mathcal{L}(\tilde{z}_{k};\theta_t)$.
Substituting this proxy estimate into Eq.~\eqref{eq:score_with_proxy_precond}, we obtain the final scoring rule:
\begin{equation}
U_z^{(t)} \leftarrow
\eta_t \Big\langle {\mathbf u}_z^{(t)},\, \mathbf g_{\text{proxy}}^{(t)} \Big\rangle
-
\eta_t^2 \Big\langle {\mathbf u}_z^{(t)},\, {\mathbf G}^{(t)} \Big\rangle.
\label{eq:score_with_proxy_gradient}
\end{equation}
This formulation ensures that selected updates not only reduce loss but specifically align with the benchmark-relevant subspace of the optimization landscape. Further details of \textsc{Bench-Proxy} construction are provided in Sec~\ref{sec:bench_proxy}.

\subsection{Scalable Utility Estimation}
\label{sec:scalable_estimation}
To score candidates at scale, we leverage the ghost technique~\citep{Greats,wang2025data,hu2025a} to avoid per-sample forward/backward passes and the materialization of full gradients. We further apply a low-dimensional sketch to efficiently compute the inner products required for the utility score in Eq.~\eqref{eq:score_with_proxy_gradient}.

\noindent \textbf{Ghost technique.}
\label{sec:ghost}
Following GREATS~\citep{Greats}, we exploit the \emph{rank-1 outer product structure} of backpropagated gradients in linear layers. 
Consider a linear layer $r$ with weights $\mathbf{W}_r$. 
For a sample $z$, let $\mathbf{a}^{(z)}_r$ denote the input activation vector and $\mathbf{b}^{(z)}_r$ the output gradient vector (error signal). 
The per-sample gradient with respect to the weights factorizes as the outer product $\nabla_{\mathbf{W}_r}\mathcal{L}(z; \theta_t) = \mathbf{a}^{(z)}_r \otimes \mathbf{b}^{(z)}_r$, where $\otimes$ denotes the outer product.
Since $\mathbf{a}^{(z)}_r$ and $\mathbf{b}^{(z)}_r$ are available during the standard forward/backward passes, \textit{we can compute gradient statistics without ever materializing the high-dimensional matrix} $\nabla_{\mathbf{W}_r}\mathcal{L}$. In OPUS, we apply it over a set of layers $\mathcal{R}$ (e.g., linear and embedding matrices). 
We concatenate the proxy batch and candidate batch within a single forward/backward pass to collect $\{\mathbf{a}^{(z)}_r, \mathbf{b}^{(z)}_r\}$ for all samples. These quantities contain all information required to compute the projected scores, and are discarded layer-by-layer to maintain low memory overhead.

\noindent \textbf{CountSketch projection.}
\label{sec:random_projection}
Computing the utility $U_z^{(t)}$ in Eq.~\eqref{eq:score_with_proxy_gradient} requires applying the optimizer preconditioner $\mathbf{P}_t$. We project the resulting effective updates into a low-dimensional sketch space using a sparse CountSketch map $\sketch:\mathbb{R}^d\to\mathbb{R}^m$ ($m \ll d$).
For a linear layer $r$ with dimensions $d_{\text{in}} \times d_{\text{out}}$, the per-sample preconditioned sketch feature $\boldsymbol{\phi}^{(t,r)}(z) \in \mathbb{R}^m$ is computed implicitly as:
\begin{equation}
\boldsymbol{\phi}^{(t,r)}(z) = \sketch_r\Big(\mathbf{P}_{t,r}\big(\mathbf{a}^{(z)}_r \otimes \mathbf{b}^{(z)}_r\big)\Big).
\end{equation}
We instantiate $\sketch_r$ using CountSketch~\citep{cormode2005improved}, which enables computing the projection by streaming over the coordinates of the outer-product gradient without explicitly materializing it.
This choice yields concrete computational benefits depending on the structure of $\mathbf{P}_{t,r}$.
For AdamW, $\mathbf{P}_{t,r}$ is diagonal (Section~\ref{sec:optimizer}), preserving the coordinate-wise separable structure of the outer-product gradient.
This allows the CountSketch projection to be interleaved with preconditioning by applying the diagonal weights on the fly, yielding a projection cost of $\mathcal{O}(d_{\text{in}} + d_{\text{out}})$ rather than the $\mathcal{O}(d_{\text{in}} d_{\text{out}})$ cost required for a dense projection.
% This preserves the coordinate-wise sparsity structure of the outer product, allowing the projection to be computed in $\mathcal{O}(d_{\text{in}} + d_{\text{out}})$—proportional to the number of non-zeros—rather than the $\mathcal{O}(d_{\text{in}} d_{\text{out}})$ required for a dense projection.
In contrast, for optimizers with dense preconditioners such as Muon, coordinate mixing destroys this separability, resulting in a projection cost of $\mathcal{O}(d_{\text{in}} d_{\text{out}})$. We approximate the alignment and redundancy terms by summing dot products in the sketch space across layers:
\begin{equation}
U_z^{(t)}
\approx
\eta_t \sum_{r\in\mathcal{R}} \inner{\boldsymbol{\phi}^{(t,r)}(z)}{ \proxySketch_{\text{proxy}}^{(t,r)} }
- \eta_t^2 \sum_{r\in\mathcal{R}} \inner{\boldsymbol{\phi}^{(t,r)}(z)}{ \boldsymbol{\Phi}^{(t,r)} },
\label{eq:opus_score_sketch}
\end{equation}
where $\boldsymbol{\Phi}^{(t,r)} = \sum_{z_j\in\widehat{\mathcal{B}}_t} \boldsymbol{\phi}^{(t,r)}(z_j)$ is the running history of selected sketches. 
Note that $\proxySketch_{\text{proxy}}^{(t,r)} := \sketch_r \Big( \frac{1}{K_{\text{proxy}}}\sum_{k=1}^{K_{\text{proxy}}} \mathbf{a}^{(\tilde{z}_k)}_r \otimes \mathbf{b}^{(\tilde{z}_k)}_r \Big)$ represents the sketched \emph{unpreconditioned} proxy gradient direction.

\subsection{Boltzmann Sampling}
\label{sec:boltzmann}
To preserve diversity under dynamic selection, we replace deterministic greedy top-$k$ with stochastic sampling.
While our utility formulation in Eq.~\eqref{eq:opus_score_sketch} explicitly penalizes \emph{geometric} redundancy (vector alignment), greedy selection remains brittle to \emph{estimation noise}: it assumes the proxy direction $\proxySketch_{\text{proxy}}^{(t,r)}$ is perfect. In practice, the proxy is a stochastic estimate from a small batch, and the data stream is non-stationary. Always picking the current top-$k$ can lock the model into transient, noisy features of the proxy batch.
We therefore adopt Boltzmann sampling to improve robustness:
\begin{equation}
    p_z^{(t)} \propto \exp\big(U_{z}^{(t)} / \tau\big).
\end{equation}
This ensures that high-utility candidates are favored, while complementary candidates maintain non-zero probability, preventing overfitting to local proxy noise.

Algorithm~\ref{alg:opus} summarizes OPUS, a step-wise dynamic selection method that scores candidates in the \emph{optimizer-induced update space}. At each step $t$, OPUS samples a candidate buffer $\mathcal{B}_t$, constructs the preconditioner $\mathbf{P}_t$ from the optimizer state, and builds a proxy \emph{target direction} from an in-distribution pool $\mathcal{D}_{\text{proxy}}$ via ghost factors, yielding per-layer proxy sketches $\proxySketch_{\text{proxy}}^{(t,r)}$ for $r\in\mathcal{R}$. For each candidate $z\in\mathcal{B}_t$, it forms a sketch feature $\phi^{(t,r)}(z)$ by applying $\mathbf{P}_{t,r}$ to the ghost outer-product gradient and projecting with CountSketch $\Pi_r$ into $\mathbb{R}^m$ for efficiency. OPUS then selects $K=\lfloor \rho N\rfloor$ samples using Boltzmann sampling with a marginal-gain objective that balances proxy alignment and redundancy control, and updates the model on the selected subset $\widehat{\mathcal{B}}_t$.

\section{Experiments}

\subsection{Experimental Setup}
\label{sec:exp_setup}
\noindent \textbf{Models and training settings.}
We pre-train GPT-2 Large and GPT-2 XL~\citep{gpt2} from scratch under a fixed optimization budget of 30B update tokens. GPT-2 Large consists of 36 layers with a hidden size of 1280, totaling approximately 774M parameters, while GPT-2 XL features a deeper architecture of 48 layers and a hidden size of 1600, amounting to 1.5B parameters. Unless stated otherwise, all methods are compute-matched by performing parameter updates on exactly 30B update tokens. For GPT-2 models, we keep most modules in FP32 but cast the token embedding layers to BF16 for efficiency. We also evaluate OPUS in a  continued pre-training setting using the Qwen3-8B-Base ~\citep{qwen3}. This model architecture comprises 36 layers with a hidden size of 4096 and approximately 8B parameters. In this configuration, the model is adapted on a science-domain stream, keep the training recipe fixed, and vary only the selection policy. We train with mixed precision in \texttt{bfloat16}. For Qwen3-8B-Base models, we cast the entire model to BF16 to maintain dtype consistency. All experiments run with synchronous data-parallel training using NCCL. Let $W$ be the number of GPUs (world size) and $G$ be the gradient accumulation steps; then the global batch size per optimizer update is $B = W \cdot G$ sequences of length $L$, i.e., $W\cdot G\cdot L$ update tokens per step. We apply global gradient-norm clipping with threshold $1.0$.

\noindent \textbf{Sequence lengths, batch sizes for OPUS.}
We use model-specific training sequence lengths due to memory constraints. For GPT-2 we set $L_{\text{train}}{=}24{,}576$ (GPT-2 Large) and $L_{\text{train}}{=}6{,}144$ (GPT-2 XL), with $L_{\text{val}}{=}32{,}768$ (Large) and $L_{\text{val}}{=}8{,}192$ (XL).\footnote{For the Qwen3-8B CPT runs, we use $L_{\text{train}}{=}4{,}096$ and $L_{\text{val}}{=}4{,}096$ with FlexAttention.} 
For OPUS, at each optimization step we score candidates using only $L_{\text{score}}{=}512$ tokens of each sequence. We form a candidate buffer of $N{=}32$ sequences for GPT-2 runs. For Qwen3-8B, we use $M{=}16$ as a buffer-size multiplier; selection is performed \emph{globally} by gathering scores across all GPUs and selecting the top $K{=}\lfloor \rho N \rfloor$ sequences with $\rho{=}0.5$. 
We use the validation split as the proxy set for scoring (proxy batch size $8$) and refresh it every step. After selection, the model performs a full forward/backward update on the selected sequences of length $L_{\text{train}}$, and the token budget is counted using $L_{\text{train}}$. The additional forward computation used for scoring is treated as overhead (Sec.~\ref{sec:efficiency}). Random projection is disabled in these runs unless stated otherwise.

\noindent \textbf{Optimizers and hyperparameters.}
We evaluate two optimizer settings under the same learning-rate schedule and training recipe. In Muon setting, we apply Muon~\citep{muon}\footnote{The optimizer employed in our implementation combines Muon and AdamW. To simplify notation, we use ``Muon'' as shorthand for this hybrid optimizer in the remainder of this paper.
} updates to matrix-shaped parameters and use AdamW~\citep{adamw} for parameter types where Muon-style matrix preconditioning is not directly applicable, such as biases and normalization parameters. In AdamW setting, we use AdamW~\citep{adamw} for all parameters as a unified baseline. 

\begin{table}[t]
\centering
\small
\setlength{\tabcolsep}{5.0pt}
\renewcommand{\arraystretch}{1.08}
\caption{\textbf{Optimizer assignment by parameter.} In our Muon+AdamW setting, Muon is applied to matrix-shaped parameters inside Transformer blocks (\texttt{model.blocks}, $\texttt{ndim}\ge 2$), while AdamW is applied to embeddings, LM head, and all 0/1D parameters. In the AdamW setting, AdamW is applied to all parameters. Patterns with \texttt{i=0..L-1} repeat per Transformer layer.}
\label{tab:opt_param_assignment}
\resizebox{0.97\textwidth}{!}{%
\begin{tabular}{l l l c l l}
\toprule
\textbf{Model} & \textbf{Parameter pattern} & \textbf{Repeats} & \textbf{ndim} & \textbf{Optimizer} & \textbf{Notes} \\
\midrule

% ===================== GPT-2 Large =====================
GPT2-Large & \texttt{embed.weight} & -- & 2D & AdamW & Token embedding table \\
GPT2-Large & \texttt{lm\_head.weight} & -- & 2D & AdamW & Tied to \texttt{embed.weight} \\
GPT2-Large & \texttt{blocks.\{i\}.attn.qkv\_proj.weight} & $i{=}0..35$ & 2D & Muon & Attention QKV projection \\
GPT2-Large & \texttt{blocks.\{i\}.attn.c\_proj.weight} & $i{=}0..35$ & 2D & Muon & Attention output projection \\
GPT2-Large & \texttt{blocks.\{i\}.mlp.c\_fc.weight} & $i{=}0..35$ & 2D & Muon & MLP expansion projection \\
GPT2-Large & \texttt{blocks.\{i\}.mlp.c\_proj.weight} & $i{=}0..35$ & 2D & Muon & MLP contraction projection \\
\midrule

% ===================== GPT-2 XL =====================
GPT2-XL & \texttt{embed.weight} & -- & 2D & AdamW & Token embedding table \\
GPT2-XL & \texttt{lm\_head.weight} & -- & 2D & AdamW & Tied to \texttt{embed.weight}\\
GPT2-XL & \texttt{blocks.\{i\}.attn.qkv\_proj.weight} & $i{=}0..47$ & 2D & Muon & Attention QKV projection \\
GPT2-XL & \texttt{blocks.\{i\}.attn.c\_proj.weight} & $i{=}0..47$ & 2D & Muon & Attention output projection \\
GPT2-XL & \texttt{blocks.\{i\}.mlp.c\_fc.weight} & $i{=}0..47$ & 2D & Muon & MLP expansion projection \\
GPT2-XL & \texttt{blocks.\{i\}.mlp.c\_proj.weight} & $i{=}0..47$ & 2D & Muon & MLP contraction projection \\
\midrule

% ===================== Qwen3-8B =====================
Qwen3-8B-Base & \texttt{embed.weight} & -- & 2D & AdamW & Token embedding table \\
Qwen3-8B-Base & \texttt{lm\_head.weight} & -- & 2D & AdamW & Tied \\
Qwen3-8B-Base & \texttt{ln\_f.weight} & -- & 1D & AdamW & Final RMSNorm weight \\
Qwen3-8B-Base & \texttt{blocks.\{i\}.input\_layernorm.weight} & $i{=}0..35$ & 1D & AdamW & RMSNorm weight \\
Qwen3-8B-Base & \texttt{blocks.\{i\}.post\_attention\_layernorm.weight} & $i{=}0..35$ & 1D & AdamW & RMSNorm weight \\
Qwen3-8B-Base & \texttt{blocks.\{i\}.self\_attn.q\_norm.weight} & $i{=}0..35$ & 1D & AdamW & QK-norm weight \\
Qwen3-8B-Base & \texttt{blocks.\{i\}.self\_attn.k\_norm.weight} & $i{=}0..35$ & 1D & AdamW & QK-norm weight \\
Qwen3-8B-Base & \texttt{blocks.\{i\}.self\_attn.q\_proj.weight} & $i{=}0..35$ & 2D & Muon & Attention Q projection \\
Qwen3-8B-Base & \texttt{blocks.\{i\}.self\_attn.k\_proj.weight} & $i{=}0..35$ & 2D & Muon & Attention K projection \\
Qwen3-8B-Base & \texttt{blocks.\{i\}.self\_attn.v\_proj.weight} & $i{=}0..35$ & 2D & Muon & Attention V projection \\
Qwen3-8B-Base & \texttt{blocks.\{i\}.self\_attn.o\_proj.weight} & $i{=}0..35$ & 2D & Muon & Attention output projection \\
Qwen3-8B-Base & \texttt{blocks.\{i\}.mlp.gate\_proj.weight} & $i{=}0..35$ & 2D & Muon & SwiGLU gate projection \\
Qwen3-8B-Base & \texttt{blocks.\{i\}.mlp.up\_proj.weight} & $i{=}0..35$ & 2D & Muon & SwiGLU up projection \\
Qwen3-8B-Base & \texttt{blocks.\{i\}.mlp.down\_proj.weight} & $i{=}0..35$ & 2D & Muon & SwiGLU down projection \\
\midrule

% ===================== Safety net =====================
All & \texttt{(any remaining parameters)} & -- & any & AdamW &  \\
\bottomrule
\end{tabular}
}
\end{table}

\smallskip
\noindent\textbf{Optimizer assignment.}
For clarity and reproducibility, we explicitly specify how parameters are assigned to optimizers in our experimental settings (Table~\ref{tab:opt_param_assignment}).
In the Muon setting, we apply Muon updates only to \emph{matrix-shaped} parameters inside Transformer blocks, i.e., parameters under \texttt{model.blocks} with $\texttt{ndim}\ge 2$ (e.g., attention and MLP projection matrices).
All remaining parameters—including token embeddings, the LM head, and all 0/1D parameters such as RMSNorm weights and biases—are optimized with a distributed AdamW optimizer.
This hybrid design follows the recommended usage of Muon, which is intended for 2D matrices and is not directly applicable to 0/1D parameter types.
In the AdamW setting, we instead optimize all parameters with AdamW optimizer.

\smallskip
\noindent\textbf{Muon optimizer configuration.}
We use a hybrid optimizer in which Muon updates the matrix parameters inside Transformer blocks (parameters with $\texttt{ndim}\ge 2$), excluding the token embedding table and the final LM head. All remaining parameters are updated with AdamW. Muon applies SGD with momentum ($\mu=0.95$) with no weight decay, followed by an orthogonalization post-processing step on each 2D update. Specifically, we run a Newton--Schulz quintic iteration for $5$ steps in BF16 to produce an approximate zeroth-power transform, serving as an efficient surrogate to the $UV^\top$ factor in SVD-based orthogonalization. To stabilize updates across differently-shaped matrices, we rescale the effective learning rate for each matrix parameter $W\in\mathbb{R}^{m\times n}$ as
\[
\eta_{\text{eff}}=\eta\cdot\sqrt{\max\!\left(1,\frac{m}{n}\right)}.
\]
For the AdamW-updated parameter groups in this hybrid setup, we use $\beta_1=0.8$, $\beta_2=0.95$, $\epsilon=10^{-8}$, and weight decay $\lambda=0$, synchronizing gradients via memory-efficient reduce-scatter when dimensions are divisible by the world size and otherwise falling back to all-reduce for correctness.

\smallskip
\noindent\textbf{AdamW optimizer configuration.}
For settings that use AdamW, we update all model parameters---including token embeddings, all Transformer block parameters, and the final LM head---with a distributed AdamW optimizer using $\beta_1=0.8$, $\beta_2=0.95$, $\epsilon=10^{-8}$, and weight decay $\lambda=0$. Gradients are synchronized using reduce-scatter when tensor dimensions are divisible by the world size, and otherwise using all-reduce to ensure numerically correct distributed updates.

\noindent \textbf{Learning rate and optimization hyperparameters.}
For GPT-2 XL, we use $\mathrm{lr}_{\text{adam}}{=}2{\times}10^{-3}$ and $\mathrm{lr}_{\text{muon}}{=}1{\times}10^{-2}$. AdamW uses $\beta_1{=}0.8$, $\beta_2{=}0.95$, $\epsilon{=}10^{-8}$, and no weight decay ($\lambda{=}0$). Muon uses momentum $\mu{=}0.95$ with a short warmup from $0.85\rightarrow 0.95$ over the first $300$ steps, and no weight decay. 
For Qwen3-8B CPT (SciPedia), we use $\mathrm{lr}_{\text{adam}}{=}10^{-6}$ and $\mathrm{lr}_{\text{muon}}{=}10^{-5}$ with AdamW hyperparameters $\beta_1{=}0.9$, $\beta_2{=}0.95$, and weight decay $\lambda{=}0.01$. We apply global gradient-norm clipping with threshold $1.0$ in all experiments. 
The global batch per optimization step is $B{=}W\cdot G$ sequences of length $L$, where $W$ is the number of GPUs and $G$ is the number of gradient-accumulation steps (Qwen3-8B uses $W{=}8$ and $G{=}1$). We train Qwen3-8B for a token budget of $1.5$B tokens and evaluate every $0.5$B tokens. The learning-rate schedule is implemented as a piecewise multiplier over the base LR with a warmup fraction of $0.01$.

\smallskip
\noindent\textbf{Random projection configuration.}
To accelerate OPUS scoring, we apply a CountSketch-based random projection to per-sample gradients, implementing the sketching operator. Concretely, for each trainable linear weight we form the per-sample gradient in outer-product form (aggregated over time when applicable) and then sketch the flattened gradient into an $m$-dimensional vector using CountSketch with a deterministic hash/sign pair; this yields an unbiased estimator of inner products, $\mathbb{E}\langle \Pi(g_1),\Pi(g_2)\rangle=\langle g_1,g_2\rangle$, enabling us to compute gradient dot-products (and similarity matrices) in the projected space. We set the sketch dimension to $m=8192$ with seed $42$, which provides substantial compression for GPT-2 XL where the largest matrix-gradient has dimension on the order of $10.24$M, corresponding to an effective compression of roughly $1250 \times$ while preserving the ranking signal used by OPUS. The projection is enabled during scoring and uses cached hash/sign tensors per parameter shape for efficiency; when disabled, we fall back to exact full-dimensional dot-products.

\begin{table}[tb!]
\caption{\textbf{Benchmark evaluation configuration.} For most benchmarks we use multiple-choice perplexity: score each candidate option by negative log-likelihood and choose the best-scoring option; we report accuracy. MMLU is evaluated separately using zero-shot and log-likelihood on the entire answer following FineWeb-Edu.}
\label{tab:eval_config_all}
\centering
\small
\setlength{\tabcolsep}{15.0pt}
\renewcommand{\arraystretch}{1.05}
\resizebox{0.99\textwidth}{!}{%
\begin{tabular}{l l c l l}
\toprule
\textbf{Benchmark} & \textbf{Domain} & \textbf{\#Choices} & \textbf{Eval mode} & \textbf{Metric} \\
\midrule
\multicolumn{5}{c}{\textit{Core Benchmarks (in-domain)}} \\
\midrule
MMLU & Knowledge & 4 & LL & Accuracy \\
ANLI & Understanding & 3 & PPL & Accuracy \\
HellaSwag & Commonsense and Reasoning & 4 & PPL & Accuracy \\
PIQA & Commonsense and Reasoning & 2 & PPL & Accuracy \\
SIQA & Commonsense and Reasoning & 3 & PPL & Accuracy \\
WinoGrande & Language & 2 & LL & Accuracy \\
ARC-Easy & Science and Reasoning & 4 & PPL & Accuracy \\
ARC-Challenge & Science and Reasoning & 4 & PPL & Accuracy \\
CommonsenseQA & Commonsense and Reasoning & 5 & PPL & Accuracy \\
WSC & Language & 2 & PPL & Accuracy \\
\midrule
\multicolumn{5}{c}{\textit{Other Benchmarks (out-of-domain)}} \\
\midrule
BBH & Reasoning (hard) & -- & Generation & Exact Match \\
RACE-Middle & Understanding & 4 & PPL & Accuracy \\
RACE-High & Understanding & 4 & PPL & Accuracy \\
AX-b & Language & 2 & PPL & Accuracy \\
AX-g & Language & 2 & PPL & Accuracy \\
StoryCloze & Understanding & 2 & PPL & Accuracy \\
\bottomrule
\end{tabular}%
}
\end{table}

\noindent \textbf{Pre-training corpus.}
For from-scratch pre-training, all methods draw candidates from the same 3T-token pool constructed from FineWeb~\citep{finewebedu}. To test robustness on a higher-quality corpus, we also run the same recipe on FineWeb-Edu~\citep{finewebedu}. FineWeb-Edu provides a document-level quality classifier that assigns each document a discrete score in \{3,4,5\}. We partition the FineWeb-Edu pool into two buckets: a 120B-token mid-quality bucket consisting of all score-3 documents, and a 80B-token high-quality bucket formed by merging score-4 and score-5 documents. For static filtering baselines, we score the full pool once and materialize a fixed 30B-token subset for training. For dynamic methods, candidates are streamed from the pool and selected during training. For CPT, we construct a 3B-token pool from SciencePedia~\citep{sciencepedia} for continued pre-training.

\noindent \textbf{Evaluation.} We evaluate all GPT-2 pretraining checkpoints on a variety of benchmarks target diverse capabilities. See Table ~\ref{tab:eval_config_all} for the summary of the configurations.

Specifically, we evaluate on the following benchmarks to test the general capabilities of our pretrained models:
\begin{itemize}[leftmargin=*,topsep=1pt,itemsep=1pt]
    \item \textbf{MMLU}~\citep{mmlu}: broad factual and academic knowledge across many subjects.
    \item \textbf{ANLI}~\citep{anli}: adversarial natural language inference, testing robust entailment and contradiction reasoning.
    \item \textbf{HellaSwag}~\citep{hellaswag}: commonsense reasoning for plausible continuations.
    \item \textbf{PIQA}~\citep{piqa}: physical commonsense reasoning about everyday actions.
    \item \textbf{SIQA}~\citep{siqa}: social commonsense and intent reasoning.
    \item \textbf{WinoGrande}~\citep{winogrande}: pronoun/coreference resolution with adversarial bias reduction.
    \item \textbf{ARC-E / ARC-C}~\citep{arc}: grade-school science questions; Easy and Challenge splits measure increasing reasoning difficulty.
    \item \textbf{CommonsenseQA}~\citep{commensenseqa}: commonsense knowledge and reasoning over concepts.
    \item \textbf{WSC}~\citep{wsc}: hard coreference requiring commonsense.
\end{itemize}

For all above benchmarks except for MMLU, we use OpenCompass~\citep{2023opencompass} with a multiple-choice perplexity scoring rule: for each candidate answer option, we compute its average negative log-likelihood conditioned on the prompt, and predict the option with the lowest perplexity; we then report accuracy.
For WinoGrande, we follow the OpenCompass log-likelihood variant that compares the likelihood of the two candidates. All these benchmarks are evaluated zero-shot. MMLU is evaluated separately with Lighteval~\citep{lighteval} following the implementation in FineWeb-Edu~\citep{finewebedu} evaluation protocol. Since the typical MMLU implementation (which uses "A", "B", etc as answer targets) gives generally random results on non instruction tuned models, instead, we use the full MMLU answer as the target. We also use zero-shot prompting and then select the answer by comparing the log-likelihood of the entire option string. 

In addition, we use the following benchmarks that are not in our bench-proxy set for the generalization evaluation:
\begin{itemize}[leftmargin=*,topsep=1pt,itemsep=1pt]
    \item \textbf{BBH}~\citep{bbh}: a challenging subset of BIG-Bench tasks emphasizing multi-step reasoning. We select a set of BBH tasks where base models produce non-degenerate outputs: Tracking Shuffled Objects, Reasoning about Colored Objects, Logical Deduction, Disambiguation QA, Penguins in a Table, and Sports Understanding.
    \item \textbf{RACE-M / RACE-H}~\citep{race}: exam-style reading comprehension with multiple choice questions; we use the Middle and High school subsets.
    \item \textbf{AX-B / AX-G}~\citep{superglue}: diagnostic evaluation sets from SuperGLUE designed to stress-test linguistic phenomena and generalization.
    \item \textbf{StoryCloze}~\citep{storycloze}: story ending prediction to test narrative coherence and commonsense continuation.
\end{itemize}

We evaluate these benchmarks using the OpenCompass framework. All these benchmarks are evaluated zero-shot except for BBH, which uses three-shot. For BBH, many subtasks are near-chance at our model scale, so an aggregate score over all subtasks becomes unstable and less informative. We therefore report results on the curated subset above, where the base model achieves non-trivial accuracy and methods exhibit meaningful separation.

\noindent\textbf{CPT evaluation.} We evaluate continued pre-training checkpoints of Qwen3-8B-Base on two science focused benchmarks, OlympicArena~\citep{olympicarena} and SciAssess~\citep{sciassess}.
For OlympicArena, we evaluate on the test split and use zero-shot prompting.
For SciAssess, we evaluate four subdomains in biology, chemistry, material, medicine using a 3-shot prompting setting with chain-of-thought enabled where available. We use stochastic decoding with temperature $0.6$, top-$p=0.95$, and top-$k=20$, and max sequence length of 1024. We report the official accuracy metric for both benchmarks.

\noindent \textbf{Baselines.}
We compare OPUS against representative data selection methods. (1) Static baselines. We evaluate five representative static filtering methods: QuRating~\citep{qurating}, DSIR~\citep{dsir}, DCLM-FastText~\citep{dclm}, FineWeb-Edu Classifier~\citep{finewebedu}, and UltraFineweb Classifier~\citep{ultrafineweb}. (2) Dynamic selection. We include \textsc{High-PPL} (PPL), which selects the highest-loss sequences under the current model following~\citep{ankner2025perplexed}, and GREATS~\citep{Greats}, which selects samples whose per-sample gradients best align with a SGD-based proxy direction in post-training. We also report results of random selection at 30B and 60B update tokens for baseline comparison.

\subsection{Bench-proxy construction}
\label{sec:bench_proxy}

% This part details how we construct the \textsc{Bench-Proxy} (\textsc{Bench-Proxy})
% used to estimate the validation direction in Eq.~\eqref{eq:score_with_proxy_precond}, following the retrieval
% pipeline in Fig.~\ref{fig:pipeline}(a).
% Our goal is to obtain a small proxy dataset $\mathcal{D}_{\text{proxy}}$ that is \emph{in-distribution} with respect
% to a target downstream benchmark, while still being drawn from the pre-training corpus so that gradients can be
% computed efficiently and consistently during pre-training.

We describe how to construct \textsc{Bench-Proxy}, which estimates the validation direction in
Eq.~\eqref{eq:score_with_proxy_precond} via the retrieval pipeline in Fig.~\ref{fig:pipeline}(a).
The goal is to build a small proxy set $\mathcal{D}_{\text{proxy}}$ that matches the target benchmark’s distribution,
while being sampled from the pre-training corpus so gradients can be computed efficiently and consistently during
pre-training.

\noindent \textbf{Similarity scoring.}
We first assign each pre-training document a benchmark relevance score based on its semantic similarity to the
benchmark validation set $\mathcal{D}_{\text{val}}$.
Concretely, we use a frozen sentence embedding model Arctic-Embed-L v2 \citep{Arctic-Embed} to encode (i) each benchmark sample and
(ii) each pre-training document into a shared embedding space, and compute cosine similarities between document
embeddings and benchmark embeddings.
To obtain a single scalar score per document, we reduce the similarity vector by taking the maximum similarity over
all benchmark samples, which captures whether a document is strongly aligned with \emph{any} benchmark instance.
This produces a scored version of the pre-training corpus, where each document is annotated with a benchmark alignment score.

\noindent \textbf{Proxy construction.}
We then construct the proxy pool $\mathcal{D}_{\text{proxy}}$ by selecting the highest-scoring documents from the scored corpus.
In practice, we sort documents by their benchmark relevance scores in descending order and greedily accumulate them
until reaching a fixed token budget (\textbf{30M tokens} in our experiments), which yields a compact but benchmark-aligned proxy shard.
During training, we repeatedly sample mini-batches from $\mathcal{D}_{\text{proxy}}$ to estimate the proxy gradient direction
used for within-step ranking.
This design keeps scoring stable and low-variance, while steering selection toward data that matches the target benchmark distribution.

\begin{table*}[tb!]
\caption{\textbf{Evaluation results after training on FineWeb dataset with 30B tokens}. Blocks correspond to model size and optimizer. Bold marks the best compute-matched method per benchmark within each block; a longer-training random-sampling reference at 60B update tokens is included for context. Abbreviations: W.G. = Winogrande; C.QA = CommonsenseQA; WSC = Winograd Schema Challenge.}
\centering
\resizebox{.99\textwidth}{!}{
\begin{tabular}{@{}lcccccccccc|c}
\toprule
Method & MMLU & ANLI & HellaSwag & PIQA & SIQA & W.G. & ARC-E & ARC-C & C.QA & WSC & Avg. \\
\midrule

% -------------------------------
% GPT-2 Large + Muon
% -------------------------------
\multicolumn{12}{c}{\textit{GPT-2 Large with Muon optimizer on 30B update tokens of FineWeb}} \\
\midrule
Random & 28.46 & 32.93 & 42.71 & 69.70 & 40.07 & 49.17 & 37.57 & 28.14 & 31.94 & 36.54 & 39.72 \\
PPL & 28.40 & 33.24 & 42.69 & 70.13 & 40.17 & 48.38 & 36.16 & 23.05 & 31.86 & 36.54 & 39.06 \\
GREATS~\citep{Greats} & 28.49 & 33.31 & 42.22 & 70.18 & 39.46 & 49.41 & 36.86 & 24.41 & 33.25 & 36.54 & 39.41 \\
QuRating~\citep{qurating} & \textbf{31.53} & \textbf{34.12} & 39.47 & 66.38 & 39.82 & \textbf{50.59} & \textbf{40.92} & \textbf{30.51} & 30.22 & \textbf{38.46} & 40.20 \\
DSIR~\citep{dsir} & 28.50 & 33.39 & 43.04 & 69.70 & \textbf{40.53} & 49.64 & 37.39 & 24.41 & 32.27 & 36.54 & 39.54 \\
DCLM-FastText~\citep{dclm} & 29.36 & 33.17 & 44.26 & \textbf{71.16} & 39.82 & 49.96 & 37.92 & 24.75 & 32.02 & 36.54 & 39.90 \\
FineWeb-Edu~\citep{finewebedu} & 28.83 & 32.67 & 43.09 & 70.02 & 40.28 & 47.75 & 39.15 & 24.75 & 33.66 & \textbf{38.46} & 39.87 \\
UltraFineweb~\citep{ultrafineweb} & 29.00 & 32.99 & \textbf{44.38} & 71.11 & 40.17 & 48.78 & 37.57 & 25.08 & \textbf{33.91} & \textbf{38.46} & 40.15 \\
\rowcolor{cyan!5} OPUS (Ours) & 28.76 & 33.12 & 42.92 & 69.97 & 39.56 & 50.43 & 38.98 & 29.15 & 33.09 & 36.54 & \textbf{40.25} \\
\midrule
Random (60B) & 28.70 & 33.23 & 45.20 & 71.16 & 40.79 & 49.41 & 39.68 & 25.42 & 31.12 & 36.54 & 40.13 \\
\midrule

% -------------------------------
% GPT-2 XL + Muon
% -------------------------------
\multicolumn{12}{c}{\textit{GPT-2 XL with Muon optimizer on 30B update tokens of FineWeb}} \\
\midrule
Random & 28.73 & 33.98 & 48.01 & 70.46 & 39.61 & 47.91 & 38.98 & 25.42 & \textbf{33.25} & 36.54 & 40.29 \\
PPL & 29.35 & 33.42 & 47.87 & 71.55 & 40.69 & 45.86 & 38.45 & 24.07 & 30.38 & 36.54 & 39.82 \\
GREATS~\citep{Greats} & 29.95 & 33.58 & 42.26 & 70.18 & 39.61 & 47.67 & 36.33 & 23.73 & 30.55 & 38.46 & 39.23 \\
QuRating~\citep{qurating} & \textbf{33.28} & 33.19 & \textbf{48.62} & 70.95 & 41.20 & \textbf{48.70} & 37.04 & 26.78 & 30.88 & 36.54 & 40.72 \\
DSIR~\citep{dsir} & 29.58 & 33.98 & 48.49 & \textbf{71.93} & 39.51 & 47.59 & 38.10 & 26.44 & 32.68 & 38.46 & 40.68 \\
DCLM-FastText~\citep{dclm} & 30.40 & \textbf{34.08} & 44.07 & 71.38 & \textbf{41.97} & 48.38 & 38.80 & \textbf{29.49} & 30.88 & 36.54 & 40.60 \\
FineWeb-Edu~\citep{finewebedu} & 29.66 & 33.12 & 48.45 & 71.71 & 41.25 & 46.17 & 39.19 & 28.14 & 31.29 & 38.46 & 40.74 \\
UltraFineweb~\citep{ultrafineweb} & 29.95 & 33.31 & 43.11 & 70.57 & 40.79 & 47.51 & 36.51 & 26.44 & 31.70 & 36.54 & 39.64 \\
\rowcolor{cyan!5} OPUS (Ours) & 29.89 & 33.29 & 48.39 & 71.27 & 41.10 & 47.99 & \textbf{39.68} & 26.44 & 31.37 & \textbf{48.08} & \textbf{41.75} \\
\midrule
Random (60B) & 30.24 & 33.84 & 51.10 & 72.25 & 40.89 & 48.78 & 41.98 & 23.05 & 32.35 & 38.46 & 41.29 \\
\midrule

% -------------------------------
% GPT-2 Large + AdamW
% -------------------------------
\multicolumn{12}{c}{\textit{GPT-2 Large with AdamW on 30B update tokens of FineWeb}} \\
\midrule
Random & 28.19 & 32.91 & 42.65 & 69.37 & \textbf{40.79} & 50.12 & 37.21 & 25.08 & 30.06 & 36.54 & 39.29 \\
PPL & 28.69 & 33.44 & 42.23 & 68.77 & 40.43 & 47.36 & 36.68 & 22.37 & 32.84 & 36.54 & 38.94 \\
GREATS~\citep{Greats} & 28.77 & 33.46 & 43.00 & 70.46 & 40.63 & 49.96 & 38.45 & 23.39 & 32.02 & 36.54 & 39.67 \\
QuRating~\citep{qurating} & \textbf{31.87} & 33.08 & 43.22 & 70.24 & 40.74 & 49.88 & 37.21 & 24.75 & 33.58 & 36.54 & 40.11 \\
DSIR~\citep{dsir} & 28.22 & 33.18 & 43.42 & 69.53 & 40.02 & 48.93 & 37.92 & 25.08 & 31.20 & \textbf{38.46} & 39.60 \\
DCLM-FastText~\citep{dclm} & 29.11 & 33.05 & 43.60 & \textbf{70.67} & 39.41 & 47.51 & 39.33 & 25.08 & 33.42 & 36.54 & 39.77 \\
FineWeb-Edu~\citep{finewebedu} & 29.03 & \textbf{35.41} & 42.82 & 70.29 & 40.38 & 47.51 & 39.51 & \textbf{27.12} & 31.86 & \textbf{38.46} & 40.24 \\
UltraFineweb~\citep{ultrafineweb} & 29.05 & 33.51 & 43.51 & \textbf{70.67} & 40.38 & 48.62 & 41.62 & 25.76 & \textbf{34.15} & 36.54 & 40.38 \\
\rowcolor{cyan!5} OPUS (Ours) & 31.09 & 34.04 & \textbf{45.52} & 69.97 & 40.69 & \textbf{51.62} & \textbf{42.50} & 26.44 & 33.99 & \textbf{38.46} & \textbf{41.43} \\
\midrule
Random (60B) & 29.08 & 33.08 & 44.40 & 70.89 & 41.15 & 48.70 & 37.74 & 22.03 & 32.43 & 36.54 & 39.60 \\
\midrule

% -------------------------------
% GPT-2 XL + AdamW
% -------------------------------
\multicolumn{12}{c}{\textit{GPT-2 XL with AdamW optimizer on 30B update tokens of FineWeb}} \\
\midrule
Random & 28.76 & 33.56 & \textbf{46.63} & 70.35 & 42.37 & 49.19 & 39.15 & 24.41 & 32.68 & 36.54 & 40.36 \\
PPL & 29.32 & 33.67 & 45.31 & 70.08 & 41.71 & 49.72 & 39.68 & 24.75 & 31.29 & \textbf{38.46} & 40.02 \\
GREATS~\citep{Greats} & 28.81 & 33.49 & 40.73 & 69.53 & \textbf{42.48} & 49.01 & 34.22 & 24.75 & 31.04 & \textbf{38.46} & 39.25 \\
QuRating~\citep{qurating} & \textbf{32.24} & 32.61 & 34.66 & 66.65 & 38.54 & \textbf{50.43} & 36.86 & 24.75 & 28.42 & 36.54 & 38.71 \\
DSIR~\citep{dsir} & 29.37 & 33.09 & 45.88 & 70.67 & 39.97 & 47.51 & 38.80 & 24.41 & 33.42 & 36.54 & 39.97 \\
DCLM-FastText~\citep{dclm} & 29.43 & \textbf{34.47} & 42.45 & 69.91 & 41.86 & 47.59 & 36.33 & 24.41 & 31.53 & 36.54 & 39.45 \\
FineWeb-Edu~\citep{finewebedu} & 29.71 & 33.51 & 46.62 & \textbf{71.93} & 41.91 & 46.88 & \textbf{40.04} & 25.08 & 32.10 & 36.54 & 40.43 \\
UltraFineweb~\citep{ultrafineweb} & 29.25 & 33.51 & 41.76 & 69.21 & 41.40 & 49.57 & 37.92 & 24.07 & 32.76 & 36.54 & 39.60 \\
\rowcolor{cyan!5} OPUS (Ours) & 29.43 & 33.51 & 46.12 & 70.35 & 41.35 & 50.36 & 39.33 & \textbf{29.15} & \textbf{33.99} & 36.54 & \textbf{41.01} \\
\midrule
Random (60B) & 29.55 & 33.57 & 48.75 & 72.09 & 41.10 & 48.78 & 40.92 & 27.12 & 34.48 & 36.54 & 41.29 \\
\bottomrule
\end{tabular}
}
\label{tab:fineweb_merged}
\end{table*}

\subsection{Pre-training from Scratch}
\label{sec:main_results}

\noindent \textbf{Performance on web-scale corpora: FineWeb.}
We first evaluate OPUS on FineWeb, a standard large-scale web corpus. Table~\ref{tab:fineweb_merged} compares OPUS against prior static and dynamic baselines under a fixed budget of 30B update tokens. Across model scales and optimizer settings, OPUS achieves the best compute-matched average and consistently improves over strong baselines. We also include a longer-training random-sampling reference at 60B update tokens to contextualize the magnitude of these efficiency gains; notably, OPUS often matches or exceeds the performance of baselines trained for twice as long.

\noindent \textbf{Robustness on curated corpora: FineWeb-Edu.}
We next evaluate performance on FineWeb-Edu. To test the limits of our method, we subject OPUS to a strict evaluation regime: it selects dynamically from the lower-quality subset (FineWeb-Edu score 3), whereas baselines are trained on the superior high-quality partition (scores 4 and 5). As shown in Table~\ref{tab:finewebedu_merged}, despite this disadvantage in raw data quality, OPUS matches or exceeds prior methods trained on the superior data. For GPT-2 XL with Muon, OPUS achieves the best compute-matched average of 44.99, outperforming all baselines trained on the higher-quality data partitions.

\noindent \textbf{Optimizer-induced selection matters: strong gains under AdamW and Muon.} Under AdamW, which utilizes diagonal preconditioning, OPUS achieves the best compute-matched performance for both GPT-2 Large and GPT-2 XL (Table~\ref{tab:fineweb_merged}). Crucially, this advantage extends to Muon, which employs non-linear matrix preconditioning via Newton-Schulz orthogonalization. For instance, on GPT-2 XL with Muon optimizer on FineWeb, OPUS outperforms Random selection by a significant margin (40.29 $\to$ 41.75). This empirically validates our central hypothesis: aligning data selection with the preconditioned update trajectory yields a more effective training signal than raw gradient-based selection.

\begin{table*}[tb!]
\caption{\textbf{Evaluation on FineWeb-Edu dataset with 30B tokens}. OPUS is evaluated under a strict constraint: selecting dynamically from the mid-quality subset (score 3), while baselines are trained on the higher-quality partitions (scores $\ge 4$). Bold marks the best compute-matched method per benchmark within each block; Random (60B) is shown as a non compute-matched reference.}
\centering
\resizebox{.99\textwidth}{!}{
\begin{tabular}{@{}lcccccccccc|c}
\toprule
Method & MMLU & ANLI & HellaSwag & PIQA & SIQA & W.G. & ARC-E & ARC-C & C.QA & WSC & Avg. \\
\midrule

\multicolumn{12}{c}{\textit{GPT-2 Large with Muon optimizer on 30B update tokens of FineWeb-Edu}} \\
\midrule
Random (Score 3) & 30.52 & 33.16 & 43.95 & 68.87 & 40.58 & 49.02 & 48.39 & 25.08 & 35.54 & 36.54 & 41.17 \\
Random (Score 4+5) & 32.92 & 33.38 & 41.95 & 67.46 & 38.84 & 47.75 & 53.97 & 29.15 & 30.79 & 36.54 & 41.28 \\
PPL (Score 4+5) & \textbf{33.17} & 33.87 & 42.25 & 67.63 & \textbf{40.33} & 48.22 & 50.79 & 28.47 & 29.48 & \textbf{38.46} & 41.27 \\
GREATS (Score 4+5) & 32.73 & \textbf{34.38} & 45.86 & \textbf{70.95} & 39.30 & 50.36 & 44.62 & 24.75 & 32.92 & \textbf{38.46} & 41.43 \\
QuRating (Score 4+5) & 31.32 & 34.07 & 41.70 & 66.92 & 39.71 & 47.83 & 50.79 & 32.88 & 31.94 & 36.54 & 41.37 \\
DSIR (Score 4+5) & 32.54 & 33.54 & 41.07 & 67.95 & 39.36 & 47.28 & 48.68 & \textbf{33.90} & 29.57 & \textbf{38.46} & 41.24 \\
DCLM-FastText (Score 4+5) & 32.64 & 33.67 & 41.66 & 66.38 & 38.74 & \textbf{51.30} & 49.38 & 30.85 & 31.04 & 36.54 & 41.22 \\
FineWeb-Edu (Score 4+5) & 32.00 & 33.46 & 39.95 & 64.74 & 39.87 & 50.51 & 52.20 & 29.15 & 30.30 & 36.54 & 40.87 \\
UltraFineweb (Score 4+5) & 32.60 & 33.02 & 40.70 & 66.05 & 38.23 & 49.72 & 48.32 & 30.17 & 29.24 & 36.54 & 40.46 \\
\rowcolor{cyan!5} OPUS (Score 3) & 30.39 & 34.31 & \textbf{46.36} & 70.51 & 39.41 & 50.20 & 45.33 & 28.47 & \textbf{33.74} & \textbf{38.46} & 41.72 \\
\rowcolor{cyan!5} OPUS (Score 4+5) & 32.17 & 33.38 & 42.52 & 67.30 & 39.51 & 51.07 & \textbf{54.14} & 30.85 & 31.04 & \textbf{38.46} & \textbf{42.04} \\
\midrule
Random (60B) (Score 4+5) & 33.21 & 34.03 & 43.66 & 67.95 & 40.07 & 50.04 & 52.56 & 31.86 & 31.61 & 36.54 & 42.15 \\

\midrule
\multicolumn{12}{c}{\textit{GPT-2 XL with Muon optimizer on 30B update tokens of FineWeb-Edu}} \\
\midrule
Random (Score 3) & 31.92 & 33.56 & 48.39 & 70.13 & 41.10 & 48.86 & 44.86 & 28.47 & 34.23 & 36.54 & 41.81 \\
Random (Score 4+5) & \textbf{34.32} & \textbf{33.78} & 46.39 & 68.72 & 39.36 & 47.59 & 50.44 & 32.54 & 29.48 & 36.54 & 41.92 \\
PPL (Score 4+5) & 32.60 & 33.58 & 46.14 & 69.10 & 40.33 & 51.70 & 50.79 & 30.17 & 31.78 & 36.54 & 42.27 \\
GREATS (Score 4+5) & 33.58 & 33.02 & 46.32 & 68.93 & 39.61 & \textbf{52.57} & 49.21 & 33.90 & 28.01 & 36.54 & 42.17 \\
QuRating (Score 4+5) & 33.10 & 33.58 & 44.22 & 66.70 & 39.97 & 49.64 & 50.09 & 32.54 & 28.99 & 36.54 & 41.54 \\
DSIR (Score 4+5) & 34.13 & 33.63 & 45.10 & 67.79 & 39.82 & 48.15 & 49.03 & 32.88 & 28.83 & 36.54 & 41.59 \\
DCLM-FastText (Score 4+5) & 33.19 & 33.02 & 44.36 & 68.23 & \textbf{41.15} & 48.86 & 51.32 & \textbf{35.59} & 30.14 & 36.54 & 42.24 \\
FineWeb-Edu (Score 4+5) & 32.94 & 33.64 & 43.14 & 68.28 & 39.61 & 51.30 & \textbf{52.73} & 32.20 & 31.37 & 36.54 & 42.18 \\
UltraFineweb (Score 4+5) & 33.41 & 33.48 & 44.34 & 68.93 & 38.64 & 48.30 & 49.38 & 33.56 & 29.07 & 36.54 & 41.57 \\
OPUS (Score 4+5) & 33.83 & 33.64 & 46.30 & 70.67 & 38.95 & 51.14 & 50.62 & 29.15 & 30.47 & 39.42 & 42.42 \\
\rowcolor{cyan!5} OPUS (Score 3) & 32.62 & 33.11 & \textbf{50.54} & \textbf{72.20} & 41.04 & 51.46 & 47.62 & 30.85 & \textbf{35.63} & \textbf{54.81} & \textbf{44.99} \\
\midrule
Random (60B) (Score 4+5) & 33.77 & 33.54 & 46.94 & 69.64 & 39.82 & 49.80 & 50.44 & 32.54 & 30.96 & 38.46 & 42.59 \\
\bottomrule
\end{tabular}
}
\label{tab:finewebedu_merged}
\end{table*}

\noindent \textbf{Generalization beyond proxy-aligned benchmarks.}
Since OPUS uses a benchmark-matched proxy direction to guide training-time selection, it is important to verify that gains are not merely driven by overfitting to the specific evaluation suite used to construct the proxy.
We therefore evaluate on a set of \textit{out-of-distribution} benchmarks covering challenging reasoning and general language comprehension for generalization evaluation. As shown in Table~\ref{tab:bench_generalization}, OPUS achieves the best performance, suggesting that it reflects more general training signal quality, rather than narrow specialization to the proxy-aligned benchmark.

\begin{table}[tb!]
\caption{\textbf{Evaluation on out-of-distribution benchmarks.} We evaluate the same GPT2-XL checkpoints from Table~\ref{tab:fineweb_merged} on out-of-distribution benchmarks that are not included in  \textsc{Bench-Proxy}. }
\label{tab:bench_generalization}
\centering
\resizebox{0.9\linewidth}{!}{%
\begin{tabular}{@{}lcccccc|c}
\toprule
Method & BBH & RACE-M & RACE-H & AX-b & AX-g & StoryCloze & Avg. \\
\midrule
Random & 9.87 & 24.58 & 25.19 & 52.54 & 50.00 & 66.38 & 38.09 \\
PPL & 9.88 & 24.37 & 25.73 & 54.98 & 51.12 & 67.34 & 38.90 \\
GREATS & 10.44 & 26.04 & 26.04 & 57.34 & 50.84 & 65.79 & 39.42 \\
QuRating & 10.65 & 24.79 & 23.33 & 54.35 & \textbf{51.97} & 66.70 & 38.63 \\
DSIR & 9.92 & 25.07 & 26.21 & 53.53 & 49.44 & \textbf{67.72} & 38.65 \\
DCLM-FastText & 10.65 & 26.53 & 25.59 & 52.08 & \textbf{51.97} & 66.86 & 38.95 \\
FineWeb-Edu & 9.73 & \textbf{26.81} & 25.90 & 55.25 & 50.00 & 66.76 & 39.08 \\
UltraFineweb & 9.69 & 23.26 & 22.58 & 48.73 & 48.31 & 67.13 & 36.62 \\
\rowcolor{cyan!5} OPUS (Ours) & \textbf{11.02} & 25.77 & \textbf{27.50} & \textbf{58.42} & 50.56 & 67.13 & \textbf{40.07} \\
\bottomrule
\end{tabular}%
}
\end{table}

\begin{figure}[htbp]
    \centering
    \includegraphics[width=0.98\linewidth]{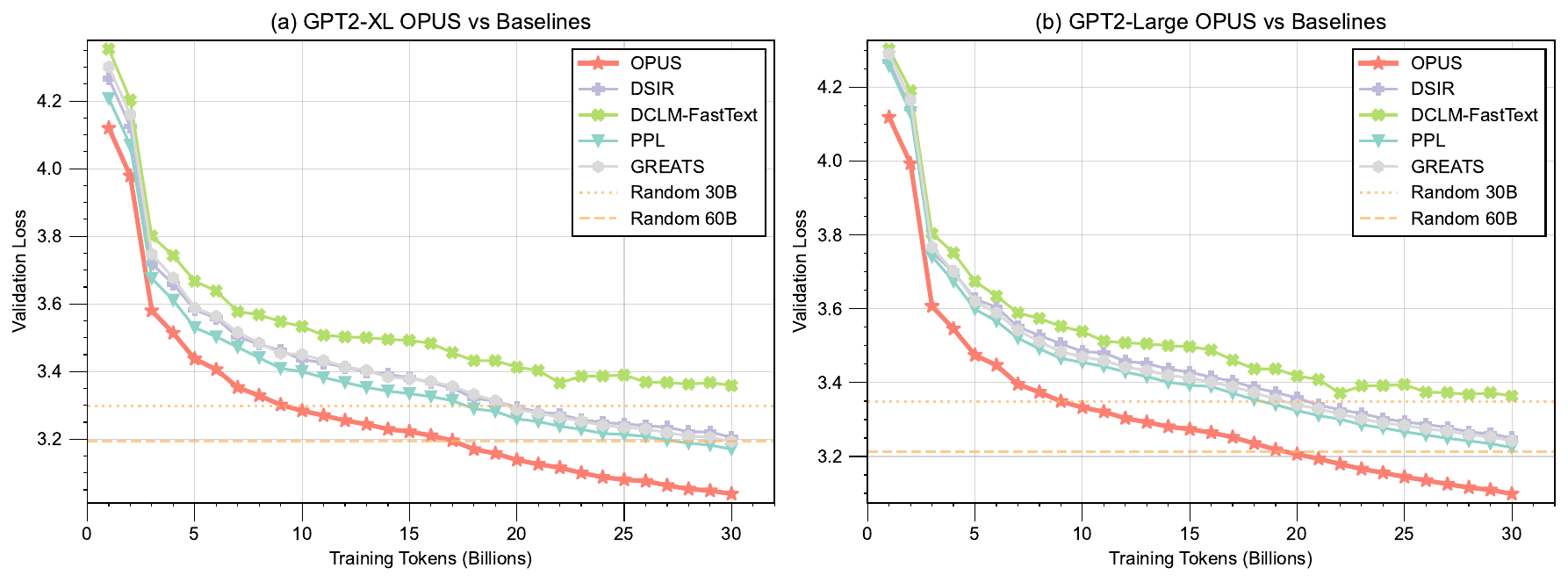}
    \caption{Validation-loss curves on GPT-2 XL and GPT-2 Large pre-trained from scratch on FineWeb-Edu dataset. Left: Results on GPT-2 XL. OPUS compared with representative baselines trained on the high-quality pool, with Random 60B shown as a non compute-matched reference. Curves are shown up to 30B update tokens for compute-matched comparison. Right: Results on GPT2-Large.}
    \label{fig:fwedu_convergence_xl}
\end{figure}

\noindent \textbf{Validation loss curves on FineWeb-Edu dataset.} We report validation-loss trajectories in Figure \ref{fig:fwedu_convergence_xl} for GPT-2 XL and GPT-2 Large trained from scratch on FineWeb-Edu under the same training recipe and a fixed budget of 30B update tokens.
To make the comparison conservative for OPUS, OPUS selects dynamically from the mid-quality pool with score 3, whereas the baselines are trained on the high-quality pool with scores 4+5.
All curves are evaluated on the same held-out FineWeb-Edu validation split.
We also include a longer-training Random reference at 60B update tokens (not compute-matched) to contextualize convergence speed.

As shown in Fig.~\ref{fig:fwedu_convergence_xl}, OPUS consistently improves optimization dynamics for both model scales: across training, it attains lower validation loss than representative baselines despite selecting from the lower-quality candidate pool. For GPT-2 XL, OPUS reaches the validation loss achieved by Random trained for 60B update tokens using only 17B update tokens, demonstrating substantially faster convergence.
For GPT-2 Large, OPUS exhibits the same trend and maintains a clear gap over baselines throughout training.

\noindent \textbf{OPUS enhances knowledge compression measured by domain perplexity across domains.}
To ensure that our selection strategy does not overfit to specific patterns at the expense of broad coverage, we evaluate domain-wise perplexity (PPL). Following the evaluation protocol of \textsc{WebOrganizer}~\citep{wettig2025organize}, we first label documents using the WebOrganizer topic classifier to classify documents into 24 topics and merge these semantically similar topics into ten domains.
We then construct a held-out test set by randomly sampling 1,000 documents from each of ten distinct domains (e.g., Health, Law, Science) to ensure a balanced evaluation. Table~\ref{tab:domain_ppl_fw_edu_xl} indicates that OPUS achieves the lowest average perplexity on FineWeb-Edu dataset.

\begin{table*}[tb!]
\caption{\textbf{Domain-specific perplexity analysis.} Perplexity (PPL; lower is better) on ten domains after 30B update tokens. We construct a validation pool of 10 domains from \citep{wettig2025organize}, containing 1000 held-out samples per domain.}
\centering
\resizebox{.99\textwidth}{!}{
\begin{tabular}{@{}l|cccccccccc|c}
\toprule
Method & Health & Business & Politics & Education & History & Lifestyle & Science & Arts \& Lit. & Entertainment & Computing & Avg. \\
\midrule

\multicolumn{12}{c}{\textit{GPT-2 Large with Muon optimizer on 30B update tokens of FineWeb}} \\
\midrule
Random (30B)     & 3.21 & 3.26 & 3.28 & 3.31 & 3.32 & 3.37 & 3.40 & 3.49 & 3.56 & 3.62 & 3.38 \\
DSIR             & 3.21 & 3.26 & 3.28 & 3.31 & 3.32 & 3.38 & 3.40 & 3.49 & 3.57 & 3.63 & 3.39 \\
DCLM-FastText    & \textbf{3.17} & 3.24 & 3.26 & 3.30 & 3.36 & 3.37 & 3.36 & 3.46 & \textbf{3.54} & 3.60 & 3.37 \\
FineWeb-Edu      & \textbf{3.17} & 3.24 & \textbf{3.25} & \textbf{3.28} & 3.26 & 3.41 & 3.34 & 3.48 & 3.58 & 3.61 & 3.36 \\
QuRating         & 3.40 & 3.60 & 3.79 & 3.57 & 3.68 & 4.05 & 3.61 & 3.92 & 4.27 & 4.11 & 3.80 \\
UltraFineweb     & 3.19 & 3.29 & 3.30 & 3.32 & \textbf{3.30} & 3.43 & 3.38 & 3.50 & 3.59 & 3.62 & 3.39 \\
PPL              & 3.22 & 3.26 & 3.28 & 3.31 & 3.32 & 3.37 & 3.39 & 3.49 & 3.56 & 3.61 & 3.38 \\
GREATS           & 3.25 & 3.31 & 3.33 & 3.36 & 3.38 & 3.42 & 3.46 & 3.55 & 3.62 & 3.66 & 3.43 \\
\rowcolor{cyan!5} OPUS (Ours) & 3.18 & \textbf{3.23} & \textbf{3.25} & \textbf{3.28} & \textbf{3.30} & \textbf{3.34} & \textbf{3.37} & \textbf{3.47} & \textbf{3.54} & \textbf{3.58} & \textbf{3.35} \\

\midrule
\multicolumn{12}{c}{\textit{GPT-2 XL with Muon optimizer on 30B update tokens of FineWeb}} \\
\midrule
Random (30B)     & 3.18 & 3.25 & 3.26 & 3.29 & 3.30 & 3.35 & 3.40 & 3.49 & 3.56 & 3.61 & 3.37 \\
DSIR             & 3.15 & 3.22 & 3.23 & 3.26 & 3.25 & 3.32 & 3.35 & 3.44 & 3.52 & 3.56 & 3.33 \\
DCLM-FastText    & 3.15 & 3.23 & 3.25 & 3.31 & 3.25 & 3.36 & 3.34 & 3.45 & 3.53 & 3.60 & 3.35 \\
FineWeb-Edu      & 3.16 & 3.23 & 3.24 & 3.28 & 3.25 & 3.40 & 3.34 & 3.47 & 3.62 & 3.60 & 3.36 \\
QuRating         & 3.27 & 3.53 & 3.67 & 3.47 & 3.59 & 3.91 & 3.51 & 3.83 & 4.14 & 3.96 & 3.69 \\
UltraFineweb     & 3.10 & 3.20 & 3.19 & 3.24 & 3.21 & 3.33 & 3.29 & 3.41 & 3.50 & 3.53 & 3.30 \\
PPL              & 3.11 & 3.17 & 3.18 & 3.21 & 3.22 & 3.27 & 3.30 & 3.40 & 3.46 & 3.50 & 3.28 \\
GREATS           & 3.22 & 3.29 & 3.29 & 3.33 & 3.32 & 3.39 & 3.42 & 3.51 & 3.58 & 3.66 & 3.40 \\
\rowcolor{cyan!5} OPUS (Ours) & \textbf{3.08} & \textbf{3.15} & \textbf{3.16} & \textbf{3.18} & \textbf{3.21} & \textbf{3.23} & \textbf{3.29} & \textbf{3.39} & \textbf{3.45} & \textbf{3.44} & \textbf{3.26} \\

\midrule
\multicolumn{12}{c}{\textit{GPT-2 Large with Muon optimizer on 30B update tokens of FineWeb-Edu Subset (score $\ge 3$)}} \\
\midrule
Random (30B)     & 3.27 & 3.52 & 3.58 & 3.49 & 3.48 & 3.81 & 3.43 & 3.75 & 4.03 & 3.82 & 3.62 \\
DSIR             & 3.29 & 3.55 & 3.61 & 3.52 & 3.49 & 3.84 & 3.46 & 3.77 & 4.05 & 3.86 & 3.64 \\
DCLM-FastText    & 3.34 & 3.61 & 3.67 & 3.59 & 3.58 & 3.89 & 3.5 & 3.82 & 4.09 & 3.89 & 3.70 \\
FineWeb-Edu      & 3.41 & 3.67 & 3.72 & 3.62 & 3.60 & 3.97 & 3.57 & 3.87 & 4.17 & 3.98 & 3.76 \\
QuRating         & 3.46 & 3.76 & 3.90 & 3.65 & 3.79 & 4.13 & 3.70 & 4.00 & 4.36 & 4.16 & 3.89 \\
UltraFineweb     & 3.42 & 3.72 & 3.87 & 3.66 & 3.77 & 4.05 & 3.58 & 3.96 & 4.26 & 4.00 & 3.83 \\
PPL              & 3.25 & 3.49 & 3.54 & 3.46 & 3.44 & 3.78 & 3.41 & 3.71 & 3.99 & 3.80 & 3.59 \\
GREATS           & 3.29 & 3.55 & 3.62 & 3.52 & 3.50 & 3.84 & 3.46 & 3.77 & 4.06 & 3.86 & 3.65 \\
\rowcolor{cyan!5} OPUS (Ours) & \textbf{3.14} & \textbf{3.34} & \textbf{3.44} & \textbf{3.37} & \textbf{3.37} & \textbf{3.63} & \textbf{3.38} & \textbf{3.63} & \textbf{3.87} & \textbf{3.71} & \textbf{3.49} \\

\midrule
\multicolumn{12}{c}{\textit{GPT-2 XL with Muon optimizer on 30B update tokens of FineWeb-Edu Subset (score $\ge 3$)}} \\
\midrule
Random (30B)     & 3.25 & 3.51 & 3.55 & 3.48 & 3.45 & 3.79 & 3.42 & 3.73 & 4.00 & 3.83 & 3.60 \\
DSIR             & 3.24 & 3.50 & 3.54 & 3.47 & 3.44 & 3.78 & 3.41 & 3.72 & 4.00 & 3.81 & 3.59 \\
DCLM-FastText    & 3.36 & 3.64 & 3.70 & 3.62 & 3.61 & 3.94 & 3.52 & 3.86 & 4.13 & 3.94 & 3.73 \\
FineWeb-Edu      & 3.29 & 3.55 & 3.58 & 3.50 & 3.49 & 3.82 & 3.45 & 3.75 & 4.02 & 3.83 & 3.63 \\
QuRating         & 3.50 & 3.79 & 3.93 & 3.70 & 3.83 & 4.18 & 3.73 & 4.04 & 4.39 & 4.24 & 3.93 \\
UltraFineweb     & 3.43 & 3.74 & 3.90 & 3.68 & 3.80 & 4.07 & 3.59 & 3.99 & 4.28 & 4.02 & 3.85 \\
PPL              & 3.22 & 3.47 & 3.50 & 3.44 & 3.40 & 3.74 & 3.39 & 3.69 & 3.96 & 3.77 & 3.56 \\
GREATS           & 3.29 & 3.55 & 3.60 & 3.52 & 3.49 & 3.84 & 3.45 & 3.77 & 4.05 & 3.88 & 3.64 \\
\rowcolor{cyan!5} OPUS (Ours) & \textbf{3.11} & \textbf{3.31} & \textbf{3.37} & \textbf{3.34} & \textbf{3.31} & \textbf{3.59} & \textbf{3.33} & \textbf{3.58} & \textbf{3.83} & \textbf{3.69} & \textbf{3.45} \\

\bottomrule
\end{tabular}
}
\label{tab:domain_ppl_fw_edu_xl}
\end{table*}

\begin{figure}[tb!]
  \centering
  \includegraphics[width=0.99\linewidth]{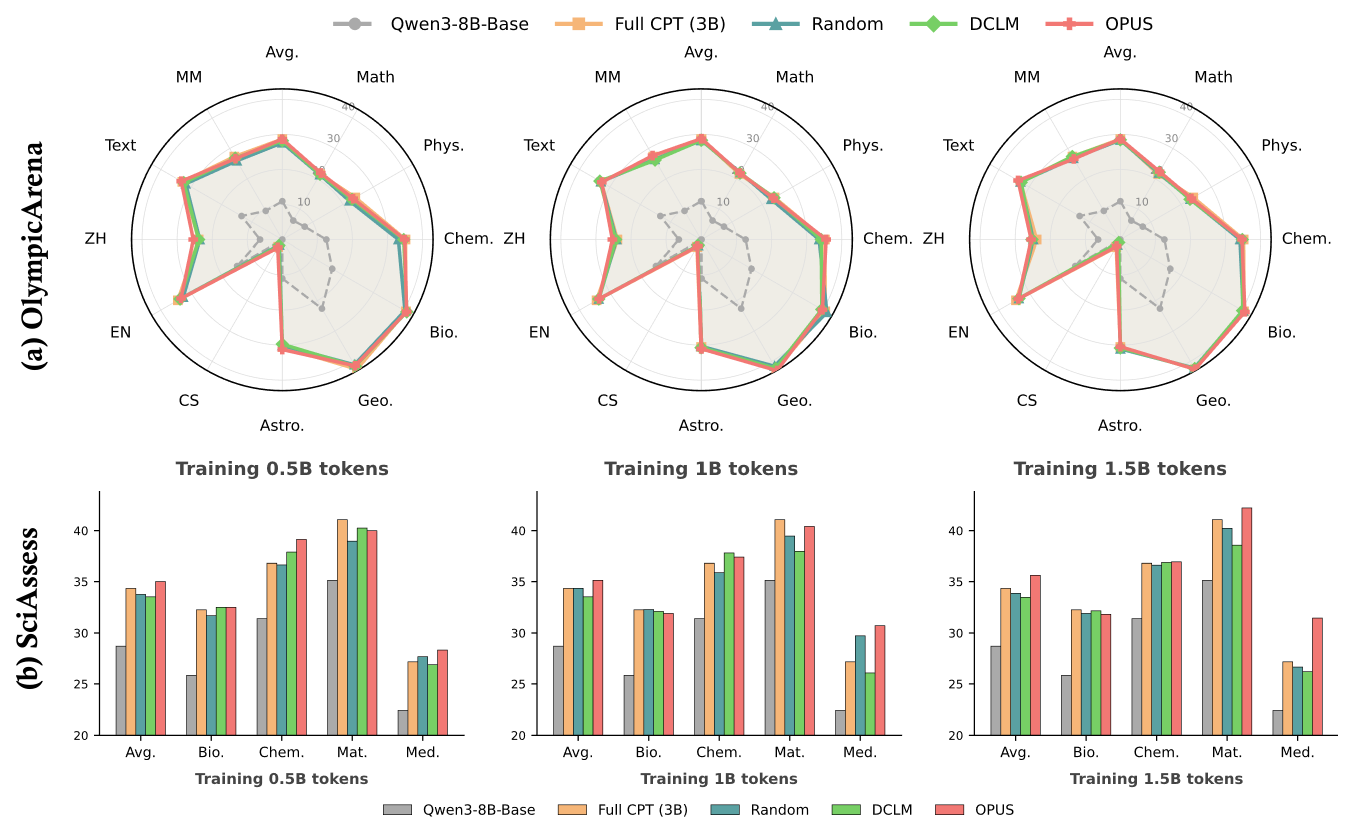}
 \vspace{-10pt} 
 \caption{\textbf{CPT domain breakdown on SciencePedia.}
Domain-level accuracy of Qwen3-8B-Base and CPT baselines across three token budgets 0.5B, 1B, and 1.5B.
Rows correspond to the CPT token budget. Columns show (a) OlympicArena domains with an appended Avg. and (b) SciAssess domains.
For each panel, we compare Qwen3-8B-Base, Full CPT (3B), Random, DCLM, and OPUS. All results use the official benchmark metrics.}
  \label{fig:cpt_domain_bars}
\end{figure}

\subsection{Continued Pre-training}
\label{sec:cpt}
% \begin{wrapfigure}{r}{0.4\linewidth}
%   \centering
%   \vspace{-40pt}
%   \includegraphics[width=\linewidth]{figs/cpt.pdf}
%   \vspace{-25pt}
%   \caption{\textbf{Continued pre-training results on SciencePedia.}}
%   \vspace{-20pt}
%   \label{fig:opus_cpt}
% \end{wrapfigure}
We extend our evaluation to  continued pre-training (CPT), a critical setting for adapting general-purpose LLMs to specialized verticals. We continue training Qwen3-8B-Base on SciencePedia. 
Figure~\ref{fig:opus_cpt} reports the average downstream performance on the specialized SciAssess benchmark and the reasoning-heavy OlympicArena versus CPT tokens.  Notably, OPUS reaches the best performance using only 0.5B tokens and already outperforms random CPT trained for 3B tokens, implying a 6$\times$ gain in data efficiency. 
\begin{wrapfigure}{r}{0.4\linewidth}
  \centering
  \vspace{-1pt}
  \includegraphics[width=\linewidth]{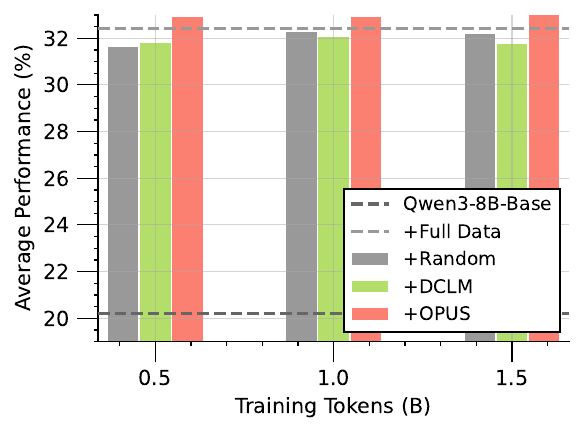}
  \vspace{-25pt}
  \caption{\textbf{Continued pre-training results on SciencePedia.}}
  \vspace{-10pt}
  \label{fig:opus_cpt}
\end{wrapfigure}

\noindent \textbf{Detailed domain-wise CPT Results.}
Figure~\ref{fig:cpt_domain_bars} reports domain breakdowns for continued pre-training on SciencePedia across three token budgets 0.5B, 1B, and 1.5B.
Across OlympicArena (Fig.~\ref{fig:cpt_domain_bars}a) OPUS consistently improves over the base Qwen3-8B-Base and the compute-matched Random baseline in most scientific domains like physics, chemistry, biology, and geography, as well as the text-only and multimodal subsets., with gains that are broadly distributed rather than concentrated in a single category. 
Importantly, OPUS is competitive with, and sometimes surpasses, DCLM and even the Full CPT reference despite using at most 1.5B update tokens, indicating strong data efficiency.
On SciAssess (Fig.~\ref{fig:cpt_domain_bars}b), OPUS yields substantial gains on the material and medicine subsets and ties the best baseline on chemistry, leading to the highest average overall, again with at most 1.5B update tokens.

\subsection{Ablation Study}
\label{sec:ablation}

\noindent \textbf{Soft sampling vs.\ greedy top-$k$.}
We replace Boltzmann soft sampling with a deterministic greedy variant that always selects the top-$K$ candidates by utility.
Table~\ref{tab:ablation_sampling_fw_xl_muon} shows that greedy selection improves over Random, but remains notably behind full OPUS: the greedy variant reaches an Avg.\ of 40.49, whereas OPUS achieves 41.75. This supports our motivation that purely greedy top-$k$ selection can over-concentrate on a narrow set of high-score but overlapping candidates, while stochastic sampling better preserves update diversity and stabilizes training.

\begin{table}[tb!]
\caption{Ablation study on sampling  and validation strategy.}
\label{tab:ablation_sampling_fw_xl_muon}
\centering
\small
\setlength{\tabcolsep}{10pt}
\resizebox{0.7\columnwidth}{!}{%
\begin{tabular}{@{}l|c|cc>{\columncolor{cyan!5}}c}
\toprule
& & \multicolumn{3}{c}{\textbf{OPUS Variants}} \\
\textbf{Benchmark} & \textbf{Random} & \textbf{Greedy} & \textbf{Std. proxy} & \textbf{OPUS} \\ 
\midrule
MMLU       & 28.73 & 29.63 & 29.50 & \textbf{29.89} \\
ANLI       & 33.98 & 33.52 & 33.70 & \textbf{33.29} \\
HellaSwag  & 48.01 & 48.17 & 48.18 & \textbf{48.39} \\
PIQA       & 70.46 & \textbf{72.25} & 71.60 & 71.27 \\
SIQA       & 39.61 & \textbf{41.61} & 40.28 & 41.10 \\
Winogrande & 47.91 & 49.88 & \textbf{51.85} & 47.99 \\
ARC-E      & 38.98 & 37.39 & 38.80 & \textbf{39.68} \\
ARC-C      & 25.42 & 24.75 & 26.10 & \textbf{26.44} \\
C.QA       & 33.25 & 31.12 & \textbf{32.76} & 31.37 \\
WSC        & 36.54 & 36.54 & 37.50 & \textbf{48.08} \\
\midrule
Average    & 40.29 & 40.49 & 41.03 & \textbf{41.75} \\
\bottomrule
\end{tabular}%
}
\end{table}

\noindent \textbf{Benchmark-matched proxy vs.\ standard proxy.}
OPUS estimates the target update direction using a small proxy pool. We compare the default proxy construction with a benchmark-matched proxy that is retrieved to better reflect the downstream evaluation distribution (Sec.~\ref{sec:method}).
As shown in Table~\ref{tab:ablation_sampling_fw_xl_muon}, the benchmark-matched proxy yields a measurable improvement over the default setting, increasing the average from 41.03 to 41.75. This indicates that sharpening the proxy direction can further increase the effectiveness of utility-based selection.
Table~\ref{tab:ablation_sampling_fw_xl_muon} also shows that the standard proxy already provides strong gains over \textsc{Random}, improving the average from 40.29 to 41.03.

\begin{table}[htbp]
\caption{FineWeb results after 30B update tokens for GPT-2 Large pre-trained on FineWeb with the Muon optimizer under varying buffer size $b_t$, temperature $\tau$ and CountSketch projection dimension $m$. See sampling and validation strategy ablations at Table~\ref{tab:ablation_sampling_fw_xl_muon}.}
\resizebox{.99\textwidth}{!}{
\begin{tabular}{@{}lcccccccccc|c}
\toprule
Method & MMLU & ANLI & HellaSwag & PIQA & SIQA & W.G. & ARC-E & ARC-C & C.QA & WSC & Avg. \\
\midrule
% -------------------------------
% GPT-2 Large + AdamW  (KEPT)
% -------------------------------
\multicolumn{12}{c}{\textit{GPT-2 Large with Muon optimizer} ($\tau = 0.9$ $m = 8192$)} \\ \midrule
Random & 28.46 & 32.93 & 42.71 & 69.70 & 40.07 & 49.17 & 37.57 & 28.14 & 31.94 & 36.54 & \textbf{39.72} \\
\midrule
\multicolumn{12}{c}{\textit{GPT-2 Large with Muon optimizer on different buffer size $b_t$} ($\tau = 0.9$ $d = 8192$)} \\
\midrule
OPUS (Buffer size 16) & 28.37 & 33.30 & 42.60 & 69.53 & 40.02 & 48.78 & 38.45 & 27.46 & 32.51 & 36.54 & \textbf{39.76} \\
OPUS (Buffer size 32) & 29.23 & 33.36 & 42.76 & 70.4 & 39.30 & 49.72 & 37.39 & 25.42 & 33.42 & 36.54 & \textbf{39.75} \\
\rowcolor{cyan!5} OPUS (Buffer size 64) & 28.76 & 33.12 & 42.92 & 69.97 & 39.56 & 50.43 & 38.98 & 29.15 & 33.09 & 36.54 & \textbf{40.25} \\
\midrule
\multicolumn{12}{c}{\textit{GPT-2 Large with Muon optimizer on different temperature $\tau$} ($b_t=64$ $m = 8192$)} \\
\midrule
OPUS (temperature 0.8) & 28.54 & 34.19 & 42.92 & 69.59 & 40.23 & 49.33 & 37.92 & 26.78 & 32.76 & 36.54 & \textbf{39.88} \\
OPUS (temperature 1.0) & 28.62 & 33.64 & 43.63 & 70.46 & 39.97 & 50.12 & 37.21 & 24.41 & 32.19 & 38.46 & \textbf{39.87} \\
\rowcolor{cyan!5} OPUS (temperature 0.9) & 28.76 & 33.12 & 42.92 & 69.97 & 39.56 & 50.43 & 38.98 & 29.15 & 33.09 & 36.54 & \textbf{40.25} \\
\midrule
\multicolumn{12}{c}{\textit{GPT-2 Large with Muon optimizer on different CountSketch projection dimension $m$} ($b_t=64$ $\tau = 0.9$)} \\
\midrule
OPUS (projection dimension 4096) & 28.57 & 33.46 & 42.75 & 68.39 & 40.79 & 48.46 & 38.27 & 26.10 & 33.01 & 36.54 & \textbf{39.63} \\
OPUS (projection dimension 16384) & 28.31 & 33.47 & 42.64 & 70.02 & 40.33 & 49.57 & 36.68 & 22.71 & 32.19 & 37.50 & \textbf{39.34} \\
\rowcolor{cyan!5} OPUS (projection dimension 8192) & 28.76 & 33.12 & 42.92 & 69.97 & 39.56 & 50.43 & 38.98 & 29.15 & 33.09 & 36.54 & \textbf{40.25} \\
\bottomrule
\end{tabular}
}
\label{tab:fineweb_ablation}
\end{table}

\noindent \textbf{Hyperparameter sensitivity analysis.}
We conduct further ablation studies on key hyperparameters of OPUS, including (i) the candidate buffer size $b_t$, (ii) the Boltzmann sampling temperature $\tau$, and (iii) the CountSketch projection dimension $m$ (Table~\ref{tab:fineweb_ablation}). Overall, OPUS is reasonably stable across the tested settings and improves over random selection in most configurations. Increasing the buffer size tends to help, with $b_t{=}64$ yielding the best average performance among the evaluated choices. For stochastic selection, a moderate temperature offers a better exploration--exploitation trade-off: $\tau{=}0.9$ performs best compared to both a lower temperature (more greedy) and a higher temperature (closer to uniform sampling). For random projection, we observe sensitivity to the sketch dimension: $m{=}8192$ provides the strongest results among the tested dimensions. Based on these results, we adopt $b_t{=}64$, $\tau{=}0.9$, and $m{=}8192$ as our default configuration.

\begin{figure}[tb!]
    \centering
 \vspace{-10pt} 
    \includegraphics[width=0.99\linewidth]{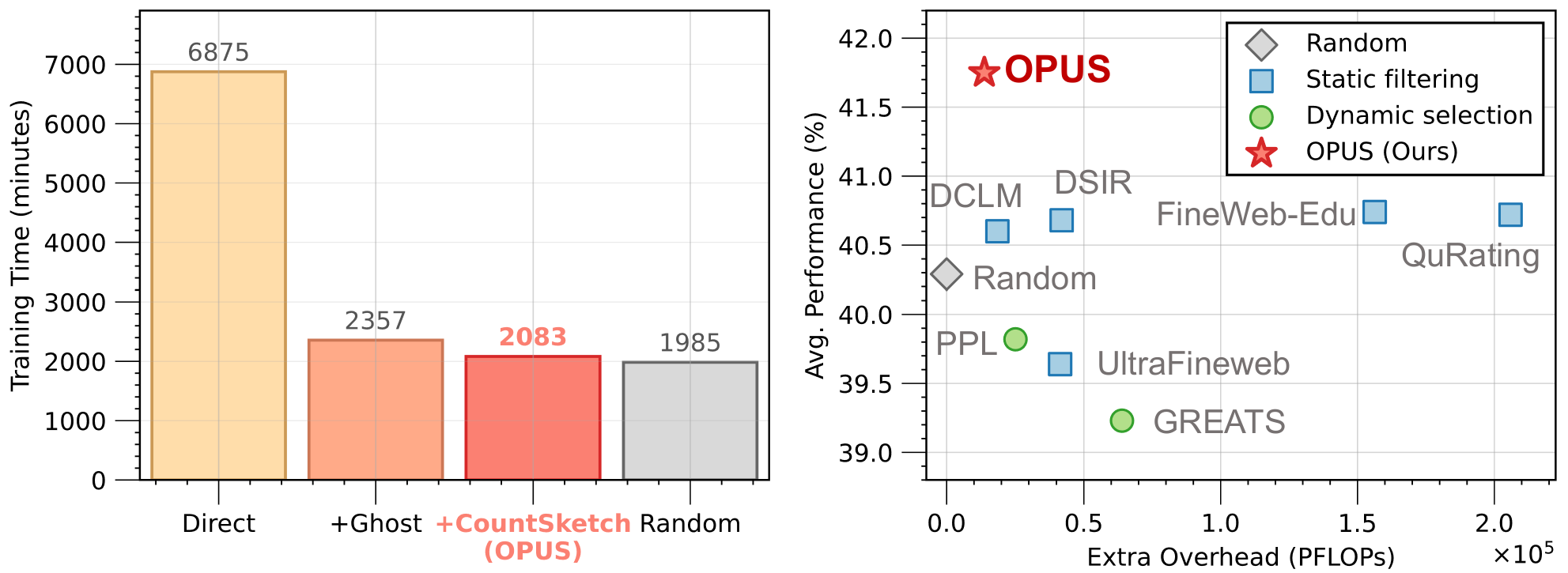}
 \vspace{-10pt} 
    \caption{\textbf{Efficiency and computational cost analysis.} Time (minutes) and total compute (PFLOPs) are evaluated on GPT-2 XL after pre-training on FineWeb (30B tokens) with Muon.}
 \vspace{-10pt} 
    \label{fig:efficiency}
\end{figure}

\subsection{Efficiency Analysis}
\label{sec:efficiency}

A key advantage of OPUS is its minimal computational overhead. Static filtering methods incur a substantial one-time cost to score the entire corpus, while dynamic selection adds per-iteration scoring during training. As shown in Figure~\ref{fig:efficiency}, a naïve direct implementation of online selection would incur \textit{over 3.5$\times$ slowdown} compared to random sampling. By incorporating ghost gradients and CountSketch projections, OPUS reduces this overhead to \textit{only 4.7\%} while achieving the best benchmark performance. In contrast, static methods like QuRating require more compute for selection yet fail to outperform OPUS.

\subsection{Qualitative comparison of selected samples.} We show the selection from a single candidate buffer of size $N{=}32$ and selected $K{=}16$ samples. For each method, we show selected candidates and not selected samples, candidate index, and the method’s raw score (see Appendix~\ref{app:qual}). Overall, OPUS tends to select a more diverse mixture of documents, covering both instructional content and broader web text, rather than concentrating on a narrow ``educational-only'' slice. In contrast, several static filtering method exhibit more extreme preferences—either strongly favoring highly low-diversity patterns or focusing on a limited subset of high-loss samples. These examples support our empirical findings: OPUS’s optimizer-aware utility and stochastic sampling encourage selections that remain broadly suitable for general-purpose pre-training, while still being guided towards high quality samples that align with the proxy direction.

\section{Conclusion and Future work}
We introduced OPUS, a dynamic data selection framework for LLM pre-training that aligns training-time selection with the optimizer’s effective update geometry. Across model scales, optimizers, and corpus quality settings, OPUS consistently improves compute-matched pre-training, suggesting that selection can be substantially strengthened by accounting for how the optimizer actually moves parameters. A natural next step is to extend this optimizer-aligned idea to richer training regimes, such as data mixtures.

\section*{Acknowledgements}
We thank Jiachen T. Wang at Princeton University and Meng Ding at University at Buffalo for helpful feedback and discussions.

\clearpage

\bibliography{colm2026_conference}
\bibliographystyle{colm2026_conference}

%%%%%%%%%%%%%%%%%%%%%%%%%%%%%%%%%%%%%%%%%%%%%%%%%%%%%%%%%%%%%%%%%%%%%%%%%%%%%%%
%%%%%%%%%%%%%%%%%%%%%%%%%%%%%%%%%%%%%%%%%%%%%%%%%%%%%%%%%%%%%%%%%%%%%%%%%%%%%%%
% APPENDIX
%%%%%%%%%%%%%%%%%%%%%%%%%%%%%%%%%%%%%%%%%%%%%%%%%%%%%%%%%%%%%%%%%%%%%%%%%%%%%%%
%%%%%%%%%%%%%%%%%%%%%%%%%%%%%%%%%%%%%%%%%%%%%%%%%%%%%%%%%%%%%%%%%%%%%%%%%%%%%%%
\newpage
\appendix
\onecolumn

\clearpage

\vspace{-15pt}
\section{Qualitative Results}
\label{app:qual}

% Auto-generated qualitative cards (no tables)
% Requires:
% \usepackage{xcolor} % already in icml
\setlength{\fboxsep}{1pt}
\vspace{-13pt}

\noindent\phantomsection
\label{app:qual_random}
\noindent \textbf{Random} \\[6pt]
\noindent
\begin{minipage}[t]{0.49\linewidth}
\fcolorbox{black!25}{gray!6}{\parbox{\dimexpr\linewidth-2\fboxsep-2\fboxrule\relax}{\scriptsize
\textbf{Sample 1}\hfill \textcolor{green!45!black}{\textbf{Selected}}\\
\textbf{Candidate \#0}\hfill \texttt{score=--}\\
\vspace{1pt}\\
\color{black!85} As it turns out, the exercises synonymous with strong, attractive abs may not be the best way to train your core—and may be doing damage to your back. Read more If you are worried about the excess holiday pounds many of us are still carrying around. There are a few easy, natural things you can do to shed them, and none of them requires an\dots}}
\end{minipage}\hfill
\begin{minipage}[t]{0.49\linewidth}
\fcolorbox{black!25}{gray!6}{\parbox{\dimexpr\linewidth-2\fboxsep-2\fboxrule\relax}{\scriptsize
\textbf{Sample 2}\hfill \textcolor{green!45!black}{\textbf{Selected}}\\
\textbf{Candidate \#1}\hfill \texttt{score=--}\\
\vspace{1pt}\\
\color{black!85} Wedding \& Party Venues - Sort By: Edgartown : (508) 627-9510 A 19th century gothic revival home transformed into the island's premier eco-boutique hotel. Guests either stay in the 17-room Hob Knob hotel or in the privacy of their own Hob Knob House. Guests can expect individualized Hob Knob hospitality and modern luxury amenities in a rel\dots}}
\end{minipage}
\vspace{3pt}
\noindent
\begin{minipage}[t]{0.49\linewidth}
\fcolorbox{black!25}{gray!6}{\parbox{\dimexpr\linewidth-2\fboxsep-2\fboxrule\relax}{\scriptsize
\textbf{Sample 3}\hfill \textcolor{green!45!black}{\textbf{Selected}}\\
\textbf{Candidate \#2}\hfill \texttt{score=--}\\
\vspace{1pt}\\
\color{black!85} With the advent of new technologies for sneakers such as Vac Tech, Hyperfuse and Flyknit, the mid 90s and early 2000s methods of production and designing are becoming obsolete in this sneaker world. Nike Running is the future for Nike, generating billions of dollars per year, and we see Nike also not afraid to experiment with technology s\dots}}
\end{minipage}\hfill
\begin{minipage}[t]{0.49\linewidth}
\fcolorbox{black!25}{gray!6}{\parbox{\dimexpr\linewidth-2\fboxsep-2\fboxrule\relax}{\scriptsize
\textbf{Sample 4}\hfill \textcolor{green!45!black}{\textbf{Selected}}\\
\textbf{Candidate \#3}\hfill \texttt{score=--}\\
\vspace{1pt}\\
\color{black!85} starring John Travolta and Sam Jackson The first thing to understand about Basic --the basic thing, let's say-- is that although the commercials make it look like a war movie, it is not, for which we can all be grateful. No, Basic is a plot-twisty whodunnit. If The Usual Suspects died, and its body turned to cheese, and then that cheese-b\dots}}
\end{minipage}
\vspace{3pt}
\noindent
\begin{minipage}[t]{0.49\linewidth}
\fcolorbox{black!25}{gray!6}{\parbox{\dimexpr\linewidth-2\fboxsep-2\fboxrule\relax}{\scriptsize
\textbf{Sample 5}\hfill \textcolor{green!45!black}{\textbf{Selected}}\\
\textbf{Candidate \#4}\hfill \texttt{score=--}\\
\vspace{1pt}\\
\color{black!85} 5 Types of Women’s Underwear That Men Love Underwear can say a lot about a woman. It’s something that men are obsessed with, to the point that, a mere glimpse of a thong waistband causes us to go into shock. On the surface we find them sexy, revealing. We’re able to see who a woman actually is—or maybe some guys are just plain horny. Howe\dots}}
\end{minipage}\hfill
\begin{minipage}[t]{0.49\linewidth}
\fcolorbox{black!25}{gray!6}{\parbox{\dimexpr\linewidth-2\fboxsep-2\fboxrule\relax}{\scriptsize
\textbf{Sample 6}\hfill \textcolor{green!45!black}{\textbf{Selected}}\\
\textbf{Candidate \#7}\hfill \texttt{score=--}\\
\vspace{1pt}\\
\color{black!85} Elizabeth Hurley played as Dalila Release: Dec 8, 1996 Mara and her husband Manoa are both upstanding and religious Israelites living under the harsh and unjust rule of the Philistines. Much to their regret, they have not been able to have children. One day, a mysterious stranger appears to Mara and promises her that she will bear a son w\dots}}
\end{minipage}
\vspace{3pt}
\noindent
\begin{minipage}[t]{0.49\linewidth}
\fcolorbox{black!25}{gray!6}{\parbox{\dimexpr\linewidth-2\fboxsep-2\fboxrule\relax}{\scriptsize
\textbf{Sample 7}\hfill \textcolor{green!45!black}{\textbf{Selected}}\\
\textbf{Candidate \#8}\hfill \texttt{score=--}\\
\vspace{1pt}\\
\color{black!85} The Unsung Heroes of Your HVAC System: Understanding the Importance of Filters When it comes to your HVAC (Heating, Ventilation, and Air Conditioning) system, you might be quick to think about the thermostat, air ducts, or even the unit itself. However, there’s an unsung hero in your HVAC system that plays a pivotal role in maintaining in\dots}}
\end{minipage}\hfill
\begin{minipage}[t]{0.49\linewidth}
\fcolorbox{black!25}{gray!6}{\parbox{\dimexpr\linewidth-2\fboxsep-2\fboxrule\relax}{\scriptsize
\textbf{Sample 8}\hfill \textcolor{green!45!black}{\textbf{Selected}}\\
\textbf{Candidate \#13}\hfill \texttt{score=--}\\
\vspace{1pt}\\
\color{black!85} In Heart of Darkness it is the white invaders for instance, who are, almost without exception, embodiments of blindness, selfishness, and cruelty; and even in the cognitive domain, where such positive phrases as “to enlighten,” for instance, are conventionally opposed to negative ones such as “to be in the dark,” the traditional expectati\dots}}
\end{minipage}
\vspace{3pt}
\noindent
\begin{minipage}[t]{0.49\linewidth}
\fcolorbox{black!25}{gray!6}{\parbox{\dimexpr\linewidth-2\fboxsep-2\fboxrule\relax}{\scriptsize
\textbf{Sample 9}\hfill \textcolor{green!45!black}{\textbf{Selected}}\\
\textbf{Candidate \#17}\hfill \texttt{score=--}\\
\vspace{1pt}\\
\color{black!85} Political Parties and Elections Political parties are an established part of modern mass democracy, and the conduct of elections in India is largely dependent on the behaviour of political parties. Although many candidates for Indian elections are independent, the winning candidates for Lok Sabha and Vidhan Sabha elections usually stand a\dots}}
\end{minipage}\hfill
\begin{minipage}[t]{0.49\linewidth}
\fcolorbox{black!25}{gray!6}{\parbox{\dimexpr\linewidth-2\fboxsep-2\fboxrule\relax}{\scriptsize
\textbf{Sample 10}\hfill \textcolor{green!45!black}{\textbf{Selected}}\\
\textbf{Candidate \#18}\hfill \texttt{score=--}\\
\vspace{1pt}\\
\color{black!85} This article originally appeared in the December 2015 issue of Resource Recycling. Subscribe today for access to all print content. Since the 1990s, curbside and drop-off recycling has grown substantially – nearly 90 percent of households now have access, according to recent surveys from Moore Recycling Associates, the American Forest and\dots}}
\end{minipage}
\vspace{3pt}
\noindent
\begin{minipage}[t]{0.49\linewidth}
\fcolorbox{black!25}{gray!6}{\parbox{\dimexpr\linewidth-2\fboxsep-2\fboxrule\relax}{\scriptsize
\textbf{Sample 11}\hfill \textcolor{green!45!black}{\textbf{Selected}}\\
\textbf{Candidate \#21}\hfill \texttt{score=--}\\
\vspace{1pt}\\
\color{black!85} Nestled in the leafy suburbs of western Berlin, the Wannsee Conference House stands as a poignant reminder of a dark chapter in human history. The Wannsee Conference: A Pivotal Moment The Wannsee Conference, held on January 20, 1942, marked a pivotal moment in the implementation of Nazi Germany's genocidal plans. Organized by SS-Obergrupp\dots}}
\end{minipage}\hfill
\begin{minipage}[t]{0.49\linewidth}
\fcolorbox{black!25}{gray!6}{\parbox{\dimexpr\linewidth-2\fboxsep-2\fboxrule\relax}{\scriptsize
\textbf{Sample 12}\hfill \textcolor{green!45!black}{\textbf{Selected}}\\
\textbf{Candidate \#23}\hfill \texttt{score=--}\\
\vspace{1pt}\\
\color{black!85} The St. James kindergarteners have been working up to Project Week over the past month. We started slowly by taking walks in our neighborhood while Ms. Meghan and I noted what caught the children’s interest. It became apparent that the class was very interested in the L trains that they saw on our walks. It started with a simple question,\dots}}
\end{minipage}
\vspace{3pt}
\noindent
\begin{minipage}[t]{0.49\linewidth}
\fcolorbox{black!25}{gray!6}{\parbox{\dimexpr\linewidth-2\fboxsep-2\fboxrule\relax}{\scriptsize
\textbf{Sample 13}\hfill \textcolor{green!45!black}{\textbf{Selected}}\\
\textbf{Candidate \#27}\hfill \texttt{score=--}\\
\vspace{1pt}\\
\color{black!85} 24/7 writing help on your phone Save to my list Remove from my list In the tumultuous 19th century, both Italy and Germany found themselves fragmented into numerous separate ruling states. The impetus for change came in the form of rising nationalism and liberalism, paving the way for the unification of these disparate entities. However,\dots}}
\end{minipage}\hfill
\begin{minipage}[t]{0.49\linewidth}
\fcolorbox{black!25}{gray!6}{\parbox{\dimexpr\linewidth-2\fboxsep-2\fboxrule\relax}{\scriptsize
\textbf{Sample 14}\hfill \textcolor{green!45!black}{\textbf{Selected}}\\
\textbf{Candidate \#29}\hfill \texttt{score=--}\\
\vspace{1pt}\\
\color{black!85} Earthquakes are the result of sudden movement along faults within the Earth. The movement releases stored-up ‘elastic strain’ energy in the form of seismic waves, which propagate through the Earth and cause the ground surface to shake. Such movement on the faults is generally a response to long-term deformation and the buildup of stress.\dots}}
\end{minipage}
\vspace{1pt}
\noindent
\begin{minipage}[t]{0.49\linewidth}
\fcolorbox{black!25}{gray!6}{\parbox{\dimexpr\linewidth-2\fboxsep-2\fboxrule\relax}{\scriptsize
\textbf{Sample 15}\hfill \textcolor{green!45!black}{\textbf{Selected}}\\
\textbf{Candidate \#30}\hfill \texttt{score=--}\\
\vspace{1pt}\\
\color{black!85} Over 1.8 million professionals use CFI to learn accounting, financial analysis, modeling and more. Start with a free account to explore 20+ always-free courses and hundreds of finance templates and cheat sheets. What is the Central Limit Theorem (CLT)? The Central Limit Theorem (CLT) is a statistical concept that states that the sample me\dots}}
\end{minipage}\hfill
\begin{minipage}[t]{0.49\linewidth}
\fcolorbox{black!25}{gray!6}{\parbox{\dimexpr\linewidth-2\fboxsep-2\fboxrule\relax}{\scriptsize
\textbf{Sample 16}\hfill \textcolor{green!45!black}{\textbf{Selected}}\\
\textbf{Candidate \#31}\hfill \texttt{score=--}\\
\vspace{1pt}\\
\color{black!85} One of the challenges of working with ancient DNA samples is that damage accumulates over time, breaking the double helix structure into ever-smaller fragments. In the samples we worked with, these fragments were scattered and mixed with contaminants, making genome reconstruction a major technical challenge. But a shocking paper published\dots}}
\end{minipage}
\vspace{3pt}
\clearpage
\noindent\phantomsection
\label{app:qual_random_cont}
% \noindent \textbf{Random} \\[6pt]
\noindent
\begin{minipage}[t]{0.49\linewidth}
\fcolorbox{black!25}{gray!6}{\parbox{\dimexpr\linewidth-2\fboxsep-2\fboxrule\relax}{\scriptsize
\textbf{Sample 17}\hfill \textcolor{black!45}{\textbf{Not selected}}\\
\textbf{Candidate \#5}\hfill \texttt{score=--}\\
\vspace{1pt}\\
\color{black!85} Well this is the big one. So big apparently, that I had to take it there and raise the number from 10 to 15. There’s just that many fails in the world of female rap. Some slight missteps, some EPIC. Nevertheless, they are all worth mentioning. You can probably think of a bunch more, but this is what I have gathered picking up from my prev\dots}}
\end{minipage}\hfill
\begin{minipage}[t]{0.49\linewidth}
\fcolorbox{black!25}{gray!6}{\parbox{\dimexpr\linewidth-2\fboxsep-2\fboxrule\relax}{\scriptsize
\textbf{Sample 18}\hfill \textcolor{black!45}{\textbf{Not selected}}\\
\textbf{Candidate \#6}\hfill \texttt{score=--}\\
\vspace{1pt}\\
\color{black!85} Skaters need to check their skate helmets every so often and ask yourself, "Is it time to replace this helmet?" Well, that depends. Did you crash in it? For starters, most people are aware that you must replace a helmet after any crash where your head hit. The foam part of a helmet is made for one-time use, and after crushing once it is n\dots}}
\end{minipage}
\vspace{3pt}
\noindent
\begin{minipage}[t]{0.49\linewidth}
\fcolorbox{black!25}{gray!6}{\parbox{\dimexpr\linewidth-2\fboxsep-2\fboxrule\relax}{\scriptsize
\textbf{Sample 19}\hfill \textcolor{black!45}{\textbf{Not selected}}\\
\textbf{Candidate \#9}\hfill \texttt{score=--}\\
\vspace{1pt}\\
\color{black!85} “Last night three cargoes of Bohea Tea were emptied into the sea. This is the most magnificent movement of all. There is a dignity, a majesty, a sublimity, in this last effort of the Patriots that I greatly admire.” - John Adams, diary entry, December 17, 1773 - John Adams, diary entry, December 17, 1773 A Novel Idea Is something so new a\dots}}
\end{minipage}\hfill
\begin{minipage}[t]{0.49\linewidth}
\fcolorbox{black!25}{gray!6}{\parbox{\dimexpr\linewidth-2\fboxsep-2\fboxrule\relax}{\scriptsize
\textbf{Sample 20}\hfill \textcolor{black!45}{\textbf{Not selected}}\\
\textbf{Candidate \#10}\hfill \texttt{score=--}\\
\vspace{1pt}\\
\color{black!85} Deforestation isn't just happening in well-known global hotspots like Indonesia and Brazil's rainforest. A new analysis says forests are also shrinking on state and private land in Oregon, where an estimated 522,000 acres of forest cover have disappeared since 2000. That's an area six times larger than the city of Portland, equal to more\dots}}
\end{minipage}
\vspace{3pt}
\noindent
\begin{minipage}[t]{0.49\linewidth}
\fcolorbox{black!25}{gray!6}{\parbox{\dimexpr\linewidth-2\fboxsep-2\fboxrule\relax}{\scriptsize
\textbf{Sample 21}\hfill \textcolor{black!45}{\textbf{Not selected}}\\
\textbf{Candidate \#11}\hfill \texttt{score=--}\\
\vspace{1pt}\\
\color{black!85} In decades past, classroom design was often an afterthought and followed a standardised layout. Plain boxed shaped classrooms, with identical chairs and tables throughout were commonplace in many schools. Read the latest issue of School News HERE Recently, though, there has been a shift away from this one-size-fits all approach to classro\dots}}
\end{minipage}\hfill
\begin{minipage}[t]{0.49\linewidth}
\fcolorbox{black!25}{gray!6}{\parbox{\dimexpr\linewidth-2\fboxsep-2\fboxrule\relax}{\scriptsize
\textbf{Sample 22}\hfill \textcolor{black!45}{\textbf{Not selected}}\\
\textbf{Candidate \#12}\hfill \texttt{score=--}\\
\vspace{1pt}\\
\color{black!85} Can you please give us a little short bio? (education, professional experiences, select publications, academic specialty, awards won) Public school teacher for 5 years BA art (UC Irvine) PhD. (UCLA) educational psychology Professor of Child Development, (25 years) CSUS Senior Research Scientist (Oregon Research Institute with Institute of\dots}}
\end{minipage}
\vspace{3pt}
\noindent
\begin{minipage}[t]{0.49\linewidth}
\fcolorbox{black!25}{gray!6}{\parbox{\dimexpr\linewidth-2\fboxsep-2\fboxrule\relax}{\scriptsize
\textbf{Sample 23}\hfill \textcolor{black!45}{\textbf{Not selected}}\\
\textbf{Candidate \#14}\hfill \texttt{score=--}\\
\vspace{1pt}\\
\color{black!85} Is your major sustainable enough? Whether you’re pursuing a sustainability degree and want to further your knowledge, or are interested in supplementing your major in another area with sustainability education, plenty of independent learning resources are available. A wide range of credit and noncredit courses—including university- and or\dots}}
\end{minipage}\hfill
\begin{minipage}[t]{0.49\linewidth}
\fcolorbox{black!25}{gray!6}{\parbox{\dimexpr\linewidth-2\fboxsep-2\fboxrule\relax}{\scriptsize
\textbf{Sample 24}\hfill \textcolor{black!45}{\textbf{Not selected}}\\
\textbf{Candidate \#15}\hfill \texttt{score=--}\\
\vspace{1pt}\\
\color{black!85} Origami is an art form that combines precision, creativity, and patience. While basic origami is obtainable to every one, mastering complex origami designs can be quite a rewarding and impressive achievement. In this article, we’ll show you with the procedure for creating intricate origami while highlighting essential techniques for achie\dots}}
\end{minipage}
\vspace{3pt}
\noindent
\begin{minipage}[t]{0.49\linewidth}
\fcolorbox{black!25}{gray!6}{\parbox{\dimexpr\linewidth-2\fboxsep-2\fboxrule\relax}{\scriptsize
\textbf{Sample 25}\hfill \textcolor{black!45}{\textbf{Not selected}}\\
\textbf{Candidate \#16}\hfill \texttt{score=--}\\
\vspace{1pt}\\
\color{black!85} What is rotavirus and why does my baby need to be immunised? Rotavirus is a very infectious virus that causes the majority of serious cases of gastroenteritis in babies. It causes diarrhoea, vomiting and abdominal pain, usually lasting around a week. Most children will be infected by rotavirus once by the age of five. Gastroenteritis (cau\dots}}
\end{minipage}\hfill
\begin{minipage}[t]{0.49\linewidth}
\fcolorbox{black!25}{gray!6}{\parbox{\dimexpr\linewidth-2\fboxsep-2\fboxrule\relax}{\scriptsize
\textbf{Sample 26}\hfill \textcolor{black!45}{\textbf{Not selected}}\\
\textbf{Candidate \#19}\hfill \texttt{score=--}\\
\vspace{1pt}\\
\color{black!85} Dividing Fractions Using Models Worksheet. This worksheet has six division with fractions issues to be solved — three must be solved with fashions and three with algorithms — options are on the second page. Answer key divide the unit fractions by whole numbers using th e fashions given. Use these resources to help reinforce the following\dots}}
\end{minipage}
\vspace{3pt}
\noindent
\begin{minipage}[t]{0.49\linewidth}
\fcolorbox{black!25}{gray!6}{\parbox{\dimexpr\linewidth-2\fboxsep-2\fboxrule\relax}{\scriptsize
\textbf{Sample 27}\hfill \textcolor{black!45}{\textbf{Not selected}}\\
\textbf{Candidate \#20}\hfill \texttt{score=--}\\
\vspace{1pt}\\
\color{black!85} Conduct Disorder (CD) is a complex and serious behavioural and emotional disorder that can occur in children and adolescents. It’s characterised by a repetitive and persistent pattern of behaviour where the basic rights of others or major age-appropriate societal norms or rules are violated. Here’s an outline of Conduct Disorder in line w\dots}}
\end{minipage}\hfill
\begin{minipage}[t]{0.49\linewidth}
\fcolorbox{black!25}{gray!6}{\parbox{\dimexpr\linewidth-2\fboxsep-2\fboxrule\relax}{\scriptsize
\textbf{Sample 28}\hfill \textcolor{black!45}{\textbf{Not selected}}\\
\textbf{Candidate \#22}\hfill \texttt{score=--}\\
\vspace{1pt}\\
\color{black!85} How To Choose Decodable Readers for First Grade To decode or not to decode: really, there is no question. To help rising first graders become successful and enthusiastic readers this summer, decodable readers are essential reading resources. Although “decodable text” might sound like yet another form of educational lingo, parents and educ\dots}}
\end{minipage}
\vspace{3pt}
\noindent
\begin{minipage}[t]{0.49\linewidth}
\fcolorbox{black!25}{gray!6}{\parbox{\dimexpr\linewidth-2\fboxsep-2\fboxrule\relax}{\scriptsize
\textbf{Sample 29}\hfill \textcolor{black!45}{\textbf{Not selected}}\\
\textbf{Candidate \#24}\hfill \texttt{score=--}\\
\vspace{1pt}\\
\color{black!85} Next we will talk about solar radiation, that is, the forms of solar radiation that we receive on earth. Solar radiation is generated by a series of nuclear fusion reactions that occur in the Sun and, as a consequence, emit electromagnetic radiation that reaches the earth. This radiation received by the earth’s surface is measured in W /\dots}}
\end{minipage}\hfill
\begin{minipage}[t]{0.49\linewidth}
\fcolorbox{black!25}{gray!6}{\parbox{\dimexpr\linewidth-2\fboxsep-2\fboxrule\relax}{\scriptsize
\textbf{Sample 30}\hfill \textcolor{black!45}{\textbf{Not selected}}\\
\textbf{Candidate \#25}\hfill \texttt{score=--}\\
\vspace{1pt}\\
\color{black!85} KS2 Maths is an important core subject in the National Curriculum and this area of the website covers all the major aspects of the curriculum including numbers, calculations, problems and measures. Each subject area is designed to help children develop their knowledge, whether they are learning in a classroom or home schooling environment\dots}}
\end{minipage}
\vspace{3pt}
\noindent
\begin{minipage}[t]{0.49\linewidth}
\fcolorbox{black!25}{gray!6}{\parbox{\dimexpr\linewidth-2\fboxsep-2\fboxrule\relax}{\scriptsize
\textbf{Sample 31}\hfill \textcolor{black!45}{\textbf{Not selected}}\\
\textbf{Candidate \#26}\hfill \texttt{score=--}\\
\vspace{1pt}\\
\color{black!85} Unveiling the Power: Key Provisions of the Civil Rights Act of 1864 What were the Civil Rights Act of 1864's key provisions? The Civil Rights Act of 1864 was a pivotal moment in American history, establishing crucial legal protections for African Americans in the face of rampant discrimination. Editor Note: The Civil Rights Act of 1864 la\dots}}
\end{minipage}\hfill
\begin{minipage}[t]{0.49\linewidth}
\fcolorbox{black!25}{gray!6}{\parbox{\dimexpr\linewidth-2\fboxsep-2\fboxrule\relax}{\scriptsize
\textbf{Sample 32}\hfill \textcolor{black!45}{\textbf{Not selected}}\\
\textbf{Candidate \#28}\hfill \texttt{score=--}\\
\vspace{1pt}\\
\color{black!85} You really have to be alert when studying science. Galaxies were created after matter. The stars in those galaxies were supposed to move slowly because there was more mass in the center of the galaxy. However, after dark matter was added, the stars appeared to move faster; however, this is not the case in our galaxy, suggesting that there\dots}}
\end{minipage}
\vspace{3pt}

\clearpage
\noindent\phantomsection
\label{app:qual_opus}
\noindent \textbf{OPUS} \\[6pt]
\noindent
\begin{minipage}[t]{0.49\linewidth}
\fcolorbox{black!25}{gray!6}{\parbox{\dimexpr\linewidth-2\fboxsep-2\fboxrule\relax}{\scriptsize
\textbf{Sample 1}\hfill \textcolor{green!45!black}{\textbf{Selected}}\\
\textbf{Candidate \#8}\hfill \texttt{score=0.00589}\\
\vspace{1pt}\\
\color{black!85} The Unsung Heroes of Your HVAC System: Understanding the Importance of Filters When it comes to your HVAC (Heating, Ventilation, and Air Conditioning) system, you might be quick to think about the thermostat, air ducts, or even the unit itself. However, there’s an unsung hero in your HVAC system that plays a pivotal role in maintaining in\dots}}
\end{minipage}\hfill
\begin{minipage}[t]{0.49\linewidth}
\fcolorbox{black!25}{gray!6}{\parbox{\dimexpr\linewidth-2\fboxsep-2\fboxrule\relax}{\scriptsize
\textbf{Sample 2}\hfill \textcolor{green!45!black}{\textbf{Selected}}\\
\textbf{Candidate \#22}\hfill \texttt{score=0.00471}\\
\vspace{1pt}\\
\color{black!85} How To Choose Decodable Readers for First Grade To decode or not to decode: really, there is no question. To help rising first graders become successful and enthusiastic readers this summer, decodable readers are essential reading resources. Although “decodable text” might sound like yet another form of educational lingo, parents and educ\dots}}
\end{minipage}
\vspace{3pt}
\noindent
\begin{minipage}[t]{0.49\linewidth}
\fcolorbox{black!25}{gray!6}{\parbox{\dimexpr\linewidth-2\fboxsep-2\fboxrule\relax}{\scriptsize
\textbf{Sample 3}\hfill \textcolor{green!45!black}{\textbf{Selected}}\\
\textbf{Candidate \#27}\hfill \texttt{score=0.00466}\\
\vspace{1pt}\\
\color{black!85} 24/7 writing help on your phone Save to my list Remove from my list In the tumultuous 19th century, both Italy and Germany found themselves fragmented into numerous separate ruling states. The impetus for change came in the form of rising nationalism and liberalism, paving the way for the unification of these disparate entities. However,\dots}}
\end{minipage}\hfill
\begin{minipage}[t]{0.49\linewidth}
\fcolorbox{black!25}{gray!6}{\parbox{\dimexpr\linewidth-2\fboxsep-2\fboxrule\relax}{\scriptsize
\textbf{Sample 4}\hfill \textcolor{green!45!black}{\textbf{Selected}}\\
\textbf{Candidate \#4}\hfill \texttt{score=0.0046}\\
\vspace{1pt}\\
\color{black!85} 5 Types of Women’s Underwear That Men Love Underwear can say a lot about a woman. It’s something that men are obsessed with, to the point that, a mere glimpse of a thong waistband causes us to go into shock. On the surface we find them sexy, revealing. We’re able to see who a woman actually is—or maybe some guys are just plain horny. Howe\dots}}
\end{minipage}
\vspace{3pt}
\noindent
\begin{minipage}[t]{0.49\linewidth}
\fcolorbox{black!25}{gray!6}{\parbox{\dimexpr\linewidth-2\fboxsep-2\fboxrule\relax}{\scriptsize
\textbf{Sample 5}\hfill \textcolor{green!45!black}{\textbf{Selected}}\\
\textbf{Candidate \#18}\hfill \texttt{score=0.0044}\\
\vspace{1pt}\\
\color{black!85} This article originally appeared in the December 2015 issue of Resource Recycling. Subscribe today for access to all print content. Since the 1990s, curbside and drop-off recycling has grown substantially – nearly 90 percent of households now have access, according to recent surveys from Moore Recycling Associates, the American Forest and\dots}}
\end{minipage}\hfill
\begin{minipage}[t]{0.49\linewidth}
\fcolorbox{black!25}{gray!6}{\parbox{\dimexpr\linewidth-2\fboxsep-2\fboxrule\relax}{\scriptsize
\textbf{Sample 6}\hfill \textcolor{green!45!black}{\textbf{Selected}}\\
\textbf{Candidate \#30}\hfill \texttt{score=0.0042}\\
\vspace{1pt}\\
\color{black!85} Over 1.8 million professionals use CFI to learn accounting, financial analysis, modeling and more. Start with a free account to explore 20+ always-free courses and hundreds of finance templates and cheat sheets. What is the Central Limit Theorem (CLT)? The Central Limit Theorem (CLT) is a statistical concept that states that the sample me\dots}}
\end{minipage}
\vspace{3pt}
\noindent
\begin{minipage}[t]{0.49\linewidth}
\fcolorbox{black!25}{gray!6}{\parbox{\dimexpr\linewidth-2\fboxsep-2\fboxrule\relax}{\scriptsize
\textbf{Sample 7}\hfill \textcolor{green!45!black}{\textbf{Selected}}\\
\textbf{Candidate \#0}\hfill \texttt{score=0.0042}\\
\vspace{1pt}\\
\color{black!85} As it turns out, the exercises synonymous with strong, attractive abs may not be the best way to train your core—and may be doing damage to your back. Read more If you are worried about the excess holiday pounds many of us are still carrying around. There are a few easy, natural things you can do to shed them, and none of them requires an\dots}}
\end{minipage}\hfill
\begin{minipage}[t]{0.49\linewidth}
\fcolorbox{black!25}{gray!6}{\parbox{\dimexpr\linewidth-2\fboxsep-2\fboxrule\relax}{\scriptsize
\textbf{Sample 8}\hfill \textcolor{green!45!black}{\textbf{Selected}}\\
\textbf{Candidate \#23}\hfill \texttt{score=0.00418}\\
\vspace{1pt}\\
\color{black!85} The St. James kindergarteners have been working up to Project Week over the past month. We started slowly by taking walks in our neighborhood while Ms. Meghan and I noted what caught the children’s interest. It became apparent that the class was very interested in the L trains that they saw on our walks. It started with a simple question,\dots}}
\end{minipage}
\vspace{3pt}
\noindent
\begin{minipage}[t]{0.49\linewidth}
\fcolorbox{black!25}{gray!6}{\parbox{\dimexpr\linewidth-2\fboxsep-2\fboxrule\relax}{\scriptsize
\textbf{Sample 9}\hfill \textcolor{green!45!black}{\textbf{Selected}}\\
\textbf{Candidate \#31}\hfill \texttt{score=0.00411}\\
\vspace{1pt}\\
\color{black!85} One of the challenges of working with ancient DNA samples is that damage accumulates over time, breaking the double helix structure into ever-smaller fragments. In the samples we worked with, these fragments were scattered and mixed with contaminants, making genome reconstruction a major technical challenge. But a shocking paper published\dots}}
\end{minipage}\hfill
\begin{minipage}[t]{0.49\linewidth}
\fcolorbox{black!25}{gray!6}{\parbox{\dimexpr\linewidth-2\fboxsep-2\fboxrule\relax}{\scriptsize
\textbf{Sample 10}\hfill \textcolor{green!45!black}{\textbf{Selected}}\\
\textbf{Candidate \#11}\hfill \texttt{score=0.00401}\\
\vspace{1pt}\\
\color{black!85} In decades past, classroom design was often an afterthought and followed a standardised layout. Plain boxed shaped classrooms, with identical chairs and tables throughout were commonplace in many schools. Read the latest issue of School News HERE Recently, though, there has been a shift away from this one-size-fits all approach to classro\dots}}
\end{minipage}
\vspace{3pt}
\noindent
\begin{minipage}[t]{0.49\linewidth}
\fcolorbox{black!25}{gray!6}{\parbox{\dimexpr\linewidth-2\fboxsep-2\fboxrule\relax}{\scriptsize
\textbf{Sample 11}\hfill \textcolor{green!45!black}{\textbf{Selected}}\\
\textbf{Candidate \#25}\hfill \texttt{score=0.00396}\\
\vspace{1pt}\\
\color{black!85} KS2 Maths is an important core subject in the National Curriculum and this area of the website covers all the major aspects of the curriculum including numbers, calculations, problems and measures. Each subject area is designed to help children develop their knowledge, whether they are learning in a classroom or home schooling environment\dots}}
\end{minipage}\hfill
\begin{minipage}[t]{0.49\linewidth}
\fcolorbox{black!25}{gray!6}{\parbox{\dimexpr\linewidth-2\fboxsep-2\fboxrule\relax}{\scriptsize
\textbf{Sample 12}\hfill \textcolor{green!45!black}{\textbf{Selected}}\\
\textbf{Candidate \#7}\hfill \texttt{score=0.0039}\\
\vspace{1pt}\\
\color{black!85} Elizabeth Hurley played as Dalila Release: Dec 8, 1996 Mara and her husband Manoa are both upstanding and religious Israelites living under the harsh and unjust rule of the Philistines. Much to their regret, they have not been able to have children. One day, a mysterious stranger appears to Mara and promises her that she will bear a son w\dots}}
\end{minipage}
\vspace{3pt}
\noindent
\begin{minipage}[t]{0.49\linewidth}
\fcolorbox{black!25}{gray!6}{\parbox{\dimexpr\linewidth-2\fboxsep-2\fboxrule\relax}{\scriptsize
\textbf{Sample 13}\hfill \textcolor{green!45!black}{\textbf{Selected}}\\
\textbf{Candidate \#5}\hfill \texttt{score=0.00389}\\
\vspace{1pt}\\
\color{black!85} Well this is the big one. So big apparently, that I had to take it there and raise the number from 10 to 15. There’s just that many fails in the world of female rap. Some slight missteps, some EPIC. Nevertheless, they are all worth mentioning. You can probably think of a bunch more, but this is what I have gathered picking up from my prev\dots}}
\end{minipage}\hfill
\begin{minipage}[t]{0.49\linewidth}
\fcolorbox{black!25}{gray!6}{\parbox{\dimexpr\linewidth-2\fboxsep-2\fboxrule\relax}{\scriptsize
\textbf{Sample 14}\hfill \textcolor{green!45!black}{\textbf{Selected}}\\
\textbf{Candidate \#9}\hfill \texttt{score=0.00384}\\
\vspace{1pt}\\
\color{black!85} “Last night three cargoes of Bohea Tea were emptied into the sea. This is the most magnificent movement of all. There is a dignity, a majesty, a sublimity, in this last effort of the Patriots that I greatly admire.” - John Adams, diary entry, December 17, 1773 - John Adams, diary entry, December 17, 1773 A Novel Idea Is something so new a\dots}}
\end{minipage}
\vspace{3pt}
\noindent
\begin{minipage}[t]{0.49\linewidth}
\fcolorbox{black!25}{gray!6}{\parbox{\dimexpr\linewidth-2\fboxsep-2\fboxrule\relax}{\scriptsize
\textbf{Sample 15}\hfill \textcolor{green!45!black}{\textbf{Selected}}\\
\textbf{Candidate \#19}\hfill \texttt{score=0.00376}\\
\vspace{1pt}\\
\color{black!85} Dividing Fractions Using Models Worksheet. This worksheet has six division with fractions issues to be solved — three must be solved with fashions and three with algorithms — options are on the second page. Answer key divide the unit fractions by whole numbers using th e fashions given. Use these resources to help reinforce the following\dots}}
\end{minipage}\hfill
\begin{minipage}[t]{0.49\linewidth}
\fcolorbox{black!25}{gray!6}{\parbox{\dimexpr\linewidth-2\fboxsep-2\fboxrule\relax}{\scriptsize
\textbf{Sample 16}\hfill \textcolor{green!45!black}{\textbf{Selected}}\\
\textbf{Candidate \#1}\hfill \texttt{score=0.00348}\\
\vspace{1pt}\\
\color{black!85} Wedding \& Party Venues - Sort By: Edgartown : (508) 627-9510 A 19th century gothic revival home transformed into the island's premier eco-boutique hotel. Guests either stay in the 17-room Hob Knob hotel or in the privacy of their own Hob Knob House. Guests can expect individualized Hob Knob hospitality and modern luxury amenities in a rel\dots}}
\end{minipage}
\vspace{3pt}
\clearpage
\noindent\phantomsection
\label{app:qual_opus_cont}
% \noindent \textbf{OPUS} \\[6pt]
\noindent
\begin{minipage}[t]{0.49\linewidth}
\fcolorbox{black!25}{gray!6}{\parbox{\dimexpr\linewidth-2\fboxsep-2\fboxrule\relax}{\scriptsize
\textbf{Sample 17}\hfill \textcolor{black!45}{\textbf{Not selected}}\\
\textbf{Candidate \#15}\hfill \texttt{score=0.00524}\\
\vspace{1pt}\\
\color{black!85} Origami is an art form that combines precision, creativity, and patience. While basic origami is obtainable to every one, mastering complex origami designs can be quite a rewarding and impressive achievement. In this article, we’ll show you with the procedure for creating intricate origami while highlighting essential techniques for achie\dots}}
\end{minipage}\hfill
\begin{minipage}[t]{0.49\linewidth}
\fcolorbox{black!25}{gray!6}{\parbox{\dimexpr\linewidth-2\fboxsep-2\fboxrule\relax}{\scriptsize
\textbf{Sample 18}\hfill \textcolor{black!45}{\textbf{Not selected}}\\
\textbf{Candidate \#20}\hfill \texttt{score=0.00518}\\
\vspace{1pt}\\
\color{black!85} Conduct Disorder (CD) is a complex and serious behavioural and emotional disorder that can occur in children and adolescents. It’s characterised by a repetitive and persistent pattern of behaviour where the basic rights of others or major age-appropriate societal norms or rules are violated. Here’s an outline of Conduct Disorder in line w\dots}}
\end{minipage}
\vspace{3pt}
\noindent
\begin{minipage}[t]{0.49\linewidth}
\fcolorbox{black!25}{gray!6}{\parbox{\dimexpr\linewidth-2\fboxsep-2\fboxrule\relax}{\scriptsize
\textbf{Sample 19}\hfill \textcolor{black!45}{\textbf{Not selected}}\\
\textbf{Candidate \#14}\hfill \texttt{score=0.00472}\\
\vspace{1pt}\\
\color{black!85} Is your major sustainable enough? Whether you’re pursuing a sustainability degree and want to further your knowledge, or are interested in supplementing your major in another area with sustainability education, plenty of independent learning resources are available. A wide range of credit and noncredit courses—including university- and or\dots}}
\end{minipage}\hfill
\begin{minipage}[t]{0.49\linewidth}
\fcolorbox{black!25}{gray!6}{\parbox{\dimexpr\linewidth-2\fboxsep-2\fboxrule\relax}{\scriptsize
\textbf{Sample 20}\hfill \textcolor{black!45}{\textbf{Not selected}}\\
\textbf{Candidate \#28}\hfill \texttt{score=0.0046}\\
\vspace{1pt}\\
\color{black!85} You really have to be alert when studying science. Galaxies were created after matter. The stars in those galaxies were supposed to move slowly because there was more mass in the center of the galaxy. However, after dark matter was added, the stars appeared to move faster; however, this is not the case in our galaxy, suggesting that there\dots}}
\end{minipage}
\vspace{3pt}
\noindent
\begin{minipage}[t]{0.49\linewidth}
\fcolorbox{black!25}{gray!6}{\parbox{\dimexpr\linewidth-2\fboxsep-2\fboxrule\relax}{\scriptsize
\textbf{Sample 21}\hfill \textcolor{black!45}{\textbf{Not selected}}\\
\textbf{Candidate \#21}\hfill \texttt{score=0.00457}\\
\vspace{1pt}\\
\color{black!85} Nestled in the leafy suburbs of western Berlin, the Wannsee Conference House stands as a poignant reminder of a dark chapter in human history. The Wannsee Conference: A Pivotal Moment The Wannsee Conference, held on January 20, 1942, marked a pivotal moment in the implementation of Nazi Germany's genocidal plans. Organized by SS-Obergrupp\dots}}
\end{minipage}\hfill
\begin{minipage}[t]{0.49\linewidth}
\fcolorbox{black!25}{gray!6}{\parbox{\dimexpr\linewidth-2\fboxsep-2\fboxrule\relax}{\scriptsize
\textbf{Sample 22}\hfill \textcolor{black!45}{\textbf{Not selected}}\\
\textbf{Candidate \#16}\hfill \texttt{score=0.00456}\\
\vspace{1pt}\\
\color{black!85} What is rotavirus and why does my baby need to be immunised? Rotavirus is a very infectious virus that causes the majority of serious cases of gastroenteritis in babies. It causes diarrhoea, vomiting and abdominal pain, usually lasting around a week. Most children will be infected by rotavirus once by the age of five. Gastroenteritis (cau\dots}}
\end{minipage}
\vspace{3pt}
\noindent
\begin{minipage}[t]{0.49\linewidth}
\fcolorbox{black!25}{gray!6}{\parbox{\dimexpr\linewidth-2\fboxsep-2\fboxrule\relax}{\scriptsize
\textbf{Sample 23}\hfill \textcolor{black!45}{\textbf{Not selected}}\\
\textbf{Candidate \#29}\hfill \texttt{score=0.00448}\\
\vspace{1pt}\\
\color{black!85} Earthquakes are the result of sudden movement along faults within the Earth. The movement releases stored-up ‘elastic strain’ energy in the form of seismic waves, which propagate through the Earth and cause the ground surface to shake. Such movement on the faults is generally a response to long-term deformation and the buildup of stress.\dots}}
\end{minipage}\hfill
\begin{minipage}[t]{0.49\linewidth}
\fcolorbox{black!25}{gray!6}{\parbox{\dimexpr\linewidth-2\fboxsep-2\fboxrule\relax}{\scriptsize
\textbf{Sample 24}\hfill \textcolor{black!45}{\textbf{Not selected}}\\
\textbf{Candidate \#13}\hfill \texttt{score=0.00445}\\
\vspace{1pt}\\
\color{black!85} In Heart of Darkness it is the white invaders for instance, who are, almost without exception, embodiments of blindness, selfishness, and cruelty; and even in the cognitive domain, where such positive phrases as “to enlighten,” for instance, are conventionally opposed to negative ones such as “to be in the dark,” the traditional expectati\dots}}
\end{minipage}
\vspace{3pt}
\noindent
\begin{minipage}[t]{0.49\linewidth}
\fcolorbox{black!25}{gray!6}{\parbox{\dimexpr\linewidth-2\fboxsep-2\fboxrule\relax}{\scriptsize
\textbf{Sample 25}\hfill \textcolor{black!45}{\textbf{Not selected}}\\
\textbf{Candidate \#26}\hfill \texttt{score=0.00443}\\
\vspace{1pt}\\
\color{black!85} Unveiling the Power: Key Provisions of the Civil Rights Act of 1864 What were the Civil Rights Act of 1864's key provisions? The Civil Rights Act of 1864 was a pivotal moment in American history, establishing crucial legal protections for African Americans in the face of rampant discrimination. Editor Note: The Civil Rights Act of 1864 la\dots}}
\end{minipage}\hfill
\begin{minipage}[t]{0.49\linewidth}
\fcolorbox{black!25}{gray!6}{\parbox{\dimexpr\linewidth-2\fboxsep-2\fboxrule\relax}{\scriptsize
\textbf{Sample 26}\hfill \textcolor{black!45}{\textbf{Not selected}}\\
\textbf{Candidate \#24}\hfill \texttt{score=0.00439}\\
\vspace{1pt}\\
\color{black!85} Next we will talk about solar radiation, that is, the forms of solar radiation that we receive on earth. Solar radiation is generated by a series of nuclear fusion reactions that occur in the Sun and, as a consequence, emit electromagnetic radiation that reaches the earth. This radiation received by the earth’s surface is measured in W /\dots}}
\end{minipage}
\vspace{3pt}
\noindent
\begin{minipage}[t]{0.49\linewidth}
\fcolorbox{black!25}{gray!6}{\parbox{\dimexpr\linewidth-2\fboxsep-2\fboxrule\relax}{\scriptsize
\textbf{Sample 27}\hfill \textcolor{black!45}{\textbf{Not selected}}\\
\textbf{Candidate \#17}\hfill \texttt{score=0.00427}\\
\vspace{1pt}\\
\color{black!85} Political Parties and Elections Political parties are an established part of modern mass democracy, and the conduct of elections in India is largely dependent on the behaviour of political parties. Although many candidates for Indian elections are independent, the winning candidates for Lok Sabha and Vidhan Sabha elections usually stand a\dots}}
\end{minipage}\hfill
\begin{minipage}[t]{0.49\linewidth}
\fcolorbox{black!25}{gray!6}{\parbox{\dimexpr\linewidth-2\fboxsep-2\fboxrule\relax}{\scriptsize
\textbf{Sample 28}\hfill \textcolor{black!45}{\textbf{Not selected}}\\
\textbf{Candidate \#10}\hfill \texttt{score=0.00427}\\
\vspace{1pt}\\
\color{black!85} Deforestation isn't just happening in well-known global hotspots like Indonesia and Brazil's rainforest. A new analysis says forests are also shrinking on state and private land in Oregon, where an estimated 522,000 acres of forest cover have disappeared since 2000. That's an area six times larger than the city of Portland, equal to more\dots}}
\end{minipage}
\vspace{3pt}
\noindent
\begin{minipage}[t]{0.49\linewidth}
\fcolorbox{black!25}{gray!6}{\parbox{\dimexpr\linewidth-2\fboxsep-2\fboxrule\relax}{\scriptsize
\textbf{Sample 29}\hfill \textcolor{black!45}{\textbf{Not selected}}\\
\textbf{Candidate \#6}\hfill \texttt{score=0.00401}\\
\vspace{1pt}\\
\color{black!85} Skaters need to check their skate helmets every so often and ask yourself, "Is it time to replace this helmet?" Well, that depends. Did you crash in it? For starters, most people are aware that you must replace a helmet after any crash where your head hit. The foam part of a helmet is made for one-time use, and after crushing once it is n\dots}}
\end{minipage}\hfill
\begin{minipage}[t]{0.49\linewidth}
\fcolorbox{black!25}{gray!6}{\parbox{\dimexpr\linewidth-2\fboxsep-2\fboxrule\relax}{\scriptsize
\textbf{Sample 30}\hfill \textcolor{black!45}{\textbf{Not selected}}\\
\textbf{Candidate \#2}\hfill \texttt{score=0.00384}\\
\vspace{1pt}\\
\color{black!85} With the advent of new technologies for sneakers such as Vac Tech, Hyperfuse and Flyknit, the mid 90s and early 2000s methods of production and designing are becoming obsolete in this sneaker world. Nike Running is the future for Nike, generating billions of dollars per year, and we see Nike also not afraid to experiment with technology s\dots}}
\end{minipage}
\vspace{3pt}
\noindent
\begin{minipage}[t]{0.49\linewidth}
\fcolorbox{black!25}{gray!6}{\parbox{\dimexpr\linewidth-2\fboxsep-2\fboxrule\relax}{\scriptsize
\textbf{Sample 31}\hfill \textcolor{black!45}{\textbf{Not selected}}\\
\textbf{Candidate \#12}\hfill \texttt{score=0.00369}\\
\vspace{1pt}\\
\color{black!85} Can you please give us a little short bio? (education, professional experiences, select publications, academic specialty, awards won) Public school teacher for 5 years BA art (UC Irvine) PhD. (UCLA) educational psychology Professor of Child Development, (25 years) CSUS Senior Research Scientist (Oregon Research Institute with Institute of\dots}}
\end{minipage}\hfill
\begin{minipage}[t]{0.49\linewidth}
\fcolorbox{black!25}{gray!6}{\parbox{\dimexpr\linewidth-2\fboxsep-2\fboxrule\relax}{\scriptsize
\textbf{Sample 32}\hfill \textcolor{black!45}{\textbf{Not selected}}\\
\textbf{Candidate \#3}\hfill \texttt{score=0.00333}\\
\vspace{1pt}\\
\color{black!85} starring John Travolta and Sam Jackson The first thing to understand about Basic --the basic thing, let's say-- is that although the commercials make it look like a war movie, it is not, for which we can all be grateful. No, Basic is a plot-twisty whodunnit. If The Usual Suspects died, and its body turned to cheese, and then that cheese-b\dots}}
\end{minipage}
\vspace{3pt}

\clearpage
\noindent\phantomsection
\label{app:qual_ppl_high}
\noindent \textbf{High-PPL} \\[6pt]
\noindent
\begin{minipage}[t]{0.49\linewidth}
\fcolorbox{black!25}{gray!6}{\parbox{\dimexpr\linewidth-2\fboxsep-2\fboxrule\relax}{\scriptsize
\textbf{Sample 1}\hfill \textcolor{green!45!black}{\textbf{Selected}}\\
\textbf{Candidate \#3}\hfill \texttt{score=4.57}\\
\vspace{1pt}\\
\color{black!85} starring John Travolta and Sam Jackson The first thing to understand about Basic --the basic thing, let's say-- is that although the commercials make it look like a war movie, it is not, for which we can all be grateful. No, Basic is a plot-twisty whodunnit. If The Usual Suspects died, and its body turned to cheese, and then that cheese-b\dots}}
\end{minipage}\hfill
\begin{minipage}[t]{0.49\linewidth}
\fcolorbox{black!25}{gray!6}{\parbox{\dimexpr\linewidth-2\fboxsep-2\fboxrule\relax}{\scriptsize
\textbf{Sample 2}\hfill \textcolor{green!45!black}{\textbf{Selected}}\\
\textbf{Candidate \#19}\hfill \texttt{score=4.26}\\
\vspace{1pt}\\
\color{black!85} Dividing Fractions Using Models Worksheet. This worksheet has six division with fractions issues to be solved — three must be solved with fashions and three with algorithms — options are on the second page. Answer key divide the unit fractions by whole numbers using th e fashions given. Use these resources to help reinforce the following\dots}}
\end{minipage}
\vspace{3pt}
\noindent
\begin{minipage}[t]{0.49\linewidth}
\fcolorbox{black!25}{gray!6}{\parbox{\dimexpr\linewidth-2\fboxsep-2\fboxrule\relax}{\scriptsize
\textbf{Sample 3}\hfill \textcolor{green!45!black}{\textbf{Selected}}\\
\textbf{Candidate \#12}\hfill \texttt{score=4.26}\\
\vspace{1pt}\\
\color{black!85} Can you please give us a little short bio? (education, professional experiences, select publications, academic specialty, awards won) Public school teacher for 5 years BA art (UC Irvine) PhD. (UCLA) educational psychology Professor of Child Development, (25 years) CSUS Senior Research Scientist (Oregon Research Institute with Institute of\dots}}
\end{minipage}\hfill
\begin{minipage}[t]{0.49\linewidth}
\fcolorbox{black!25}{gray!6}{\parbox{\dimexpr\linewidth-2\fboxsep-2\fboxrule\relax}{\scriptsize
\textbf{Sample 4}\hfill \textcolor{green!45!black}{\textbf{Selected}}\\
\textbf{Candidate \#5}\hfill \texttt{score=4.21}\\
\vspace{1pt}\\
\color{black!85} Well this is the big one. So big apparently, that I had to take it there and raise the number from 10 to 15. There’s just that many fails in the world of female rap. Some slight missteps, some EPIC. Nevertheless, they are all worth mentioning. You can probably think of a bunch more, but this is what I have gathered picking up from my prev\dots}}
\end{minipage}
\vspace{3pt}
\noindent
\begin{minipage}[t]{0.49\linewidth}
\fcolorbox{black!25}{gray!6}{\parbox{\dimexpr\linewidth-2\fboxsep-2\fboxrule\relax}{\scriptsize
\textbf{Sample 5}\hfill \textcolor{green!45!black}{\textbf{Selected}}\\
\textbf{Candidate \#2}\hfill \texttt{score=4.04}\\
\vspace{1pt}\\
\color{black!85} With the advent of new technologies for sneakers such as Vac Tech, Hyperfuse and Flyknit, the mid 90s and early 2000s methods of production and designing are becoming obsolete in this sneaker world. Nike Running is the future for Nike, generating billions of dollars per year, and we see Nike also not afraid to experiment with technology s\dots}}
\end{minipage}\hfill
\begin{minipage}[t]{0.49\linewidth}
\fcolorbox{black!25}{gray!6}{\parbox{\dimexpr\linewidth-2\fboxsep-2\fboxrule\relax}{\scriptsize
\textbf{Sample 6}\hfill \textcolor{green!45!black}{\textbf{Selected}}\\
\textbf{Candidate \#1}\hfill \texttt{score=3.89}\\
\vspace{1pt}\\
\color{black!85} Wedding \& Party Venues - Sort By: Edgartown : (508) 627-9510 A 19th century gothic revival home transformed into the island's premier eco-boutique hotel. Guests either stay in the 17-room Hob Knob hotel or in the privacy of their own Hob Knob House. Guests can expect individualized Hob Knob hospitality and modern luxury amenities in a rel\dots}}
\end{minipage}
\vspace{3pt}
\noindent
\begin{minipage}[t]{0.49\linewidth}
\fcolorbox{black!25}{gray!6}{\parbox{\dimexpr\linewidth-2\fboxsep-2\fboxrule\relax}{\scriptsize
\textbf{Sample 7}\hfill \textcolor{green!45!black}{\textbf{Selected}}\\
\textbf{Candidate \#7}\hfill \texttt{score=3.89}\\
\vspace{1pt}\\
\color{black!85} Elizabeth Hurley played as Dalila Release: Dec 8, 1996 Mara and her husband Manoa are both upstanding and religious Israelites living under the harsh and unjust rule of the Philistines. Much to their regret, they have not been able to have children. One day, a mysterious stranger appears to Mara and promises her that she will bear a son w\dots}}
\end{minipage}\hfill
\begin{minipage}[t]{0.49\linewidth}
\fcolorbox{black!25}{gray!6}{\parbox{\dimexpr\linewidth-2\fboxsep-2\fboxrule\relax}{\scriptsize
\textbf{Sample 8}\hfill \textcolor{green!45!black}{\textbf{Selected}}\\
\textbf{Candidate \#0}\hfill \texttt{score=3.82}\\
\vspace{1pt}\\
\color{black!85} As it turns out, the exercises synonymous with strong, attractive abs may not be the best way to train your core—and may be doing damage to your back. Read more If you are worried about the excess holiday pounds many of us are still carrying around. There are a few easy, natural things you can do to shed them, and none of them requires an\dots}}
\end{minipage}
\vspace{3pt}
\noindent
\begin{minipage}[t]{0.49\linewidth}
\fcolorbox{black!25}{gray!6}{\parbox{\dimexpr\linewidth-2\fboxsep-2\fboxrule\relax}{\scriptsize
\textbf{Sample 9}\hfill \textcolor{green!45!black}{\textbf{Selected}}\\
\textbf{Candidate \#13}\hfill \texttt{score=3.79}\\
\vspace{1pt}\\
\color{black!85} In Heart of Darkness it is the white invaders for instance, who are, almost without exception, embodiments of blindness, selfishness, and cruelty; and even in the cognitive domain, where such positive phrases as “to enlighten,” for instance, are conventionally opposed to negative ones such as “to be in the dark,” the traditional expectati\dots}}
\end{minipage}\hfill
\begin{minipage}[t]{0.49\linewidth}
\fcolorbox{black!25}{gray!6}{\parbox{\dimexpr\linewidth-2\fboxsep-2\fboxrule\relax}{\scriptsize
\textbf{Sample 10}\hfill \textcolor{green!45!black}{\textbf{Selected}}\\
\textbf{Candidate \#6}\hfill \texttt{score=3.76}\\
\vspace{1pt}\\
\color{black!85} Skaters need to check their skate helmets every so often and ask yourself, "Is it time to replace this helmet?" Well, that depends. Did you crash in it? For starters, most people are aware that you must replace a helmet after any crash where your head hit. The foam part of a helmet is made for one-time use, and after crushing once it is n\dots}}
\end{minipage}
\vspace{3pt}
\noindent
\begin{minipage}[t]{0.49\linewidth}
\fcolorbox{black!25}{gray!6}{\parbox{\dimexpr\linewidth-2\fboxsep-2\fboxrule\relax}{\scriptsize
\textbf{Sample 11}\hfill \textcolor{green!45!black}{\textbf{Selected}}\\
\textbf{Candidate \#4}\hfill \texttt{score=3.74}\\
\vspace{1pt}\\
\color{black!85} 5 Types of Women’s Underwear That Men Love Underwear can say a lot about a woman. It’s something that men are obsessed with, to the point that, a mere glimpse of a thong waistband causes us to go into shock. On the surface we find them sexy, revealing. We’re able to see who a woman actually is—or maybe some guys are just plain horny. Howe\dots}}
\end{minipage}\hfill
\begin{minipage}[t]{0.49\linewidth}
\fcolorbox{black!25}{gray!6}{\parbox{\dimexpr\linewidth-2\fboxsep-2\fboxrule\relax}{\scriptsize
\textbf{Sample 12}\hfill \textcolor{green!45!black}{\textbf{Selected}}\\
\textbf{Candidate \#9}\hfill \texttt{score=3.64}\\
\vspace{1pt}\\
\color{black!85} “Last night three cargoes of Bohea Tea were emptied into the sea. This is the most magnificent movement of all. There is a dignity, a majesty, a sublimity, in this last effort of the Patriots that I greatly admire.” - John Adams, diary entry, December 17, 1773 - John Adams, diary entry, December 17, 1773 A Novel Idea Is something so new a\dots}}
\end{minipage}
\vspace{3pt}
\noindent
\begin{minipage}[t]{0.49\linewidth}
\fcolorbox{black!25}{gray!6}{\parbox{\dimexpr\linewidth-2\fboxsep-2\fboxrule\relax}{\scriptsize
\textbf{Sample 13}\hfill \textcolor{green!45!black}{\textbf{Selected}}\\
\textbf{Candidate \#23}\hfill \texttt{score=3.43}\\
\vspace{1pt}\\
\color{black!85} The St. James kindergarteners have been working up to Project Week over the past month. We started slowly by taking walks in our neighborhood while Ms. Meghan and I noted what caught the children’s interest. It became apparent that the class was very interested in the L trains that they saw on our walks. It started with a simple question,\dots}}
\end{minipage}\hfill
\begin{minipage}[t]{0.49\linewidth}
\fcolorbox{black!25}{gray!6}{\parbox{\dimexpr\linewidth-2\fboxsep-2\fboxrule\relax}{\scriptsize
\textbf{Sample 14}\hfill \textcolor{green!45!black}{\textbf{Selected}}\\
\textbf{Candidate \#31}\hfill \texttt{score=3.43}\\
\vspace{1pt}\\
\color{black!85} One of the challenges of working with ancient DNA samples is that damage accumulates over time, breaking the double helix structure into ever-smaller fragments. In the samples we worked with, these fragments were scattered and mixed with contaminants, making genome reconstruction a major technical challenge. But a shocking paper published\dots}}
\end{minipage}
\vspace{3pt}
\noindent
\begin{minipage}[t]{0.49\linewidth}
\fcolorbox{black!25}{gray!6}{\parbox{\dimexpr\linewidth-2\fboxsep-2\fboxrule\relax}{\scriptsize
\textbf{Sample 15}\hfill \textcolor{green!45!black}{\textbf{Selected}}\\
\textbf{Candidate \#11}\hfill \texttt{score=3.40}\\
\vspace{1pt}\\
\color{black!85} In decades past, classroom design was often an afterthought and followed a standardised layout. Plain boxed shaped classrooms, with identical chairs and tables throughout were commonplace in many schools. Read the latest issue of School News HERE Recently, though, there has been a shift away from this one-size-fits all approach to classro\dots}}
\end{minipage}\hfill
\begin{minipage}[t]{0.49\linewidth}
\fcolorbox{black!25}{gray!6}{\parbox{\dimexpr\linewidth-2\fboxsep-2\fboxrule\relax}{\scriptsize
\textbf{Sample 16}\hfill \textcolor{green!45!black}{\textbf{Selected}}\\
\textbf{Candidate \#10}\hfill \texttt{score=3.38}\\
\vspace{1pt}\\
\color{black!85} Deforestation isn't just happening in well-known global hotspots like Indonesia and Brazil's rainforest. A new analysis says forests are also shrinking on state and private land in Oregon, where an estimated 522,000 acres of forest cover have disappeared since 2000. That's an area six times larger than the city of Portland, equal to more\dots}}
\end{minipage}
\vspace{3pt}
\clearpage
\noindent\phantomsection
\label{app:qual_ppl_high_cont}
% \noindent \textbf{High-PPL} \\[6pt]
\noindent
\begin{minipage}[t]{0.49\linewidth}
\fcolorbox{black!25}{gray!6}{\parbox{\dimexpr\linewidth-2\fboxsep-2\fboxrule\relax}{\scriptsize
\textbf{Sample 17}\hfill \textcolor{black!45}{\textbf{Not selected}}\\
\textbf{Candidate \#28}\hfill \texttt{score=3.22}\\
\vspace{1pt}\\
\color{black!85} You really have to be alert when studying science. Galaxies were created after matter. The stars in those galaxies were supposed to move slowly because there was more mass in the center of the galaxy. However, after dark matter was added, the stars appeared to move faster; however, this is not the case in our galaxy, suggesting that there\dots}}
\end{minipage}\hfill
\begin{minipage}[t]{0.49\linewidth}
\fcolorbox{black!25}{gray!6}{\parbox{\dimexpr\linewidth-2\fboxsep-2\fboxrule\relax}{\scriptsize
\textbf{Sample 18}\hfill \textcolor{black!45}{\textbf{Not selected}}\\
\textbf{Candidate \#17}\hfill \texttt{score=3.19}\\
\vspace{1pt}\\
\color{black!85} Political Parties and Elections Political parties are an established part of modern mass democracy, and the conduct of elections in India is largely dependent on the behaviour of political parties. Although many candidates for Indian elections are independent, the winning candidates for Lok Sabha and Vidhan Sabha elections usually stand a\dots}}
\end{minipage}
\vspace{3pt}
\noindent
\begin{minipage}[t]{0.49\linewidth}
\fcolorbox{black!25}{gray!6}{\parbox{\dimexpr\linewidth-2\fboxsep-2\fboxrule\relax}{\scriptsize
\textbf{Sample 19}\hfill \textcolor{black!45}{\textbf{Not selected}}\\
\textbf{Candidate \#25}\hfill \texttt{score=3.05}\\
\vspace{1pt}\\
\color{black!85} KS2 Maths is an important core subject in the National Curriculum and this area of the website covers all the major aspects of the curriculum including numbers, calculations, problems and measures. Each subject area is designed to help children develop their knowledge, whether they are learning in a classroom or home schooling environment\dots}}
\end{minipage}\hfill
\begin{minipage}[t]{0.49\linewidth}
\fcolorbox{black!25}{gray!6}{\parbox{\dimexpr\linewidth-2\fboxsep-2\fboxrule\relax}{\scriptsize
\textbf{Sample 20}\hfill \textcolor{black!45}{\textbf{Not selected}}\\
\textbf{Candidate \#27}\hfill \texttt{score=3.05}\\
\vspace{1pt}\\
\color{black!85} 24/7 writing help on your phone Save to my list Remove from my list In the tumultuous 19th century, both Italy and Germany found themselves fragmented into numerous separate ruling states. The impetus for change came in the form of rising nationalism and liberalism, paving the way for the unification of these disparate entities. However,\dots}}
\end{minipage}
\vspace{3pt}
\noindent
\begin{minipage}[t]{0.49\linewidth}
\fcolorbox{black!25}{gray!6}{\parbox{\dimexpr\linewidth-2\fboxsep-2\fboxrule\relax}{\scriptsize
\textbf{Sample 21}\hfill \textcolor{black!45}{\textbf{Not selected}}\\
\textbf{Candidate \#18}\hfill \texttt{score=3.03}\\
\vspace{1pt}\\
\color{black!85} This article originally appeared in the December 2015 issue of Resource Recycling. Subscribe today for access to all print content. Since the 1990s, curbside and drop-off recycling has grown substantially – nearly 90 percent of households now have access, according to recent surveys from Moore Recycling Associates, the American Forest and\dots}}
\end{minipage}\hfill
\begin{minipage}[t]{0.49\linewidth}
\fcolorbox{black!25}{gray!6}{\parbox{\dimexpr\linewidth-2\fboxsep-2\fboxrule\relax}{\scriptsize
\textbf{Sample 22}\hfill \textcolor{black!45}{\textbf{Not selected}}\\
\textbf{Candidate \#15}\hfill \texttt{score=2.95}\\
\vspace{1pt}\\
\color{black!85} Origami is an art form that combines precision, creativity, and patience. While basic origami is obtainable to every one, mastering complex origami designs can be quite a rewarding and impressive achievement. In this article, we’ll show you with the procedure for creating intricate origami while highlighting essential techniques for achie\dots}}
\end{minipage}
\vspace{3pt}
\noindent
\begin{minipage}[t]{0.49\linewidth}
\fcolorbox{black!25}{gray!6}{\parbox{\dimexpr\linewidth-2\fboxsep-2\fboxrule\relax}{\scriptsize
\textbf{Sample 23}\hfill \textcolor{black!45}{\textbf{Not selected}}\\
\textbf{Candidate \#29}\hfill \texttt{score=2.91}\\
\vspace{1pt}\\
\color{black!85} Earthquakes are the result of sudden movement along faults within the Earth. The movement releases stored-up ‘elastic strain’ energy in the form of seismic waves, which propagate through the Earth and cause the ground surface to shake. Such movement on the faults is generally a response to long-term deformation and the buildup of stress.\dots}}
\end{minipage}\hfill
\begin{minipage}[t]{0.49\linewidth}
\fcolorbox{black!25}{gray!6}{\parbox{\dimexpr\linewidth-2\fboxsep-2\fboxrule\relax}{\scriptsize
\textbf{Sample 24}\hfill \textcolor{black!45}{\textbf{Not selected}}\\
\textbf{Candidate \#22}\hfill \texttt{score=2.90}\\
\vspace{1pt}\\
\color{black!85} How To Choose Decodable Readers for First Grade To decode or not to decode: really, there is no question. To help rising first graders become successful and enthusiastic readers this summer, decodable readers are essential reading resources. Although “decodable text” might sound like yet another form of educational lingo, parents and educ\dots}}
\end{minipage}
\vspace{3pt}
\noindent
\begin{minipage}[t]{0.49\linewidth}
\fcolorbox{black!25}{gray!6}{\parbox{\dimexpr\linewidth-2\fboxsep-2\fboxrule\relax}{\scriptsize
\textbf{Sample 25}\hfill \textcolor{black!45}{\textbf{Not selected}}\\
\textbf{Candidate \#24}\hfill \texttt{score=2.83}\\
\vspace{1pt}\\
\color{black!85} Next we will talk about solar radiation, that is, the forms of solar radiation that we receive on earth. Solar radiation is generated by a series of nuclear fusion reactions that occur in the Sun and, as a consequence, emit electromagnetic radiation that reaches the earth. This radiation received by the earth’s surface is measured in W /\dots}}
\end{minipage}\hfill
\begin{minipage}[t]{0.49\linewidth}
\fcolorbox{black!25}{gray!6}{\parbox{\dimexpr\linewidth-2\fboxsep-2\fboxrule\relax}{\scriptsize
\textbf{Sample 26}\hfill \textcolor{black!45}{\textbf{Not selected}}\\
\textbf{Candidate \#14}\hfill \texttt{score=2.73}\\
\vspace{1pt}\\
\color{black!85} Is your major sustainable enough? Whether you’re pursuing a sustainability degree and want to further your knowledge, or are interested in supplementing your major in another area with sustainability education, plenty of independent learning resources are available. A wide range of credit and noncredit courses—including university- and or\dots}}
\end{minipage}
\vspace{3pt}
\noindent
\begin{minipage}[t]{0.49\linewidth}
\fcolorbox{black!25}{gray!6}{\parbox{\dimexpr\linewidth-2\fboxsep-2\fboxrule\relax}{\scriptsize
\textbf{Sample 27}\hfill \textcolor{black!45}{\textbf{Not selected}}\\
\textbf{Candidate \#21}\hfill \texttt{score=2.61}\\
\vspace{1pt}\\
\color{black!85} Nestled in the leafy suburbs of western Berlin, the Wannsee Conference House stands as a poignant reminder of a dark chapter in human history. The Wannsee Conference: A Pivotal Moment The Wannsee Conference, held on January 20, 1942, marked a pivotal moment in the implementation of Nazi Germany's genocidal plans. Organized by SS-Obergrupp\dots}}
\end{minipage}\hfill
\begin{minipage}[t]{0.49\linewidth}
\fcolorbox{black!25}{gray!6}{\parbox{\dimexpr\linewidth-2\fboxsep-2\fboxrule\relax}{\scriptsize
\textbf{Sample 28}\hfill \textcolor{black!45}{\textbf{Not selected}}\\
\textbf{Candidate \#20}\hfill \texttt{score=2.60}\\
\vspace{1pt}\\
\color{black!85} Conduct Disorder (CD) is a complex and serious behavioural and emotional disorder that can occur in children and adolescents. It’s characterised by a repetitive and persistent pattern of behaviour where the basic rights of others or major age-appropriate societal norms or rules are violated. Here’s an outline of Conduct Disorder in line w\dots}}
\end{minipage}
\vspace{3pt}
\noindent
\begin{minipage}[t]{0.49\linewidth}
\fcolorbox{black!25}{gray!6}{\parbox{\dimexpr\linewidth-2\fboxsep-2\fboxrule\relax}{\scriptsize
\textbf{Sample 29}\hfill \textcolor{black!45}{\textbf{Not selected}}\\
\textbf{Candidate \#16}\hfill \texttt{score=2.42}\\
\vspace{1pt}\\
\color{black!85} What is rotavirus and why does my baby need to be immunised? Rotavirus is a very infectious virus that causes the majority of serious cases of gastroenteritis in babies. It causes diarrhoea, vomiting and abdominal pain, usually lasting around a week. Most children will be infected by rotavirus once by the age of five. Gastroenteritis (cau\dots}}
\end{minipage}\hfill
\begin{minipage}[t]{0.49\linewidth}
\fcolorbox{black!25}{gray!6}{\parbox{\dimexpr\linewidth-2\fboxsep-2\fboxrule\relax}{\scriptsize
\textbf{Sample 30}\hfill \textcolor{black!45}{\textbf{Not selected}}\\
\textbf{Candidate \#26}\hfill \texttt{score=2.31}\\
\vspace{1pt}\\
\color{black!85} Unveiling the Power: Key Provisions of the Civil Rights Act of 1864 What were the Civil Rights Act of 1864's key provisions? The Civil Rights Act of 1864 was a pivotal moment in American history, establishing crucial legal protections for African Americans in the face of rampant discrimination. Editor Note: The Civil Rights Act of 1864 la\dots}}
\end{minipage}
\vspace{3pt}
\noindent
\begin{minipage}[t]{0.49\linewidth}
\fcolorbox{black!25}{gray!6}{\parbox{\dimexpr\linewidth-2\fboxsep-2\fboxrule\relax}{\scriptsize
\textbf{Sample 31}\hfill \textcolor{black!45}{\textbf{Not selected}}\\
\textbf{Candidate \#30}\hfill \texttt{score=2.28}\\
\vspace{1pt}\\
\color{black!85} Over 1.8 million professionals use CFI to learn accounting, financial analysis, modeling and more. Start with a free account to explore 20+ always-free courses and hundreds of finance templates and cheat sheets. What is the Central Limit Theorem (CLT)? The Central Limit Theorem (CLT) is a statistical concept that states that the sample me\dots}}
\end{minipage}\hfill
\begin{minipage}[t]{0.49\linewidth}
\fcolorbox{black!25}{gray!6}{\parbox{\dimexpr\linewidth-2\fboxsep-2\fboxrule\relax}{\scriptsize
\textbf{Sample 32}\hfill \textcolor{black!45}{\textbf{Not selected}}\\
\textbf{Candidate \#8}\hfill \texttt{score=1.58}\\
\vspace{1pt}\\
\color{black!85} The Unsung Heroes of Your HVAC System: Understanding the Importance of Filters When it comes to your HVAC (Heating, Ventilation, and Air Conditioning) system, you might be quick to think about the thermostat, air ducts, or even the unit itself. However, there’s an unsung hero in your HVAC system that plays a pivotal role in maintaining in\dots}}
\end{minipage}
\vspace{3pt}

\clearpage
\noindent\phantomsection
\label{app:qual_sgd}
\noindent \textbf{GREATS} \\[6pt]
\noindent
\begin{minipage}[t]{0.49\linewidth}
\fcolorbox{black!25}{gray!6}{\parbox{\dimexpr\linewidth-2\fboxsep-2\fboxrule\relax}{\scriptsize
\textbf{Sample 1}\hfill \textcolor{green!45!black}{\textbf{Selected}}\\
\textbf{Candidate \#8}\hfill \texttt{score=17.40}\\
\vspace{1pt}\\
\color{black!85} The Unsung Heroes of Your HVAC System: Understanding the Importance of Filters When it comes to your HVAC (Heating, Ventilation, and Air Conditioning) system, you might be quick to think about the thermostat, air ducts, or even the unit itself. However, there’s an unsung hero in your HVAC system that plays a pivotal role in maintaining in\dots}}
\end{minipage}\hfill
\begin{minipage}[t]{0.49\linewidth}
\fcolorbox{black!25}{gray!6}{\parbox{\dimexpr\linewidth-2\fboxsep-2\fboxrule\relax}{\scriptsize
\textbf{Sample 2}\hfill \textcolor{green!45!black}{\textbf{Selected}}\\
\textbf{Candidate \#15}\hfill \texttt{score=15.47}\\
\vspace{1pt}\\
\color{black!85} Origami is an art form that combines precision, creativity, and patience. While basic origami is obtainable to every one, mastering complex origami designs can be quite a rewarding and impressive achievement. In this article, we’ll show you with the procedure for creating intricate origami while highlighting essential techniques for achie\dots}}
\end{minipage}
\vspace{3pt}
\noindent
\begin{minipage}[t]{0.49\linewidth}
\fcolorbox{black!25}{gray!6}{\parbox{\dimexpr\linewidth-2\fboxsep-2\fboxrule\relax}{\scriptsize
\textbf{Sample 3}\hfill \textcolor{green!45!black}{\textbf{Selected}}\\
\textbf{Candidate \#20}\hfill \texttt{score=15.18}\\
\vspace{1pt}\\
\color{black!85} Conduct Disorder (CD) is a complex and serious behavioural and emotional disorder that can occur in children and adolescents. It’s characterised by a repetitive and persistent pattern of behaviour where the basic rights of others or major age-appropriate societal norms or rules are violated. Here’s an outline of Conduct Disorder in line w\dots}}
\end{minipage}\hfill
\begin{minipage}[t]{0.49\linewidth}
\fcolorbox{black!25}{gray!6}{\parbox{\dimexpr\linewidth-2\fboxsep-2\fboxrule\relax}{\scriptsize
\textbf{Sample 4}\hfill \textcolor{green!45!black}{\textbf{Selected}}\\
\textbf{Candidate \#14}\hfill \texttt{score=13.88}\\
\vspace{1pt}\\
\color{black!85} Is your major sustainable enough? Whether you’re pursuing a sustainability degree and want to further your knowledge, or are interested in supplementing your major in another area with sustainability education, plenty of independent learning resources are available. A wide range of credit and noncredit courses—including university- and or\dots}}
\end{minipage}
\vspace{3pt}
\noindent
\begin{minipage}[t]{0.49\linewidth}
\fcolorbox{black!25}{gray!6}{\parbox{\dimexpr\linewidth-2\fboxsep-2\fboxrule\relax}{\scriptsize
\textbf{Sample 5}\hfill \textcolor{green!45!black}{\textbf{Selected}}\\
\textbf{Candidate \#22}\hfill \texttt{score=13.87}\\
\vspace{1pt}\\
\color{black!85} How To Choose Decodable Readers for First Grade To decode or not to decode: really, there is no question. To help rising first graders become successful and enthusiastic readers this summer, decodable readers are essential reading resources. Although “decodable text” might sound like yet another form of educational lingo, parents and educ\dots}}
\end{minipage}\hfill
\begin{minipage}[t]{0.49\linewidth}
\fcolorbox{black!25}{gray!6}{\parbox{\dimexpr\linewidth-2\fboxsep-2\fboxrule\relax}{\scriptsize
\textbf{Sample 6}\hfill \textcolor{green!45!black}{\textbf{Selected}}\\
\textbf{Candidate \#27}\hfill \texttt{score=13.71}\\
\vspace{1pt}\\
\color{black!85} 24/7 writing help on your phone Save to my list Remove from my list In the tumultuous 19th century, both Italy and Germany found themselves fragmented into numerous separate ruling states. The impetus for change came in the form of rising nationalism and liberalism, paving the way for the unification of these disparate entities. However,\dots}}
\end{minipage}
\vspace{3pt}
\noindent
\begin{minipage}[t]{0.49\linewidth}
\fcolorbox{black!25}{gray!6}{\parbox{\dimexpr\linewidth-2\fboxsep-2\fboxrule\relax}{\scriptsize
\textbf{Sample 7}\hfill \textcolor{green!45!black}{\textbf{Selected}}\\
\textbf{Candidate \#4}\hfill \texttt{score=13.64}\\
\vspace{1pt}\\
\color{black!85} 5 Types of Women’s Underwear That Men Love Underwear can say a lot about a woman. It’s something that men are obsessed with, to the point that, a mere glimpse of a thong waistband causes us to go into shock. On the surface we find them sexy, revealing. We’re able to see who a woman actually is—or maybe some guys are just plain horny. Howe\dots}}
\end{minipage}\hfill
\begin{minipage}[t]{0.49\linewidth}
\fcolorbox{black!25}{gray!6}{\parbox{\dimexpr\linewidth-2\fboxsep-2\fboxrule\relax}{\scriptsize
\textbf{Sample 8}\hfill \textcolor{green!45!black}{\textbf{Selected}}\\
\textbf{Candidate \#28}\hfill \texttt{score=13.60}\\
\vspace{1pt}\\
\color{black!85} You really have to be alert when studying science. Galaxies were created after matter. The stars in those galaxies were supposed to move slowly because there was more mass in the center of the galaxy. However, after dark matter was added, the stars appeared to move faster; however, this is not the case in our galaxy, suggesting that there\dots}}
\end{minipage}
\vspace{3pt}
\noindent
\begin{minipage}[t]{0.49\linewidth}
\fcolorbox{black!25}{gray!6}{\parbox{\dimexpr\linewidth-2\fboxsep-2\fboxrule\relax}{\scriptsize
\textbf{Sample 9}\hfill \textcolor{green!45!black}{\textbf{Selected}}\\
\textbf{Candidate \#21}\hfill \texttt{score=13.45}\\
\vspace{1pt}\\
\color{black!85} Nestled in the leafy suburbs of western Berlin, the Wannsee Conference House stands as a poignant reminder of a dark chapter in human history. The Wannsee Conference: A Pivotal Moment The Wannsee Conference, held on January 20, 1942, marked a pivotal moment in the implementation of Nazi Germany's genocidal plans. Organized by SS-Obergrupp\dots}}
\end{minipage}\hfill
\begin{minipage}[t]{0.49\linewidth}
\fcolorbox{black!25}{gray!6}{\parbox{\dimexpr\linewidth-2\fboxsep-2\fboxrule\relax}{\scriptsize
\textbf{Sample 10}\hfill \textcolor{green!45!black}{\textbf{Selected}}\\
\textbf{Candidate \#16}\hfill \texttt{score=13.43}\\
\vspace{1pt}\\
\color{black!85} What is rotavirus and why does my baby need to be immunised? Rotavirus is a very infectious virus that causes the majority of serious cases of gastroenteritis in babies. It causes diarrhoea, vomiting and abdominal pain, usually lasting around a week. Most children will be infected by rotavirus once by the age of five. Gastroenteritis (cau\dots}}
\end{minipage}
\vspace{3pt}
\noindent
\begin{minipage}[t]{0.49\linewidth}
\fcolorbox{black!25}{gray!6}{\parbox{\dimexpr\linewidth-2\fboxsep-2\fboxrule\relax}{\scriptsize
\textbf{Sample 11}\hfill \textcolor{green!45!black}{\textbf{Selected}}\\
\textbf{Candidate \#29}\hfill \texttt{score=13.17}\\
\vspace{1pt}\\
\color{black!85} Earthquakes are the result of sudden movement along faults within the Earth. The movement releases stored-up ‘elastic strain’ energy in the form of seismic waves, which propagate through the Earth and cause the ground surface to shake. Such movement on the faults is generally a response to long-term deformation and the buildup of stress.\dots}}
\end{minipage}\hfill
\begin{minipage}[t]{0.49\linewidth}
\fcolorbox{black!25}{gray!6}{\parbox{\dimexpr\linewidth-2\fboxsep-2\fboxrule\relax}{\scriptsize
\textbf{Sample 12}\hfill \textcolor{green!45!black}{\textbf{Selected}}\\
\textbf{Candidate \#13}\hfill \texttt{score=13.14}\\
\vspace{1pt}\\
\color{black!85} In Heart of Darkness it is the white invaders for instance, who are, almost without exception, embodiments of blindness, selfishness, and cruelty; and even in the cognitive domain, where such positive phrases as “to enlighten,” for instance, are conventionally opposed to negative ones such as “to be in the dark,” the traditional expectati\dots}}
\end{minipage}
\vspace{3pt}
\noindent
\begin{minipage}[t]{0.49\linewidth}
\fcolorbox{black!25}{gray!6}{\parbox{\dimexpr\linewidth-2\fboxsep-2\fboxrule\relax}{\scriptsize
\textbf{Sample 13}\hfill \textcolor{green!45!black}{\textbf{Selected}}\\
\textbf{Candidate \#26}\hfill \texttt{score=13.01}\\
\vspace{1pt}\\
\color{black!85} Unveiling the Power: Key Provisions of the Civil Rights Act of 1864 What were the Civil Rights Act of 1864's key provisions? The Civil Rights Act of 1864 was a pivotal moment in American history, establishing crucial legal protections for African Americans in the face of rampant discrimination. Editor Note: The Civil Rights Act of 1864 la\dots}}
\end{minipage}\hfill
\begin{minipage}[t]{0.49\linewidth}
\fcolorbox{black!25}{gray!6}{\parbox{\dimexpr\linewidth-2\fboxsep-2\fboxrule\relax}{\scriptsize
\textbf{Sample 14}\hfill \textcolor{green!45!black}{\textbf{Selected}}\\
\textbf{Candidate \#18}\hfill \texttt{score=12.93}\\
\vspace{1pt}\\
\color{black!85} This article originally appeared in the December 2015 issue of Resource Recycling. Subscribe today for access to all print content. Since the 1990s, curbside and drop-off recycling has grown substantially – nearly 90 percent of households now have access, according to recent surveys from Moore Recycling Associates, the American Forest and\dots}}
\end{minipage}
\vspace{3pt}
\noindent
\begin{minipage}[t]{0.49\linewidth}
\fcolorbox{black!25}{gray!6}{\parbox{\dimexpr\linewidth-2\fboxsep-2\fboxrule\relax}{\scriptsize
\textbf{Sample 15}\hfill \textcolor{green!45!black}{\textbf{Selected}}\\
\textbf{Candidate \#24}\hfill \texttt{score=12.92}\\
\vspace{1pt}\\
\color{black!85} Next we will talk about solar radiation, that is, the forms of solar radiation that we receive on earth. Solar radiation is generated by a series of nuclear fusion reactions that occur in the Sun and, as a consequence, emit electromagnetic radiation that reaches the earth. This radiation received by the earth’s surface is measured in W /\dots}}
\end{minipage}\hfill
\begin{minipage}[t]{0.49\linewidth}
\fcolorbox{black!25}{gray!6}{\parbox{\dimexpr\linewidth-2\fboxsep-2\fboxrule\relax}{\scriptsize
\textbf{Sample 16}\hfill \textcolor{green!45!black}{\textbf{Selected}}\\
\textbf{Candidate \#17}\hfill \texttt{score=12.64}\\
\vspace{1pt}\\
\color{black!85} Political Parties and Elections Political parties are an established part of modern mass democracy, and the conduct of elections in India is largely dependent on the behaviour of political parties. Although many candidates for Indian elections are independent, the winning candidates for Lok Sabha and Vidhan Sabha elections usually stand a\dots}}
\end{minipage}
\vspace{3pt}
\clearpage
\noindent\phantomsection
\label{app:qual_sgd_cont}
% \noindent \textbf{GREATS} \\[6pt]
\noindent
\begin{minipage}[t]{0.49\linewidth}
\fcolorbox{black!25}{gray!6}{\parbox{\dimexpr\linewidth-2\fboxsep-2\fboxrule\relax}{\scriptsize
\textbf{Sample 17}\hfill \textcolor{black!45}{\textbf{Not selected}}\\
\textbf{Candidate \#10}\hfill \texttt{score=12.53}\\
\vspace{1pt}\\
\color{black!85} Deforestation isn't just happening in well-known global hotspots like Indonesia and Brazil's rainforest. A new analysis says forests are also shrinking on state and private land in Oregon, where an estimated 522,000 acres of forest cover have disappeared since 2000. That's an area six times larger than the city of Portland, equal to more\dots}}
\end{minipage}\hfill
\begin{minipage}[t]{0.49\linewidth}
\fcolorbox{black!25}{gray!6}{\parbox{\dimexpr\linewidth-2\fboxsep-2\fboxrule\relax}{\scriptsize
\textbf{Sample 18}\hfill \textcolor{black!45}{\textbf{Not selected}}\\
\textbf{Candidate \#0}\hfill \texttt{score=12.38}\\
\vspace{1pt}\\
\color{black!85} As it turns out, the exercises synonymous with strong, attractive abs may not be the best way to train your core—and may be doing damage to your back. Read more If you are worried about the excess holiday pounds many of us are still carrying around. There are a few easy, natural things you can do to shed them, and none of them requires an\dots}}
\end{minipage}
\vspace{3pt}
\noindent
\begin{minipage}[t]{0.49\linewidth}
\fcolorbox{black!25}{gray!6}{\parbox{\dimexpr\linewidth-2\fboxsep-2\fboxrule\relax}{\scriptsize
\textbf{Sample 19}\hfill \textcolor{black!45}{\textbf{Not selected}}\\
\textbf{Candidate \#23}\hfill \texttt{score=12.36}\\
\vspace{1pt}\\
\color{black!85} The St. James kindergarteners have been working up to Project Week over the past month. We started slowly by taking walks in our neighborhood while Ms. Meghan and I noted what caught the children’s interest. It became apparent that the class was very interested in the L trains that they saw on our walks. It started with a simple question,\dots}}
\end{minipage}\hfill
\begin{minipage}[t]{0.49\linewidth}
\fcolorbox{black!25}{gray!6}{\parbox{\dimexpr\linewidth-2\fboxsep-2\fboxrule\relax}{\scriptsize
\textbf{Sample 20}\hfill \textcolor{black!45}{\textbf{Not selected}}\\
\textbf{Candidate \#30}\hfill \texttt{score=12.31}\\
\vspace{1pt}\\
\color{black!85} Over 1.8 million professionals use CFI to learn accounting, financial analysis, modeling and more. Start with a free account to explore 20+ always-free courses and hundreds of finance templates and cheat sheets. What is the Central Limit Theorem (CLT)? The Central Limit Theorem (CLT) is a statistical concept that states that the sample me\dots}}
\end{minipage}
\vspace{3pt}
\noindent
\begin{minipage}[t]{0.49\linewidth}
\fcolorbox{black!25}{gray!6}{\parbox{\dimexpr\linewidth-2\fboxsep-2\fboxrule\relax}{\scriptsize
\textbf{Sample 21}\hfill \textcolor{black!45}{\textbf{Not selected}}\\
\textbf{Candidate \#31}\hfill \texttt{score=12.04}\\
\vspace{1pt}\\
\color{black!85} One of the challenges of working with ancient DNA samples is that damage accumulates over time, breaking the double helix structure into ever-smaller fragments. In the samples we worked with, these fragments were scattered and mixed with contaminants, making genome reconstruction a major technical challenge. But a shocking paper published\dots}}
\end{minipage}\hfill
\begin{minipage}[t]{0.49\linewidth}
\fcolorbox{black!25}{gray!6}{\parbox{\dimexpr\linewidth-2\fboxsep-2\fboxrule\relax}{\scriptsize
\textbf{Sample 22}\hfill \textcolor{black!45}{\textbf{Not selected}}\\
\textbf{Candidate \#11}\hfill \texttt{score=11.85}\\
\vspace{1pt}\\
\color{black!85} In decades past, classroom design was often an afterthought and followed a standardised layout. Plain boxed shaped classrooms, with identical chairs and tables throughout were commonplace in many schools. Read the latest issue of School News HERE Recently, though, there has been a shift away from this one-size-fits all approach to classro\dots}}
\end{minipage}
\vspace{3pt}
\noindent
\begin{minipage}[t]{0.49\linewidth}
\fcolorbox{black!25}{gray!6}{\parbox{\dimexpr\linewidth-2\fboxsep-2\fboxrule\relax}{\scriptsize
\textbf{Sample 23}\hfill \textcolor{black!45}{\textbf{Not selected}}\\
\textbf{Candidate \#6}\hfill \texttt{score=11.82}\\
\vspace{1pt}\\
\color{black!85} Skaters need to check their skate helmets every so often and ask yourself, "Is it time to replace this helmet?" Well, that depends. Did you crash in it? For starters, most people are aware that you must replace a helmet after any crash where your head hit. The foam part of a helmet is made for one-time use, and after crushing once it is n\dots}}
\end{minipage}\hfill
\begin{minipage}[t]{0.49\linewidth}
\fcolorbox{black!25}{gray!6}{\parbox{\dimexpr\linewidth-2\fboxsep-2\fboxrule\relax}{\scriptsize
\textbf{Sample 24}\hfill \textcolor{black!45}{\textbf{Not selected}}\\
\textbf{Candidate \#25}\hfill \texttt{score=11.51}\\
\vspace{1pt}\\
\color{black!85} KS2 Maths is an important core subject in the National Curriculum and this area of the website covers all the major aspects of the curriculum including numbers, calculations, problems and measures. Each subject area is designed to help children develop their knowledge, whether they are learning in a classroom or home schooling environment\dots}}
\end{minipage}
\vspace{3pt}
\noindent
\begin{minipage}[t]{0.49\linewidth}
\fcolorbox{black!25}{gray!6}{\parbox{\dimexpr\linewidth-2\fboxsep-2\fboxrule\relax}{\scriptsize
\textbf{Sample 25}\hfill \textcolor{black!45}{\textbf{Not selected}}\\
\textbf{Candidate \#7}\hfill \texttt{score=11.49}\\
\vspace{1pt}\\
\color{black!85} Elizabeth Hurley played as Dalila Release: Dec 8, 1996 Mara and her husband Manoa are both upstanding and religious Israelites living under the harsh and unjust rule of the Philistines. Much to their regret, they have not been able to have children. One day, a mysterious stranger appears to Mara and promises her that she will bear a son w\dots}}
\end{minipage}\hfill
\begin{minipage}[t]{0.49\linewidth}
\fcolorbox{black!25}{gray!6}{\parbox{\dimexpr\linewidth-2\fboxsep-2\fboxrule\relax}{\scriptsize
\textbf{Sample 26}\hfill \textcolor{black!45}{\textbf{Not selected}}\\
\textbf{Candidate \#5}\hfill \texttt{score=11.49}\\
\vspace{1pt}\\
\color{black!85} Well this is the big one. So big apparently, that I had to take it there and raise the number from 10 to 15. There’s just that many fails in the world of female rap. Some slight missteps, some EPIC. Nevertheless, they are all worth mentioning. You can probably think of a bunch more, but this is what I have gathered picking up from my prev\dots}}
\end{minipage}
\vspace{3pt}
\noindent
\begin{minipage}[t]{0.49\linewidth}
\fcolorbox{black!25}{gray!6}{\parbox{\dimexpr\linewidth-2\fboxsep-2\fboxrule\relax}{\scriptsize
\textbf{Sample 27}\hfill \textcolor{black!45}{\textbf{Not selected}}\\
\textbf{Candidate \#9}\hfill \texttt{score=11.29}\\
\vspace{1pt}\\
\color{black!85} “Last night three cargoes of Bohea Tea were emptied into the sea. This is the most magnificent movement of all. There is a dignity, a majesty, a sublimity, in this last effort of the Patriots that I greatly admire.” - John Adams, diary entry, December 17, 1773 - John Adams, diary entry, December 17, 1773 A Novel Idea Is something so new a\dots}}
\end{minipage}\hfill
\begin{minipage}[t]{0.49\linewidth}
\fcolorbox{black!25}{gray!6}{\parbox{\dimexpr\linewidth-2\fboxsep-2\fboxrule\relax}{\scriptsize
\textbf{Sample 28}\hfill \textcolor{black!45}{\textbf{Not selected}}\\
\textbf{Candidate \#2}\hfill \texttt{score=11.27}\\
\vspace{1pt}\\
\color{black!85} With the advent of new technologies for sneakers such as Vac Tech, Hyperfuse and Flyknit, the mid 90s and early 2000s methods of production and designing are becoming obsolete in this sneaker world. Nike Running is the future for Nike, generating billions of dollars per year, and we see Nike also not afraid to experiment with technology s\dots}}
\end{minipage}
\vspace{3pt}
\noindent
\begin{minipage}[t]{0.49\linewidth}
\fcolorbox{black!25}{gray!6}{\parbox{\dimexpr\linewidth-2\fboxsep-2\fboxrule\relax}{\scriptsize
\textbf{Sample 29}\hfill \textcolor{black!45}{\textbf{Not selected}}\\
\textbf{Candidate \#19}\hfill \texttt{score=11.04}\\
\vspace{1pt}\\
\color{black!85} Dividing Fractions Using Models Worksheet. This worksheet has six division with fractions issues to be solved — three must be solved with fashions and three with algorithms — options are on the second page. Answer key divide the unit fractions by whole numbers using th e fashions given. Use these resources to help reinforce the following\dots}}
\end{minipage}\hfill
\begin{minipage}[t]{0.49\linewidth}
\fcolorbox{black!25}{gray!6}{\parbox{\dimexpr\linewidth-2\fboxsep-2\fboxrule\relax}{\scriptsize
\textbf{Sample 30}\hfill \textcolor{black!45}{\textbf{Not selected}}\\
\textbf{Candidate \#12}\hfill \texttt{score=10.94}\\
\vspace{1pt}\\
\color{black!85} Can you please give us a little short bio? (education, professional experiences, select publications, academic specialty, awards won) Public school teacher for 5 years BA art (UC Irvine) PhD. (UCLA) educational psychology Professor of Child Development, (25 years) CSUS Senior Research Scientist (Oregon Research Institute with Institute of\dots}}
\end{minipage}
\vspace{3pt}
\noindent
\begin{minipage}[t]{0.49\linewidth}
\fcolorbox{black!25}{gray!6}{\parbox{\dimexpr\linewidth-2\fboxsep-2\fboxrule\relax}{\scriptsize
\textbf{Sample 31}\hfill \textcolor{black!45}{\textbf{Not selected}}\\
\textbf{Candidate \#1}\hfill \texttt{score=10.23}\\
\vspace{1pt}\\
\color{black!85} Wedding \& Party Venues - Sort By: Edgartown : (508) 627-9510 A 19th century gothic revival home transformed into the island's premier eco-boutique hotel. Guests either stay in the 17-room Hob Knob hotel or in the privacy of their own Hob Knob House. Guests can expect individualized Hob Knob hospitality and modern luxury amenities in a rel\dots}}
\end{minipage}\hfill
\begin{minipage}[t]{0.49\linewidth}
\fcolorbox{black!25}{gray!6}{\parbox{\dimexpr\linewidth-2\fboxsep-2\fboxrule\relax}{\scriptsize
\textbf{Sample 32}\hfill \textcolor{black!45}{\textbf{Not selected}}\\
\textbf{Candidate \#3}\hfill \texttt{score=9.77}\\
\vspace{1pt}\\
\color{black!85} starring John Travolta and Sam Jackson The first thing to understand about Basic --the basic thing, let's say-- is that although the commercials make it look like a war movie, it is not, for which we can all be grateful. No, Basic is a plot-twisty whodunnit. If The Usual Suspects died, and its body turned to cheese, and then that cheese-b\dots}}
\end{minipage}
\vspace{3pt}

\clearpage
\noindent\phantomsection
\label{app:qual_qurating}
\noindent \textbf{QuRating} \\[6pt]
\noindent
\begin{minipage}[t]{0.49\linewidth}
\fcolorbox{black!25}{gray!6}{\parbox{\dimexpr\linewidth-2\fboxsep-2\fboxrule\relax}{\scriptsize
\textbf{Sample 1}\hfill \textcolor{green!45!black}{\textbf{Selected}}\\
\textbf{Candidate \#26}\hfill \texttt{score=11.87}\\
\vspace{1pt}\\
\color{black!85} Unveiling the Power: Key Provisions of the Civil Rights Act of 1864 What were the Civil Rights Act of 1864's key provisions? The Civil Rights Act of 1864 was a pivotal moment in American history, establishing crucial legal protections for African Americans in the face of rampant discrimination. Editor Note: The Civil Rights Act of 1864 la\dots}}
\end{minipage}\hfill
\begin{minipage}[t]{0.49\linewidth}
\fcolorbox{black!25}{gray!6}{\parbox{\dimexpr\linewidth-2\fboxsep-2\fboxrule\relax}{\scriptsize
\textbf{Sample 2}\hfill \textcolor{green!45!black}{\textbf{Selected}}\\
\textbf{Candidate \#14}\hfill \texttt{score=10.85}\\
\vspace{1pt}\\
\color{black!85} Is your major sustainable enough? Whether you’re pursuing a sustainability degree and want to further your knowledge, or are interested in supplementing your major in another area with sustainability education, plenty of independent learning resources are available. A wide range of credit and noncredit courses—including university- and or\dots}}
\end{minipage}
\vspace{3pt}
\noindent
\begin{minipage}[t]{0.49\linewidth}
\fcolorbox{black!25}{gray!6}{\parbox{\dimexpr\linewidth-2\fboxsep-2\fboxrule\relax}{\scriptsize
\textbf{Sample 3}\hfill \textcolor{green!45!black}{\textbf{Selected}}\\
\textbf{Candidate \#22}\hfill \texttt{score=10.82}\\
\vspace{1pt}\\
\color{black!85} How To Choose Decodable Readers for First Grade To decode or not to decode: really, there is no question. To help rising first graders become successful and enthusiastic readers this summer, decodable readers are essential reading resources. Although “decodable text” might sound like yet another form of educational lingo, parents and educ\dots}}
\end{minipage}\hfill
\begin{minipage}[t]{0.49\linewidth}
\fcolorbox{black!25}{gray!6}{\parbox{\dimexpr\linewidth-2\fboxsep-2\fboxrule\relax}{\scriptsize
\textbf{Sample 4}\hfill \textcolor{green!45!black}{\textbf{Selected}}\\
\textbf{Candidate \#31}\hfill \texttt{score=10.42}\\
\vspace{1pt}\\
\color{black!85} One of the challenges of working with ancient DNA samples is that damage accumulates over time, breaking the double helix structure into ever-smaller fragments. In the samples we worked with, these fragments were scattered and mixed with contaminants, making genome reconstruction a major technical challenge. But a shocking paper published\dots}}
\end{minipage}
\vspace{3pt}
\noindent
\begin{minipage}[t]{0.49\linewidth}
\fcolorbox{black!25}{gray!6}{\parbox{\dimexpr\linewidth-2\fboxsep-2\fboxrule\relax}{\scriptsize
\textbf{Sample 5}\hfill \textcolor{green!45!black}{\textbf{Selected}}\\
\textbf{Candidate \#23}\hfill \texttt{score=10.27}\\
\vspace{1pt}\\
\color{black!85} The St. James kindergarteners have been working up to Project Week over the past month. We started slowly by taking walks in our neighborhood while Ms. Meghan and I noted what caught the children’s interest. It became apparent that the class was very interested in the L trains that they saw on our walks. It started with a simple question,\dots}}
\end{minipage}\hfill
\begin{minipage}[t]{0.49\linewidth}
\fcolorbox{black!25}{gray!6}{\parbox{\dimexpr\linewidth-2\fboxsep-2\fboxrule\relax}{\scriptsize
\textbf{Sample 6}\hfill \textcolor{green!45!black}{\textbf{Selected}}\\
\textbf{Candidate \#12}\hfill \texttt{score=10.05}\\
\vspace{1pt}\\
\color{black!85} Can you please give us a little short bio? (education, professional experiences, select publications, academic specialty, awards won) Public school teacher for 5 years BA art (UC Irvine) PhD. (UCLA) educational psychology Professor of Child Development, (25 years) CSUS Senior Research Scientist (Oregon Research Institute with Institute of\dots}}
\end{minipage}
\vspace{3pt}
\noindent
\begin{minipage}[t]{0.49\linewidth}
\fcolorbox{black!25}{gray!6}{\parbox{\dimexpr\linewidth-2\fboxsep-2\fboxrule\relax}{\scriptsize
\textbf{Sample 7}\hfill \textcolor{green!45!black}{\textbf{Selected}}\\
\textbf{Candidate \#29}\hfill \texttt{score=9.76}\\
\vspace{1pt}\\
\color{black!85} Earthquakes are the result of sudden movement along faults within the Earth. The movement releases stored-up ‘elastic strain’ energy in the form of seismic waves, which propagate through the Earth and cause the ground surface to shake. Such movement on the faults is generally a response to long-term deformation and the buildup of stress.\dots}}
\end{minipage}\hfill
\begin{minipage}[t]{0.49\linewidth}
\fcolorbox{black!25}{gray!6}{\parbox{\dimexpr\linewidth-2\fboxsep-2\fboxrule\relax}{\scriptsize
\textbf{Sample 8}\hfill \textcolor{green!45!black}{\textbf{Selected}}\\
\textbf{Candidate \#20}\hfill \texttt{score=9.62}\\
\vspace{1pt}\\
\color{black!85} Conduct Disorder (CD) is a complex and serious behavioural and emotional disorder that can occur in children and adolescents. It’s characterised by a repetitive and persistent pattern of behaviour where the basic rights of others or major age-appropriate societal norms or rules are violated. Here’s an outline of Conduct Disorder in line w\dots}}
\end{minipage}
\vspace{3pt}
\noindent
\begin{minipage}[t]{0.49\linewidth}
\fcolorbox{black!25}{gray!6}{\parbox{\dimexpr\linewidth-2\fboxsep-2\fboxrule\relax}{\scriptsize
\textbf{Sample 9}\hfill \textcolor{green!45!black}{\textbf{Selected}}\\
\textbf{Candidate \#25}\hfill \texttt{score=8.96}\\
\vspace{1pt}\\
\color{black!85} KS2 Maths is an important core subject in the National Curriculum and this area of the website covers all the major aspects of the curriculum including numbers, calculations, problems and measures. Each subject area is designed to help children develop their knowledge, whether they are learning in a classroom or home schooling environment\dots}}
\end{minipage}\hfill
\begin{minipage}[t]{0.49\linewidth}
\fcolorbox{black!25}{gray!6}{\parbox{\dimexpr\linewidth-2\fboxsep-2\fboxrule\relax}{\scriptsize
\textbf{Sample 10}\hfill \textcolor{green!45!black}{\textbf{Selected}}\\
\textbf{Candidate \#30}\hfill \texttt{score=8.95}\\
\vspace{1pt}\\
\color{black!85} Over 1.8 million professionals use CFI to learn accounting, financial analysis, modeling and more. Start with a free account to explore 20+ always-free courses and hundreds of finance templates and cheat sheets. What is the Central Limit Theorem (CLT)? The Central Limit Theorem (CLT) is a statistical concept that states that the sample me\dots}}
\end{minipage}
\vspace{3pt}
\noindent
\begin{minipage}[t]{0.49\linewidth}
\fcolorbox{black!25}{gray!6}{\parbox{\dimexpr\linewidth-2\fboxsep-2\fboxrule\relax}{\scriptsize
\textbf{Sample 11}\hfill \textcolor{green!45!black}{\textbf{Selected}}\\
\textbf{Candidate \#15}\hfill \texttt{score=8.39}\\
\vspace{1pt}\\
\color{black!85} Origami is an art form that combines precision, creativity, and patience. While basic origami is obtainable to every one, mastering complex origami designs can be quite a rewarding and impressive achievement. In this article, we’ll show you with the procedure for creating intricate origami while highlighting essential techniques for achie\dots}}
\end{minipage}\hfill
\begin{minipage}[t]{0.49\linewidth}
\fcolorbox{black!25}{gray!6}{\parbox{\dimexpr\linewidth-2\fboxsep-2\fboxrule\relax}{\scriptsize
\textbf{Sample 12}\hfill \textcolor{green!45!black}{\textbf{Selected}}\\
\textbf{Candidate \#28}\hfill \texttt{score=8.17}\\
\vspace{1pt}\\
\color{black!85} You really have to be alert when studying science. Galaxies were created after matter. The stars in those galaxies were supposed to move slowly because there was more mass in the center of the galaxy. However, after dark matter was added, the stars appeared to move faster; however, this is not the case in our galaxy, suggesting that there\dots}}
\end{minipage}
\vspace{3pt}
\noindent
\begin{minipage}[t]{0.49\linewidth}
\fcolorbox{black!25}{gray!6}{\parbox{\dimexpr\linewidth-2\fboxsep-2\fboxrule\relax}{\scriptsize
\textbf{Sample 13}\hfill \textcolor{green!45!black}{\textbf{Selected}}\\
\textbf{Candidate \#24}\hfill \texttt{score=8.12}\\
\vspace{1pt}\\
\color{black!85} Next we will talk about solar radiation, that is, the forms of solar radiation that we receive on earth. Solar radiation is generated by a series of nuclear fusion reactions that occur in the Sun and, as a consequence, emit electromagnetic radiation that reaches the earth. This radiation received by the earth’s surface is measured in W /\dots}}
\end{minipage}\hfill
\begin{minipage}[t]{0.49\linewidth}
\fcolorbox{black!25}{gray!6}{\parbox{\dimexpr\linewidth-2\fboxsep-2\fboxrule\relax}{\scriptsize
\textbf{Sample 14}\hfill \textcolor{green!45!black}{\textbf{Selected}}\\
\textbf{Candidate \#16}\hfill \texttt{score=8.04}\\
\vspace{1pt}\\
\color{black!85} What is rotavirus and why does my baby need to be immunised? Rotavirus is a very infectious virus that causes the majority of serious cases of gastroenteritis in babies. It causes diarrhoea, vomiting and abdominal pain, usually lasting around a week. Most children will be infected by rotavirus once by the age of five. Gastroenteritis (cau\dots}}
\end{minipage}
\vspace{3pt}
\noindent
\begin{minipage}[t]{0.49\linewidth}
\fcolorbox{black!25}{gray!6}{\parbox{\dimexpr\linewidth-2\fboxsep-2\fboxrule\relax}{\scriptsize
\textbf{Sample 15}\hfill \textcolor{green!45!black}{\textbf{Selected}}\\
\textbf{Candidate \#21}\hfill \texttt{score=8.02}\\
\vspace{1pt}\\
\color{black!85} Nestled in the leafy suburbs of western Berlin, the Wannsee Conference House stands as a poignant reminder of a dark chapter in human history. The Wannsee Conference: A Pivotal Moment The Wannsee Conference, held on January 20, 1942, marked a pivotal moment in the implementation of Nazi Germany's genocidal plans. Organized by SS-Obergrupp\dots}}
\end{minipage}\hfill
\begin{minipage}[t]{0.49\linewidth}
\fcolorbox{black!25}{gray!6}{\parbox{\dimexpr\linewidth-2\fboxsep-2\fboxrule\relax}{\scriptsize
\textbf{Sample 16}\hfill \textcolor{green!45!black}{\textbf{Selected}}\\
\textbf{Candidate \#11}\hfill \texttt{score=7.76}\\
\vspace{1pt}\\
\color{black!85} In decades past, classroom design was often an afterthought and followed a standardised layout. Plain boxed shaped classrooms, with identical chairs and tables throughout were commonplace in many schools. Read the latest issue of School News HERE Recently, though, there has been a shift away from this one-size-fits all approach to classro\dots}}
\end{minipage}
\vspace{3pt}
\clearpage
\noindent\phantomsection
\label{app:qual_qurating_cont}
% \noindent \textbf{QuRating} \\[6pt]
\noindent
\begin{minipage}[t]{0.49\linewidth}
\fcolorbox{black!25}{gray!6}{\parbox{\dimexpr\linewidth-2\fboxsep-2\fboxrule\relax}{\scriptsize
\textbf{Sample 17}\hfill \textcolor{black!45}{\textbf{Not selected}}\\
\textbf{Candidate \#18}\hfill \texttt{score=7.47}\\
\vspace{1pt}\\
\color{black!85} This article originally appeared in the December 2015 issue of Resource Recycling. Subscribe today for access to all print content. Since the 1990s, curbside and drop-off recycling has grown substantially – nearly 90 percent of households now have access, according to recent surveys from Moore Recycling Associates, the American Forest and\dots}}
\end{minipage}\hfill
\begin{minipage}[t]{0.49\linewidth}
\fcolorbox{black!25}{gray!6}{\parbox{\dimexpr\linewidth-2\fboxsep-2\fboxrule\relax}{\scriptsize
\textbf{Sample 18}\hfill \textcolor{black!45}{\textbf{Not selected}}\\
\textbf{Candidate \#8}\hfill \texttt{score=7.46}\\
\vspace{1pt}\\
\color{black!85} The Unsung Heroes of Your HVAC System: Understanding the Importance of Filters When it comes to your HVAC (Heating, Ventilation, and Air Conditioning) system, you might be quick to think about the thermostat, air ducts, or even the unit itself. However, there’s an unsung hero in your HVAC system that plays a pivotal role in maintaining in\dots}}
\end{minipage}
\vspace{3pt}
\noindent
\begin{minipage}[t]{0.49\linewidth}
\fcolorbox{black!25}{gray!6}{\parbox{\dimexpr\linewidth-2\fboxsep-2\fboxrule\relax}{\scriptsize
\textbf{Sample 19}\hfill \textcolor{black!45}{\textbf{Not selected}}\\
\textbf{Candidate \#10}\hfill \texttt{score=7.21}\\
\vspace{1pt}\\
\color{black!85} Deforestation isn't just happening in well-known global hotspots like Indonesia and Brazil's rainforest. A new analysis says forests are also shrinking on state and private land in Oregon, where an estimated 522,000 acres of forest cover have disappeared since 2000. That's an area six times larger than the city of Portland, equal to more\dots}}
\end{minipage}\hfill
\begin{minipage}[t]{0.49\linewidth}
\fcolorbox{black!25}{gray!6}{\parbox{\dimexpr\linewidth-2\fboxsep-2\fboxrule\relax}{\scriptsize
\textbf{Sample 20}\hfill \textcolor{black!45}{\textbf{Not selected}}\\
\textbf{Candidate \#27}\hfill \texttt{score=7.05}\\
\vspace{1pt}\\
\color{black!85} 24/7 writing help on your phone Save to my list Remove from my list In the tumultuous 19th century, both Italy and Germany found themselves fragmented into numerous separate ruling states. The impetus for change came in the form of rising nationalism and liberalism, paving the way for the unification of these disparate entities. However,\dots}}
\end{minipage}
\vspace{3pt}
\noindent
\begin{minipage}[t]{0.49\linewidth}
\fcolorbox{black!25}{gray!6}{\parbox{\dimexpr\linewidth-2\fboxsep-2\fboxrule\relax}{\scriptsize
\textbf{Sample 21}\hfill \textcolor{black!45}{\textbf{Not selected}}\\
\textbf{Candidate \#19}\hfill \texttt{score=5.30}\\
\vspace{1pt}\\
\color{black!85} Dividing Fractions Using Models Worksheet. This worksheet has six division with fractions issues to be solved — three must be solved with fashions and three with algorithms — options are on the second page. Answer key divide the unit fractions by whole numbers using th e fashions given. Use these resources to help reinforce the following\dots}}
\end{minipage}\hfill
\begin{minipage}[t]{0.49\linewidth}
\fcolorbox{black!25}{gray!6}{\parbox{\dimexpr\linewidth-2\fboxsep-2\fboxrule\relax}{\scriptsize
\textbf{Sample 22}\hfill \textcolor{black!45}{\textbf{Not selected}}\\
\textbf{Candidate \#17}\hfill \texttt{score=5.29}\\
\vspace{1pt}\\
\color{black!85} Political Parties and Elections Political parties are an established part of modern mass democracy, and the conduct of elections in India is largely dependent on the behaviour of political parties. Although many candidates for Indian elections are independent, the winning candidates for Lok Sabha and Vidhan Sabha elections usually stand a\dots}}
\end{minipage}
\vspace{3pt}
\noindent
\begin{minipage}[t]{0.49\linewidth}
\fcolorbox{black!25}{gray!6}{\parbox{\dimexpr\linewidth-2\fboxsep-2\fboxrule\relax}{\scriptsize
\textbf{Sample 23}\hfill \textcolor{black!45}{\textbf{Not selected}}\\
\textbf{Candidate \#9}\hfill \texttt{score=4.21}\\
\vspace{1pt}\\
\color{black!85} “Last night three cargoes of Bohea Tea were emptied into the sea. This is the most magnificent movement of all. There is a dignity, a majesty, a sublimity, in this last effort of the Patriots that I greatly admire.” - John Adams, diary entry, December 17, 1773 - John Adams, diary entry, December 17, 1773 A Novel Idea Is something so new a\dots}}
\end{minipage}\hfill
\begin{minipage}[t]{0.49\linewidth}
\fcolorbox{black!25}{gray!6}{\parbox{\dimexpr\linewidth-2\fboxsep-2\fboxrule\relax}{\scriptsize
\textbf{Sample 24}\hfill \textcolor{black!45}{\textbf{Not selected}}\\
\textbf{Candidate \#0}\hfill \texttt{score=3.94}\\
\vspace{1pt}\\
\color{black!85} As it turns out, the exercises synonymous with strong, attractive abs may not be the best way to train your core—and may be doing damage to your back. Read more If you are worried about the excess holiday pounds many of us are still carrying around. There are a few easy, natural things you can do to shed them, and none of them requires an\dots}}
\end{minipage}
\vspace{3pt}
\noindent
\begin{minipage}[t]{0.49\linewidth}
\fcolorbox{black!25}{gray!6}{\parbox{\dimexpr\linewidth-2\fboxsep-2\fboxrule\relax}{\scriptsize
\textbf{Sample 25}\hfill \textcolor{black!45}{\textbf{Not selected}}\\
\textbf{Candidate \#6}\hfill \texttt{score=2.82}\\
\vspace{1pt}\\
\color{black!85} Skaters need to check their skate helmets every so often and ask yourself, "Is it time to replace this helmet?" Well, that depends. Did you crash in it? For starters, most people are aware that you must replace a helmet after any crash where your head hit. The foam part of a helmet is made for one-time use, and after crushing once it is n\dots}}
\end{minipage}\hfill
\begin{minipage}[t]{0.49\linewidth}
\fcolorbox{black!25}{gray!6}{\parbox{\dimexpr\linewidth-2\fboxsep-2\fboxrule\relax}{\scriptsize
\textbf{Sample 26}\hfill \textcolor{black!45}{\textbf{Not selected}}\\
\textbf{Candidate \#13}\hfill \texttt{score=2.34}\\
\vspace{1pt}\\
\color{black!85} In Heart of Darkness it is the white invaders for instance, who are, almost without exception, embodiments of blindness, selfishness, and cruelty; and even in the cognitive domain, where such positive phrases as “to enlighten,” for instance, are conventionally opposed to negative ones such as “to be in the dark,” the traditional expectati\dots}}
\end{minipage}
\vspace{3pt}
\noindent
\begin{minipage}[t]{0.49\linewidth}
\fcolorbox{black!25}{gray!6}{\parbox{\dimexpr\linewidth-2\fboxsep-2\fboxrule\relax}{\scriptsize
\textbf{Sample 27}\hfill \textcolor{black!45}{\textbf{Not selected}}\\
\textbf{Candidate \#7}\hfill \texttt{score=1.46}\\
\vspace{1pt}\\
\color{black!85} Elizabeth Hurley played as Dalila Release: Dec 8, 1996 Mara and her husband Manoa are both upstanding and religious Israelites living under the harsh and unjust rule of the Philistines. Much to their regret, they have not been able to have children. One day, a mysterious stranger appears to Mara and promises her that she will bear a son w\dots}}
\end{minipage}\hfill
\begin{minipage}[t]{0.49\linewidth}
\fcolorbox{black!25}{gray!6}{\parbox{\dimexpr\linewidth-2\fboxsep-2\fboxrule\relax}{\scriptsize
\textbf{Sample 28}\hfill \textcolor{black!45}{\textbf{Not selected}}\\
\textbf{Candidate \#2}\hfill \texttt{score=0.321}\\
\vspace{1pt}\\
\color{black!85} With the advent of new technologies for sneakers such as Vac Tech, Hyperfuse and Flyknit, the mid 90s and early 2000s methods of production and designing are becoming obsolete in this sneaker world. Nike Running is the future for Nike, generating billions of dollars per year, and we see Nike also not afraid to experiment with technology s\dots}}
\end{minipage}
\vspace{3pt}
\noindent
\begin{minipage}[t]{0.49\linewidth}
\fcolorbox{black!25}{gray!6}{\parbox{\dimexpr\linewidth-2\fboxsep-2\fboxrule\relax}{\scriptsize
\textbf{Sample 29}\hfill \textcolor{black!45}{\textbf{Not selected}}\\
\textbf{Candidate \#1}\hfill \texttt{score=-0.429}\\
\vspace{1pt}\\
\color{black!85} Wedding \& Party Venues - Sort By: Edgartown : (508) 627-9510 A 19th century gothic revival home transformed into the island's premier eco-boutique hotel. Guests either stay in the 17-room Hob Knob hotel or in the privacy of their own Hob Knob House. Guests can expect individualized Hob Knob hospitality and modern luxury amenities in a rel\dots}}
\end{minipage}\hfill
\begin{minipage}[t]{0.49\linewidth}
\fcolorbox{black!25}{gray!6}{\parbox{\dimexpr\linewidth-2\fboxsep-2\fboxrule\relax}{\scriptsize
\textbf{Sample 30}\hfill \textcolor{black!45}{\textbf{Not selected}}\\
\textbf{Candidate \#3}\hfill \texttt{score=-2.09}\\
\vspace{1pt}\\
\color{black!85} starring John Travolta and Sam Jackson The first thing to understand about Basic --the basic thing, let's say-- is that although the commercials make it look like a war movie, it is not, for which we can all be grateful. No, Basic is a plot-twisty whodunnit. If The Usual Suspects died, and its body turned to cheese, and then that cheese-b\dots}}
\end{minipage}
\vspace{3pt}
\noindent
\begin{minipage}[t]{0.49\linewidth}
\fcolorbox{black!25}{gray!6}{\parbox{\dimexpr\linewidth-2\fboxsep-2\fboxrule\relax}{\scriptsize
\textbf{Sample 31}\hfill \textcolor{black!45}{\textbf{Not selected}}\\
\textbf{Candidate \#4}\hfill \texttt{score=-2.84}\\
\vspace{1pt}\\
\color{black!85} 5 Types of Women’s Underwear That Men Love Underwear can say a lot about a woman. It’s something that men are obsessed with, to the point that, a mere glimpse of a thong waistband causes us to go into shock. On the surface we find them sexy, revealing. We’re able to see who a woman actually is—or maybe some guys are just plain horny. Howe\dots}}
\end{minipage}\hfill
\begin{minipage}[t]{0.49\linewidth}
\fcolorbox{black!25}{gray!6}{\parbox{\dimexpr\linewidth-2\fboxsep-2\fboxrule\relax}{\scriptsize
\textbf{Sample 32}\hfill \textcolor{black!45}{\textbf{Not selected}}\\
\textbf{Candidate \#5}\hfill \texttt{score=-4.08}\\
\vspace{1pt}\\
\color{black!85} Well this is the big one. So big apparently, that I had to take it there and raise the number from 10 to 15. There’s just that many fails in the world of female rap. Some slight missteps, some EPIC. Nevertheless, they are all worth mentioning. You can probably think of a bunch more, but this is what I have gathered picking up from my prev\dots}}
\end{minipage}
\vspace{3pt}

\clearpage
\noindent\phantomsection
\label{app:qual_fineweb_edu}
\noindent \textbf{FineWeb-Edu} \\[6pt]
\noindent
\begin{minipage}[t]{0.49\linewidth}
\fcolorbox{black!25}{gray!6}{\parbox{\dimexpr\linewidth-2\fboxsep-2\fboxrule\relax}{\scriptsize
\textbf{Sample 1}\hfill \textcolor{green!45!black}{\textbf{Selected}}\\
\textbf{Candidate \#28}\hfill \texttt{score=4.62}\\
\vspace{1pt}\\
\color{black!85} You really have to be alert when studying science. Galaxies were created after matter. The stars in those galaxies were supposed to move slowly because there was more mass in the center of the galaxy. However, after dark matter was added, the stars appeared to move faster; however, this is not the case in our galaxy, suggesting that there\dots}}
\end{minipage}\hfill
\begin{minipage}[t]{0.49\linewidth}
\fcolorbox{black!25}{gray!6}{\parbox{\dimexpr\linewidth-2\fboxsep-2\fboxrule\relax}{\scriptsize
\textbf{Sample 2}\hfill \textcolor{green!45!black}{\textbf{Selected}}\\
\textbf{Candidate \#25}\hfill \texttt{score=4.61}\\
\vspace{1pt}\\
\color{black!85} KS2 Maths is an important core subject in the National Curriculum and this area of the website covers all the major aspects of the curriculum including numbers, calculations, problems and measures. Each subject area is designed to help children develop their knowledge, whether they are learning in a classroom or home schooling environment\dots}}
\end{minipage}
\vspace{3pt}
\noindent
\begin{minipage}[t]{0.49\linewidth}
\fcolorbox{black!25}{gray!6}{\parbox{\dimexpr\linewidth-2\fboxsep-2\fboxrule\relax}{\scriptsize
\textbf{Sample 3}\hfill \textcolor{green!45!black}{\textbf{Selected}}\\
\textbf{Candidate \#29}\hfill \texttt{score=4.61}\\
\vspace{1pt}\\
\color{black!85} Earthquakes are the result of sudden movement along faults within the Earth. The movement releases stored-up ‘elastic strain’ energy in the form of seismic waves, which propagate through the Earth and cause the ground surface to shake. Such movement on the faults is generally a response to long-term deformation and the buildup of stress.\dots}}
\end{minipage}\hfill
\begin{minipage}[t]{0.49\linewidth}
\fcolorbox{black!25}{gray!6}{\parbox{\dimexpr\linewidth-2\fboxsep-2\fboxrule\relax}{\scriptsize
\textbf{Sample 4}\hfill \textcolor{green!45!black}{\textbf{Selected}}\\
\textbf{Candidate \#31}\hfill \texttt{score=4.57}\\
\vspace{1pt}\\
\color{black!85} One of the challenges of working with ancient DNA samples is that damage accumulates over time, breaking the double helix structure into ever-smaller fragments. In the samples we worked with, these fragments were scattered and mixed with contaminants, making genome reconstruction a major technical challenge. But a shocking paper published\dots}}
\end{minipage}
\vspace{3pt}
\noindent
\begin{minipage}[t]{0.49\linewidth}
\fcolorbox{black!25}{gray!6}{\parbox{\dimexpr\linewidth-2\fboxsep-2\fboxrule\relax}{\scriptsize
\textbf{Sample 5}\hfill \textcolor{green!45!black}{\textbf{Selected}}\\
\textbf{Candidate \#24}\hfill \texttt{score=4.54}\\
\vspace{1pt}\\
\color{black!85} Next we will talk about solar radiation, that is, the forms of solar radiation that we receive on earth. Solar radiation is generated by a series of nuclear fusion reactions that occur in the Sun and, as a consequence, emit electromagnetic radiation that reaches the earth. This radiation received by the earth’s surface is measured in W /\dots}}
\end{minipage}\hfill
\begin{minipage}[t]{0.49\linewidth}
\fcolorbox{black!25}{gray!6}{\parbox{\dimexpr\linewidth-2\fboxsep-2\fboxrule\relax}{\scriptsize
\textbf{Sample 6}\hfill \textcolor{green!45!black}{\textbf{Selected}}\\
\textbf{Candidate \#26}\hfill \texttt{score=4.53}\\
\vspace{1pt}\\
\color{black!85} Unveiling the Power: Key Provisions of the Civil Rights Act of 1864 What were the Civil Rights Act of 1864's key provisions? The Civil Rights Act of 1864 was a pivotal moment in American history, establishing crucial legal protections for African Americans in the face of rampant discrimination. Editor Note: The Civil Rights Act of 1864 la\dots}}
\end{minipage}
\vspace{3pt}
\noindent
\begin{minipage}[t]{0.49\linewidth}
\fcolorbox{black!25}{gray!6}{\parbox{\dimexpr\linewidth-2\fboxsep-2\fboxrule\relax}{\scriptsize
\textbf{Sample 7}\hfill \textcolor{green!45!black}{\textbf{Selected}}\\
\textbf{Candidate \#27}\hfill \texttt{score=4.53}\\
\vspace{1pt}\\
\color{black!85} 24/7 writing help on your phone Save to my list Remove from my list In the tumultuous 19th century, both Italy and Germany found themselves fragmented into numerous separate ruling states. The impetus for change came in the form of rising nationalism and liberalism, paving the way for the unification of these disparate entities. However,\dots}}
\end{minipage}\hfill
\begin{minipage}[t]{0.49\linewidth}
\fcolorbox{black!25}{gray!6}{\parbox{\dimexpr\linewidth-2\fboxsep-2\fboxrule\relax}{\scriptsize
\textbf{Sample 8}\hfill \textcolor{green!45!black}{\textbf{Selected}}\\
\textbf{Candidate \#30}\hfill \texttt{score=4.50}\\
\vspace{1pt}\\
\color{black!85} Over 1.8 million professionals use CFI to learn accounting, financial analysis, modeling and more. Start with a free account to explore 20+ always-free courses and hundreds of finance templates and cheat sheets. What is the Central Limit Theorem (CLT)? The Central Limit Theorem (CLT) is a statistical concept that states that the sample me\dots}}
\end{minipage}
\vspace{3pt}
\noindent
\begin{minipage}[t]{0.49\linewidth}
\fcolorbox{black!25}{gray!6}{\parbox{\dimexpr\linewidth-2\fboxsep-2\fboxrule\relax}{\scriptsize
\textbf{Sample 9}\hfill \textcolor{green!45!black}{\textbf{Selected}}\\
\textbf{Candidate \#20}\hfill \texttt{score=4.18}\\
\vspace{1pt}\\
\color{black!85} Conduct Disorder (CD) is a complex and serious behavioural and emotional disorder that can occur in children and adolescents. It’s characterised by a repetitive and persistent pattern of behaviour where the basic rights of others or major age-appropriate societal norms or rules are violated. Here’s an outline of Conduct Disorder in line w\dots}}
\end{minipage}\hfill
\begin{minipage}[t]{0.49\linewidth}
\fcolorbox{black!25}{gray!6}{\parbox{\dimexpr\linewidth-2\fboxsep-2\fboxrule\relax}{\scriptsize
\textbf{Sample 10}\hfill \textcolor{green!45!black}{\textbf{Selected}}\\
\textbf{Candidate \#19}\hfill \texttt{score=4.08}\\
\vspace{1pt}\\
\color{black!85} Dividing Fractions Using Models Worksheet. This worksheet has six division with fractions issues to be solved — three must be solved with fashions and three with algorithms — options are on the second page. Answer key divide the unit fractions by whole numbers using th e fashions given. Use these resources to help reinforce the following\dots}}
\end{minipage}
\vspace{3pt}
\noindent
\begin{minipage}[t]{0.49\linewidth}
\fcolorbox{black!25}{gray!6}{\parbox{\dimexpr\linewidth-2\fboxsep-2\fboxrule\relax}{\scriptsize
\textbf{Sample 11}\hfill \textcolor{green!45!black}{\textbf{Selected}}\\
\textbf{Candidate \#21}\hfill \texttt{score=3.96}\\
\vspace{1pt}\\
\color{black!85} Nestled in the leafy suburbs of western Berlin, the Wannsee Conference House stands as a poignant reminder of a dark chapter in human history. The Wannsee Conference: A Pivotal Moment The Wannsee Conference, held on January 20, 1942, marked a pivotal moment in the implementation of Nazi Germany's genocidal plans. Organized by SS-Obergrupp\dots}}
\end{minipage}\hfill
\begin{minipage}[t]{0.49\linewidth}
\fcolorbox{black!25}{gray!6}{\parbox{\dimexpr\linewidth-2\fboxsep-2\fboxrule\relax}{\scriptsize
\textbf{Sample 12}\hfill \textcolor{green!45!black}{\textbf{Selected}}\\
\textbf{Candidate \#22}\hfill \texttt{score=3.92}\\
\vspace{1pt}\\
\color{black!85} How To Choose Decodable Readers for First Grade To decode or not to decode: really, there is no question. To help rising first graders become successful and enthusiastic readers this summer, decodable readers are essential reading resources. Although “decodable text” might sound like yet another form of educational lingo, parents and educ\dots}}
\end{minipage}
\vspace{3pt}
\noindent
\begin{minipage}[t]{0.49\linewidth}
\fcolorbox{black!25}{gray!6}{\parbox{\dimexpr\linewidth-2\fboxsep-2\fboxrule\relax}{\scriptsize
\textbf{Sample 13}\hfill \textcolor{green!45!black}{\textbf{Selected}}\\
\textbf{Candidate \#17}\hfill \texttt{score=3.85}\\
\vspace{1pt}\\
\color{black!85} Political Parties and Elections Political parties are an established part of modern mass democracy, and the conduct of elections in India is largely dependent on the behaviour of political parties. Although many candidates for Indian elections are independent, the winning candidates for Lok Sabha and Vidhan Sabha elections usually stand a\dots}}
\end{minipage}\hfill
\begin{minipage}[t]{0.49\linewidth}
\fcolorbox{black!25}{gray!6}{\parbox{\dimexpr\linewidth-2\fboxsep-2\fboxrule\relax}{\scriptsize
\textbf{Sample 14}\hfill \textcolor{green!45!black}{\textbf{Selected}}\\
\textbf{Candidate \#23}\hfill \texttt{score=3.73}\\
\vspace{1pt}\\
\color{black!85} The St. James kindergarteners have been working up to Project Week over the past month. We started slowly by taking walks in our neighborhood while Ms. Meghan and I noted what caught the children’s interest. It became apparent that the class was very interested in the L trains that they saw on our walks. It started with a simple question,\dots}}
\end{minipage}
\vspace{3pt}
\noindent
\begin{minipage}[t]{0.49\linewidth}
\fcolorbox{black!25}{gray!6}{\parbox{\dimexpr\linewidth-2\fboxsep-2\fboxrule\relax}{\scriptsize
\textbf{Sample 15}\hfill \textcolor{green!45!black}{\textbf{Selected}}\\
\textbf{Candidate \#16}\hfill \texttt{score=3.63}\\
\vspace{1pt}\\
\color{black!85} What is rotavirus and why does my baby need to be immunised? Rotavirus is a very infectious virus that causes the majority of serious cases of gastroenteritis in babies. It causes diarrhoea, vomiting and abdominal pain, usually lasting around a week. Most children will be infected by rotavirus once by the age of five. Gastroenteritis (cau\dots}}
\end{minipage}\hfill
\begin{minipage}[t]{0.49\linewidth}
\fcolorbox{black!25}{gray!6}{\parbox{\dimexpr\linewidth-2\fboxsep-2\fboxrule\relax}{\scriptsize
\textbf{Sample 16}\hfill \textcolor{green!45!black}{\textbf{Selected}}\\
\textbf{Candidate \#18}\hfill \texttt{score=3.56}\\
\vspace{1pt}\\
\color{black!85} This article originally appeared in the December 2015 issue of Resource Recycling. Subscribe today for access to all print content. Since the 1990s, curbside and drop-off recycling has grown substantially – nearly 90 percent of households now have access, according to recent surveys from Moore Recycling Associates, the American Forest and\dots}}
\end{minipage}
\vspace{3pt}
\clearpage
\noindent\phantomsection
\label{app:qual_fineweb_edu_cont}
% \noindent \textbf{FineWeb-Edu classifier} \\[6pt]
\noindent
\begin{minipage}[t]{0.49\linewidth}
\fcolorbox{black!25}{gray!6}{\parbox{\dimexpr\linewidth-2\fboxsep-2\fboxrule\relax}{\scriptsize
\textbf{Sample 17}\hfill \textcolor{black!45}{\textbf{Not selected}}\\
\textbf{Candidate \#10}\hfill \texttt{score=3.30}\\
\vspace{1pt}\\
\color{black!85} Deforestation isn't just happening in well-known global hotspots like Indonesia and Brazil's rainforest. A new analysis says forests are also shrinking on state and private land in Oregon, where an estimated 522,000 acres of forest cover have disappeared since 2000. That's an area six times larger than the city of Portland, equal to more\dots}}
\end{minipage}\hfill
\begin{minipage}[t]{0.49\linewidth}
\fcolorbox{black!25}{gray!6}{\parbox{\dimexpr\linewidth-2\fboxsep-2\fboxrule\relax}{\scriptsize
\textbf{Sample 18}\hfill \textcolor{black!45}{\textbf{Not selected}}\\
\textbf{Candidate \#9}\hfill \texttt{score=3.30}\\
\vspace{1pt}\\
\color{black!85} “Last night three cargoes of Bohea Tea were emptied into the sea. This is the most magnificent movement of all. There is a dignity, a majesty, a sublimity, in this last effort of the Patriots that I greatly admire.” - John Adams, diary entry, December 17, 1773 - John Adams, diary entry, December 17, 1773 A Novel Idea Is something so new a\dots}}
\end{minipage}
\vspace{3pt}
\noindent
\begin{minipage}[t]{0.49\linewidth}
\fcolorbox{black!25}{gray!6}{\parbox{\dimexpr\linewidth-2\fboxsep-2\fboxrule\relax}{\scriptsize
\textbf{Sample 19}\hfill \textcolor{black!45}{\textbf{Not selected}}\\
\textbf{Candidate \#11}\hfill \texttt{score=2.95}\\
\vspace{1pt}\\
\color{black!85} In decades past, classroom design was often an afterthought and followed a standardised layout. Plain boxed shaped classrooms, with identical chairs and tables throughout were commonplace in many schools. Read the latest issue of School News HERE Recently, though, there has been a shift away from this one-size-fits all approach to classro\dots}}
\end{minipage}\hfill
\begin{minipage}[t]{0.49\linewidth}
\fcolorbox{black!25}{gray!6}{\parbox{\dimexpr\linewidth-2\fboxsep-2\fboxrule\relax}{\scriptsize
\textbf{Sample 20}\hfill \textcolor{black!45}{\textbf{Not selected}}\\
\textbf{Candidate \#15}\hfill \texttt{score=2.93}\\
\vspace{1pt}\\
\color{black!85} Origami is an art form that combines precision, creativity, and patience. While basic origami is obtainable to every one, mastering complex origami designs can be quite a rewarding and impressive achievement. In this article, we’ll show you with the procedure for creating intricate origami while highlighting essential techniques for achie\dots}}
\end{minipage}
\vspace{3pt}
\noindent
\begin{minipage}[t]{0.49\linewidth}
\fcolorbox{black!25}{gray!6}{\parbox{\dimexpr\linewidth-2\fboxsep-2\fboxrule\relax}{\scriptsize
\textbf{Sample 21}\hfill \textcolor{black!45}{\textbf{Not selected}}\\
\textbf{Candidate \#13}\hfill \texttt{score=2.86}\\
\vspace{1pt}\\
\color{black!85} In Heart of Darkness it is the white invaders for instance, who are, almost without exception, embodiments of blindness, selfishness, and cruelty; and even in the cognitive domain, where such positive phrases as “to enlighten,” for instance, are conventionally opposed to negative ones such as “to be in the dark,” the traditional expectati\dots}}
\end{minipage}\hfill
\begin{minipage}[t]{0.49\linewidth}
\fcolorbox{black!25}{gray!6}{\parbox{\dimexpr\linewidth-2\fboxsep-2\fboxrule\relax}{\scriptsize
\textbf{Sample 22}\hfill \textcolor{black!45}{\textbf{Not selected}}\\
\textbf{Candidate \#8}\hfill \texttt{score=2.83}\\
\vspace{1pt}\\
\color{black!85} The Unsung Heroes of Your HVAC System: Understanding the Importance of Filters When it comes to your HVAC (Heating, Ventilation, and Air Conditioning) system, you might be quick to think about the thermostat, air ducts, or even the unit itself. However, there’s an unsung hero in your HVAC system that plays a pivotal role in maintaining in\dots}}
\end{minipage}
\vspace{3pt}
\noindent
\begin{minipage}[t]{0.49\linewidth}
\fcolorbox{black!25}{gray!6}{\parbox{\dimexpr\linewidth-2\fboxsep-2\fboxrule\relax}{\scriptsize
\textbf{Sample 23}\hfill \textcolor{black!45}{\textbf{Not selected}}\\
\textbf{Candidate \#12}\hfill \texttt{score=2.72}\\
\vspace{1pt}\\
\color{black!85} Can you please give us a little short bio? (education, professional experiences, select publications, academic specialty, awards won) Public school teacher for 5 years BA art (UC Irvine) PhD. (UCLA) educational psychology Professor of Child Development, (25 years) CSUS Senior Research Scientist (Oregon Research Institute with Institute of\dots}}
\end{minipage}\hfill
\begin{minipage}[t]{0.49\linewidth}
\fcolorbox{black!25}{gray!6}{\parbox{\dimexpr\linewidth-2\fboxsep-2\fboxrule\relax}{\scriptsize
\textbf{Sample 24}\hfill \textcolor{black!45}{\textbf{Not selected}}\\
\textbf{Candidate \#14}\hfill \texttt{score=2.68}\\
\vspace{1pt}\\
\color{black!85} Is your major sustainable enough? Whether you’re pursuing a sustainability degree and want to further your knowledge, or are interested in supplementing your major in another area with sustainability education, plenty of independent learning resources are available. A wide range of credit and noncredit courses—including university- and or\dots}}
\end{minipage}
\vspace{3pt}
\noindent
\begin{minipage}[t]{0.49\linewidth}
\fcolorbox{black!25}{gray!6}{\parbox{\dimexpr\linewidth-2\fboxsep-2\fboxrule\relax}{\scriptsize
\textbf{Sample 25}\hfill \textcolor{black!45}{\textbf{Not selected}}\\
\textbf{Candidate \#6}\hfill \texttt{score=1.77}\\
\vspace{1pt}\\
\color{black!85} Skaters need to check their skate helmets every so often and ask yourself, "Is it time to replace this helmet?" Well, that depends. Did you crash in it? For starters, most people are aware that you must replace a helmet after any crash where your head hit. The foam part of a helmet is made for one-time use, and after crushing once it is n\dots}}
\end{minipage}\hfill
\begin{minipage}[t]{0.49\linewidth}
\fcolorbox{black!25}{gray!6}{\parbox{\dimexpr\linewidth-2\fboxsep-2\fboxrule\relax}{\scriptsize
\textbf{Sample 26}\hfill \textcolor{black!45}{\textbf{Not selected}}\\
\textbf{Candidate \#0}\hfill \texttt{score=1.76}\\
\vspace{1pt}\\
\color{black!85} As it turns out, the exercises synonymous with strong, attractive abs may not be the best way to train your core—and may be doing damage to your back. Read more If you are worried about the excess holiday pounds many of us are still carrying around. There are a few easy, natural things you can do to shed them, and none of them requires an\dots}}
\end{minipage}
\vspace{3pt}
\noindent
\begin{minipage}[t]{0.49\linewidth}
\fcolorbox{black!25}{gray!6}{\parbox{\dimexpr\linewidth-2\fboxsep-2\fboxrule\relax}{\scriptsize
\textbf{Sample 27}\hfill \textcolor{black!45}{\textbf{Not selected}}\\
\textbf{Candidate \#7}\hfill \texttt{score=1.39}\\
\vspace{1pt}\\
\color{black!85} Elizabeth Hurley played as Dalila Release: Dec 8, 1996 Mara and her husband Manoa are both upstanding and religious Israelites living under the harsh and unjust rule of the Philistines. Much to their regret, they have not been able to have children. One day, a mysterious stranger appears to Mara and promises her that she will bear a son w\dots}}
\end{minipage}\hfill
\begin{minipage}[t]{0.49\linewidth}
\fcolorbox{black!25}{gray!6}{\parbox{\dimexpr\linewidth-2\fboxsep-2\fboxrule\relax}{\scriptsize
\textbf{Sample 28}\hfill \textcolor{black!45}{\textbf{Not selected}}\\
\textbf{Candidate \#5}\hfill \texttt{score=0.957}\\
\vspace{1pt}\\
\color{black!85} Well this is the big one. So big apparently, that I had to take it there and raise the number from 10 to 15. There’s just that many fails in the world of female rap. Some slight missteps, some EPIC. Nevertheless, they are all worth mentioning. You can probably think of a bunch more, but this is what I have gathered picking up from my prev\dots}}
\end{minipage}
\vspace{3pt}
\noindent
\begin{minipage}[t]{0.49\linewidth}
\fcolorbox{black!25}{gray!6}{\parbox{\dimexpr\linewidth-2\fboxsep-2\fboxrule\relax}{\scriptsize
\textbf{Sample 29}\hfill \textcolor{black!45}{\textbf{Not selected}}\\
\textbf{Candidate \#3}\hfill \texttt{score=0.919}\\
\vspace{1pt}\\
\color{black!85} starring John Travolta and Sam Jackson The first thing to understand about Basic --the basic thing, let's say-- is that although the commercials make it look like a war movie, it is not, for which we can all be grateful. No, Basic is a plot-twisty whodunnit. If The Usual Suspects died, and its body turned to cheese, and then that cheese-b\dots}}
\end{minipage}\hfill
\begin{minipage}[t]{0.49\linewidth}
\fcolorbox{black!25}{gray!6}{\parbox{\dimexpr\linewidth-2\fboxsep-2\fboxrule\relax}{\scriptsize
\textbf{Sample 30}\hfill \textcolor{black!45}{\textbf{Not selected}}\\
\textbf{Candidate \#2}\hfill \texttt{score=0.880}\\
\vspace{1pt}\\
\color{black!85} With the advent of new technologies for sneakers such as Vac Tech, Hyperfuse and Flyknit, the mid 90s and early 2000s methods of production and designing are becoming obsolete in this sneaker world. Nike Running is the future for Nike, generating billions of dollars per year, and we see Nike also not afraid to experiment with technology s\dots}}
\end{minipage}
\vspace{3pt}
\noindent
\begin{minipage}[t]{0.49\linewidth}
\fcolorbox{black!25}{gray!6}{\parbox{\dimexpr\linewidth-2\fboxsep-2\fboxrule\relax}{\scriptsize
\textbf{Sample 31}\hfill \textcolor{black!45}{\textbf{Not selected}}\\
\textbf{Candidate \#1}\hfill \texttt{score=0.798}\\
\vspace{1pt}\\
\color{black!85} Wedding \& Party Venues - Sort By: Edgartown : (508) 627-9510 A 19th century gothic revival home transformed into the island's premier eco-boutique hotel. Guests either stay in the 17-room Hob Knob hotel or in the privacy of their own Hob Knob House. Guests can expect individualized Hob Knob hospitality and modern luxury amenities in a rel\dots}}
\end{minipage}\hfill
\begin{minipage}[t]{0.49\linewidth}
\fcolorbox{black!25}{gray!6}{\parbox{\dimexpr\linewidth-2\fboxsep-2\fboxrule\relax}{\scriptsize
\textbf{Sample 32}\hfill \textcolor{black!45}{\textbf{Not selected}}\\
\textbf{Candidate \#4}\hfill \texttt{score=0.163}\\
\vspace{1pt}\\
\color{black!85} 5 Types of Women’s Underwear That Men Love Underwear can say a lot about a woman. It’s something that men are obsessed with, to the point that, a mere glimpse of a thong waistband causes us to go into shock. On the surface we find them sexy, revealing. We’re able to see who a woman actually is—or maybe some guys are just plain horny. Howe\dots}}
\end{minipage}
\vspace{3pt}

\clearpage
\noindent\phantomsection
\label{app:qual_ultrafineweb}
\noindent \textbf{Ultra-FineWeb} \\[6pt]
\noindent
\begin{minipage}[t]{0.49\linewidth}
\fcolorbox{black!25}{gray!6}{\parbox{\dimexpr\linewidth-2\fboxsep-2\fboxrule\relax}{\scriptsize
\textbf{Sample 1}\hfill \textcolor{green!45!black}{\textbf{Selected}}\\
\textbf{Candidate \#26}\hfill \texttt{score=1.000}\\
\vspace{1pt}\\
\color{black!85} Unveiling the Power: Key Provisions of the Civil Rights Act of 1864 What were the Civil Rights Act of 1864's key provisions? The Civil Rights Act of 1864 was a pivotal moment in American history, establishing crucial legal protections for African Americans in the face of rampant discrimination. Editor Note: The Civil Rights Act of 1864 la\dots}}
\end{minipage}\hfill
\begin{minipage}[t]{0.49\linewidth}
\fcolorbox{black!25}{gray!6}{\parbox{\dimexpr\linewidth-2\fboxsep-2\fboxrule\relax}{\scriptsize
\textbf{Sample 2}\hfill \textcolor{green!45!black}{\textbf{Selected}}\\
\textbf{Candidate \#29}\hfill \texttt{score=0.999}\\
\vspace{1pt}\\
\color{black!85} Earthquakes are the result of sudden movement along faults within the Earth. The movement releases stored-up ‘elastic strain’ energy in the form of seismic waves, which propagate through the Earth and cause the ground surface to shake. Such movement on the faults is generally a response to long-term deformation and the buildup of stress.\dots}}
\end{minipage}
\vspace{3pt}
\noindent
\begin{minipage}[t]{0.49\linewidth}
\fcolorbox{black!25}{gray!6}{\parbox{\dimexpr\linewidth-2\fboxsep-2\fboxrule\relax}{\scriptsize
\textbf{Sample 3}\hfill \textcolor{green!45!black}{\textbf{Selected}}\\
\textbf{Candidate \#20}\hfill \texttt{score=0.998}\\
\vspace{1pt}\\
\color{black!85} Conduct Disorder (CD) is a complex and serious behavioural and emotional disorder that can occur in children and adolescents. It’s characterised by a repetitive and persistent pattern of behaviour where the basic rights of others or major age-appropriate societal norms or rules are violated. Here’s an outline of Conduct Disorder in line w\dots}}
\end{minipage}\hfill
\begin{minipage}[t]{0.49\linewidth}
\fcolorbox{black!25}{gray!6}{\parbox{\dimexpr\linewidth-2\fboxsep-2\fboxrule\relax}{\scriptsize
\textbf{Sample 4}\hfill \textcolor{green!45!black}{\textbf{Selected}}\\
\textbf{Candidate \#22}\hfill \texttt{score=0.997}\\
\vspace{1pt}\\
\color{black!85} How To Choose Decodable Readers for First Grade To decode or not to decode: really, there is no question. To help rising first graders become successful and enthusiastic readers this summer, decodable readers are essential reading resources. Although “decodable text” might sound like yet another form of educational lingo, parents and educ\dots}}
\end{minipage}
\vspace{3pt}
\noindent
\begin{minipage}[t]{0.49\linewidth}
\fcolorbox{black!25}{gray!6}{\parbox{\dimexpr\linewidth-2\fboxsep-2\fboxrule\relax}{\scriptsize
\textbf{Sample 5}\hfill \textcolor{green!45!black}{\textbf{Selected}}\\
\textbf{Candidate \#31}\hfill \texttt{score=0.994}\\
\vspace{1pt}\\
\color{black!85} One of the challenges of working with ancient DNA samples is that damage accumulates over time, breaking the double helix structure into ever-smaller fragments. In the samples we worked with, these fragments were scattered and mixed with contaminants, making genome reconstruction a major technical challenge. But a shocking paper published\dots}}
\end{minipage}\hfill
\begin{minipage}[t]{0.49\linewidth}
\fcolorbox{black!25}{gray!6}{\parbox{\dimexpr\linewidth-2\fboxsep-2\fboxrule\relax}{\scriptsize
\textbf{Sample 6}\hfill \textcolor{green!45!black}{\textbf{Selected}}\\
\textbf{Candidate \#19}\hfill \texttt{score=0.987}\\
\vspace{1pt}\\
\color{black!85} Dividing Fractions Using Models Worksheet. This worksheet has six division with fractions issues to be solved — three must be solved with fashions and three with algorithms — options are on the second page. Answer key divide the unit fractions by whole numbers using th e fashions given. Use these resources to help reinforce the following\dots}}
\end{minipage}
\vspace{3pt}
\noindent
\begin{minipage}[t]{0.49\linewidth}
\fcolorbox{black!25}{gray!6}{\parbox{\dimexpr\linewidth-2\fboxsep-2\fboxrule\relax}{\scriptsize
\textbf{Sample 7}\hfill \textcolor{green!45!black}{\textbf{Selected}}\\
\textbf{Candidate \#30}\hfill \texttt{score=0.978}\\
\vspace{1pt}\\
\color{black!85} Over 1.8 million professionals use CFI to learn accounting, financial analysis, modeling and more. Start with a free account to explore 20+ always-free courses and hundreds of finance templates and cheat sheets. What is the Central Limit Theorem (CLT)? The Central Limit Theorem (CLT) is a statistical concept that states that the sample me\dots}}
\end{minipage}\hfill
\begin{minipage}[t]{0.49\linewidth}
\fcolorbox{black!25}{gray!6}{\parbox{\dimexpr\linewidth-2\fboxsep-2\fboxrule\relax}{\scriptsize
\textbf{Sample 8}\hfill \textcolor{green!45!black}{\textbf{Selected}}\\
\textbf{Candidate \#24}\hfill \texttt{score=0.971}\\
\vspace{1pt}\\
\color{black!85} Next we will talk about solar radiation, that is, the forms of solar radiation that we receive on earth. Solar radiation is generated by a series of nuclear fusion reactions that occur in the Sun and, as a consequence, emit electromagnetic radiation that reaches the earth. This radiation received by the earth’s surface is measured in W /\dots}}
\end{minipage}
\vspace{3pt}
\noindent
\begin{minipage}[t]{0.49\linewidth}
\fcolorbox{black!25}{gray!6}{\parbox{\dimexpr\linewidth-2\fboxsep-2\fboxrule\relax}{\scriptsize
\textbf{Sample 9}\hfill \textcolor{green!45!black}{\textbf{Selected}}\\
\textbf{Candidate \#28}\hfill \texttt{score=0.964}\\
\vspace{1pt}\\
\color{black!85} You really have to be alert when studying science. Galaxies were created after matter. The stars in those galaxies were supposed to move slowly because there was more mass in the center of the galaxy. However, after dark matter was added, the stars appeared to move faster; however, this is not the case in our galaxy, suggesting that there\dots}}
\end{minipage}\hfill
\begin{minipage}[t]{0.49\linewidth}
\fcolorbox{black!25}{gray!6}{\parbox{\dimexpr\linewidth-2\fboxsep-2\fboxrule\relax}{\scriptsize
\textbf{Sample 10}\hfill \textcolor{green!45!black}{\textbf{Selected}}\\
\textbf{Candidate \#25}\hfill \texttt{score=0.958}\\
\vspace{1pt}\\
\color{black!85} KS2 Maths is an important core subject in the National Curriculum and this area of the website covers all the major aspects of the curriculum including numbers, calculations, problems and measures. Each subject area is designed to help children develop their knowledge, whether they are learning in a classroom or home schooling environment\dots}}
\end{minipage}
\vspace{3pt}
\noindent
\begin{minipage}[t]{0.49\linewidth}
\fcolorbox{black!25}{gray!6}{\parbox{\dimexpr\linewidth-2\fboxsep-2\fboxrule\relax}{\scriptsize
\textbf{Sample 11}\hfill \textcolor{green!45!black}{\textbf{Selected}}\\
\textbf{Candidate \#8}\hfill \texttt{score=0.955}\\
\vspace{1pt}\\
\color{black!85} The Unsung Heroes of Your HVAC System: Understanding the Importance of Filters When it comes to your HVAC (Heating, Ventilation, and Air Conditioning) system, you might be quick to think about the thermostat, air ducts, or even the unit itself. However, there’s an unsung hero in your HVAC system that plays a pivotal role in maintaining in\dots}}
\end{minipage}\hfill
\begin{minipage}[t]{0.49\linewidth}
\fcolorbox{black!25}{gray!6}{\parbox{\dimexpr\linewidth-2\fboxsep-2\fboxrule\relax}{\scriptsize
\textbf{Sample 12}\hfill \textcolor{green!45!black}{\textbf{Selected}}\\
\textbf{Candidate \#27}\hfill \texttt{score=0.928}\\
\vspace{1pt}\\
\color{black!85} 24/7 writing help on your phone Save to my list Remove from my list In the tumultuous 19th century, both Italy and Germany found themselves fragmented into numerous separate ruling states. The impetus for change came in the form of rising nationalism and liberalism, paving the way for the unification of these disparate entities. However,\dots}}
\end{minipage}
\vspace{3pt}
\noindent
\begin{minipage}[t]{0.49\linewidth}
\fcolorbox{black!25}{gray!6}{\parbox{\dimexpr\linewidth-2\fboxsep-2\fboxrule\relax}{\scriptsize
\textbf{Sample 13}\hfill \textcolor{green!45!black}{\textbf{Selected}}\\
\textbf{Candidate \#15}\hfill \texttt{score=0.928}\\
\vspace{1pt}\\
\color{black!85} Origami is an art form that combines precision, creativity, and patience. While basic origami is obtainable to every one, mastering complex origami designs can be quite a rewarding and impressive achievement. In this article, we’ll show you with the procedure for creating intricate origami while highlighting essential techniques for achie\dots}}
\end{minipage}\hfill
\begin{minipage}[t]{0.49\linewidth}
\fcolorbox{black!25}{gray!6}{\parbox{\dimexpr\linewidth-2\fboxsep-2\fboxrule\relax}{\scriptsize
\textbf{Sample 14}\hfill \textcolor{green!45!black}{\textbf{Selected}}\\
\textbf{Candidate \#23}\hfill \texttt{score=0.927}\\
\vspace{1pt}\\
\color{black!85} The St. James kindergarteners have been working up to Project Week over the past month. We started slowly by taking walks in our neighborhood while Ms. Meghan and I noted what caught the children’s interest. It became apparent that the class was very interested in the L trains that they saw on our walks. It started with a simple question,\dots}}
\end{minipage}
\vspace{3pt}
\noindent
\begin{minipage}[t]{0.49\linewidth}
\fcolorbox{black!25}{gray!6}{\parbox{\dimexpr\linewidth-2\fboxsep-2\fboxrule\relax}{\scriptsize
\textbf{Sample 15}\hfill \textcolor{green!45!black}{\textbf{Selected}}\\
\textbf{Candidate \#21}\hfill \texttt{score=0.745}\\
\vspace{1pt}\\
\color{black!85} Nestled in the leafy suburbs of western Berlin, the Wannsee Conference House stands as a poignant reminder of a dark chapter in human history. The Wannsee Conference: A Pivotal Moment The Wannsee Conference, held on January 20, 1942, marked a pivotal moment in the implementation of Nazi Germany's genocidal plans. Organized by SS-Obergrupp\dots}}
\end{minipage}\hfill
\begin{minipage}[t]{0.49\linewidth}
\fcolorbox{black!25}{gray!6}{\parbox{\dimexpr\linewidth-2\fboxsep-2\fboxrule\relax}{\scriptsize
\textbf{Sample 16}\hfill \textcolor{green!45!black}{\textbf{Selected}}\\
\textbf{Candidate \#12}\hfill \texttt{score=0.718}\\
\vspace{1pt}\\
\color{black!85} Can you please give us a little short bio? (education, professional experiences, select publications, academic specialty, awards won) Public school teacher for 5 years BA art (UC Irvine) PhD. (UCLA) educational psychology Professor of Child Development, (25 years) CSUS Senior Research Scientist (Oregon Research Institute with Institute of\dots}}
\end{minipage}
\vspace{3pt}
\clearpage
\noindent\phantomsection
\label{app:qual_ultrafineweb_cont}
% \noindent \textbf{Ultra-FineWeb classifier} \\[6pt]
\noindent
\begin{minipage}[t]{0.49\linewidth}
\fcolorbox{black!25}{gray!6}{\parbox{\dimexpr\linewidth-2\fboxsep-2\fboxrule\relax}{\scriptsize
\textbf{Sample 17}\hfill \textcolor{black!45}{\textbf{Not selected}}\\
\textbf{Candidate \#14}\hfill \texttt{score=0.695}\\
\vspace{1pt}\\
\color{black!85} Is your major sustainable enough? Whether you’re pursuing a sustainability degree and want to further your knowledge, or are interested in supplementing your major in another area with sustainability education, plenty of independent learning resources are available. A wide range of credit and noncredit courses—including university- and or\dots}}
\end{minipage}\hfill
\begin{minipage}[t]{0.49\linewidth}
\fcolorbox{black!25}{gray!6}{\parbox{\dimexpr\linewidth-2\fboxsep-2\fboxrule\relax}{\scriptsize
\textbf{Sample 18}\hfill \textcolor{black!45}{\textbf{Not selected}}\\
\textbf{Candidate \#18}\hfill \texttt{score=0.648}\\
\vspace{1pt}\\
\color{black!85} This article originally appeared in the December 2015 issue of Resource Recycling. Subscribe today for access to all print content. Since the 1990s, curbside and drop-off recycling has grown substantially – nearly 90 percent of households now have access, according to recent surveys from Moore Recycling Associates, the American Forest and\dots}}
\end{minipage}
\vspace{3pt}
\noindent
\begin{minipage}[t]{0.49\linewidth}
\fcolorbox{black!25}{gray!6}{\parbox{\dimexpr\linewidth-2\fboxsep-2\fboxrule\relax}{\scriptsize
\textbf{Sample 19}\hfill \textcolor{black!45}{\textbf{Not selected}}\\
\textbf{Candidate \#11}\hfill \texttt{score=0.547}\\
\vspace{1pt}\\
\color{black!85} In decades past, classroom design was often an afterthought and followed a standardised layout. Plain boxed shaped classrooms, with identical chairs and tables throughout were commonplace in many schools. Read the latest issue of School News HERE Recently, though, there has been a shift away from this one-size-fits all approach to classro\dots}}
\end{minipage}\hfill
\begin{minipage}[t]{0.49\linewidth}
\fcolorbox{black!25}{gray!6}{\parbox{\dimexpr\linewidth-2\fboxsep-2\fboxrule\relax}{\scriptsize
\textbf{Sample 20}\hfill \textcolor{black!45}{\textbf{Not selected}}\\
\textbf{Candidate \#13}\hfill \texttt{score=0.532}\\
\vspace{1pt}\\
\color{black!85} In Heart of Darkness it is the white invaders for instance, who are, almost without exception, embodiments of blindness, selfishness, and cruelty; and even in the cognitive domain, where such positive phrases as “to enlighten,” for instance, are conventionally opposed to negative ones such as “to be in the dark,” the traditional expectati\dots}}
\end{minipage}
\vspace{3pt}
\noindent
\begin{minipage}[t]{0.49\linewidth}
\fcolorbox{black!25}{gray!6}{\parbox{\dimexpr\linewidth-2\fboxsep-2\fboxrule\relax}{\scriptsize
\textbf{Sample 21}\hfill \textcolor{black!45}{\textbf{Not selected}}\\
\textbf{Candidate \#10}\hfill \texttt{score=0.477}\\
\vspace{1pt}\\
\color{black!85} Deforestation isn't just happening in well-known global hotspots like Indonesia and Brazil's rainforest. A new analysis says forests are also shrinking on state and private land in Oregon, where an estimated 522,000 acres of forest cover have disappeared since 2000. That's an area six times larger than the city of Portland, equal to more\dots}}
\end{minipage}\hfill
\begin{minipage}[t]{0.49\linewidth}
\fcolorbox{black!25}{gray!6}{\parbox{\dimexpr\linewidth-2\fboxsep-2\fboxrule\relax}{\scriptsize
\textbf{Sample 22}\hfill \textcolor{black!45}{\textbf{Not selected}}\\
\textbf{Candidate \#17}\hfill \texttt{score=0.470}\\
\vspace{1pt}\\
\color{black!85} Political Parties and Elections Political parties are an established part of modern mass democracy, and the conduct of elections in India is largely dependent on the behaviour of political parties. Although many candidates for Indian elections are independent, the winning candidates for Lok Sabha and Vidhan Sabha elections usually stand a\dots}}
\end{minipage}
\vspace{3pt}
\noindent
\begin{minipage}[t]{0.49\linewidth}
\fcolorbox{black!25}{gray!6}{\parbox{\dimexpr\linewidth-2\fboxsep-2\fboxrule\relax}{\scriptsize
\textbf{Sample 23}\hfill \textcolor{black!45}{\textbf{Not selected}}\\
\textbf{Candidate \#9}\hfill \texttt{score=0.224}\\
\vspace{1pt}\\
\color{black!85} “Last night three cargoes of Bohea Tea were emptied into the sea. This is the most magnificent movement of all. There is a dignity, a majesty, a sublimity, in this last effort of the Patriots that I greatly admire.” - John Adams, diary entry, December 17, 1773 - John Adams, diary entry, December 17, 1773 A Novel Idea Is something so new a\dots}}
\end{minipage}\hfill
\begin{minipage}[t]{0.49\linewidth}
\fcolorbox{black!25}{gray!6}{\parbox{\dimexpr\linewidth-2\fboxsep-2\fboxrule\relax}{\scriptsize
\textbf{Sample 24}\hfill \textcolor{black!45}{\textbf{Not selected}}\\
\textbf{Candidate \#16}\hfill \texttt{score=0.211}\\
\vspace{1pt}\\
\color{black!85} What is rotavirus and why does my baby need to be immunised? Rotavirus is a very infectious virus that causes the majority of serious cases of gastroenteritis in babies. It causes diarrhoea, vomiting and abdominal pain, usually lasting around a week. Most children will be infected by rotavirus once by the age of five. Gastroenteritis (cau\dots}}
\end{minipage}
\vspace{3pt}
\noindent
\begin{minipage}[t]{0.49\linewidth}
\fcolorbox{black!25}{gray!6}{\parbox{\dimexpr\linewidth-2\fboxsep-2\fboxrule\relax}{\scriptsize
\textbf{Sample 25}\hfill \textcolor{black!45}{\textbf{Not selected}}\\
\textbf{Candidate \#7}\hfill \texttt{score=0.095}\\
\vspace{1pt}\\
\color{black!85} Elizabeth Hurley played as Dalila Release: Dec 8, 1996 Mara and her husband Manoa are both upstanding and religious Israelites living under the harsh and unjust rule of the Philistines. Much to their regret, they have not been able to have children. One day, a mysterious stranger appears to Mara and promises her that she will bear a son w\dots}}
\end{minipage}\hfill
\begin{minipage}[t]{0.49\linewidth}
\fcolorbox{black!25}{gray!6}{\parbox{\dimexpr\linewidth-2\fboxsep-2\fboxrule\relax}{\scriptsize
\textbf{Sample 26}\hfill \textcolor{black!45}{\textbf{Not selected}}\\
\textbf{Candidate \#3}\hfill \texttt{score=0.069}\\
\vspace{1pt}\\
\color{black!85} starring John Travolta and Sam Jackson The first thing to understand about Basic --the basic thing, let's say-- is that although the commercials make it look like a war movie, it is not, for which we can all be grateful. No, Basic is a plot-twisty whodunnit. If The Usual Suspects died, and its body turned to cheese, and then that cheese-b\dots}}
\end{minipage}
\vspace{3pt}
\noindent
\begin{minipage}[t]{0.49\linewidth}
\fcolorbox{black!25}{gray!6}{\parbox{\dimexpr\linewidth-2\fboxsep-2\fboxrule\relax}{\scriptsize
\textbf{Sample 27}\hfill \textcolor{black!45}{\textbf{Not selected}}\\
\textbf{Candidate \#2}\hfill \texttt{score=0.058}\\
\vspace{1pt}\\
\color{black!85} With the advent of new technologies for sneakers such as Vac Tech, Hyperfuse and Flyknit, the mid 90s and early 2000s methods of production and designing are becoming obsolete in this sneaker world. Nike Running is the future for Nike, generating billions of dollars per year, and we see Nike also not afraid to experiment with technology s\dots}}
\end{minipage}\hfill
\begin{minipage}[t]{0.49\linewidth}
\fcolorbox{black!25}{gray!6}{\parbox{\dimexpr\linewidth-2\fboxsep-2\fboxrule\relax}{\scriptsize
\textbf{Sample 28}\hfill \textcolor{black!45}{\textbf{Not selected}}\\
\textbf{Candidate \#4}\hfill \texttt{score=0.024}\\
\vspace{1pt}\\
\color{black!85} 5 Types of Women’s Underwear That Men Love Underwear can say a lot about a woman. It’s something that men are obsessed with, to the point that, a mere glimpse of a thong waistband causes us to go into shock. On the surface we find them sexy, revealing. We’re able to see who a woman actually is—or maybe some guys are just plain horny. Howe\dots}}
\end{minipage}
\vspace{3pt}
\noindent
\begin{minipage}[t]{0.49\linewidth}
\fcolorbox{black!25}{gray!6}{\parbox{\dimexpr\linewidth-2\fboxsep-2\fboxrule\relax}{\scriptsize
\textbf{Sample 29}\hfill \textcolor{black!45}{\textbf{Not selected}}\\
\textbf{Candidate \#5}\hfill \texttt{score=0.019}\\
\vspace{1pt}\\
\color{black!85} Well this is the big one. So big apparently, that I had to take it there and raise the number from 10 to 15. There’s just that many fails in the world of female rap. Some slight missteps, some EPIC. Nevertheless, they are all worth mentioning. You can probably think of a bunch more, but this is what I have gathered picking up from my prev\dots}}
\end{minipage}\hfill
\begin{minipage}[t]{0.49\linewidth}
\fcolorbox{black!25}{gray!6}{\parbox{\dimexpr\linewidth-2\fboxsep-2\fboxrule\relax}{\scriptsize
\textbf{Sample 30}\hfill \textcolor{black!45}{\textbf{Not selected}}\\
\textbf{Candidate \#0}\hfill \texttt{score=0.018}\\
\vspace{1pt}\\
\color{black!85} As it turns out, the exercises synonymous with strong, attractive abs may not be the best way to train your core—and may be doing damage to your back. Read more If you are worried about the excess holiday pounds many of us are still carrying around. There are a few easy, natural things you can do to shed them, and none of them requires an\dots}}
\end{minipage}
\vspace{3pt}
\noindent
\begin{minipage}[t]{0.49\linewidth}
\fcolorbox{black!25}{gray!6}{\parbox{\dimexpr\linewidth-2\fboxsep-2\fboxrule\relax}{\scriptsize
\textbf{Sample 31}\hfill \textcolor{black!45}{\textbf{Not selected}}\\
\textbf{Candidate \#6}\hfill \texttt{score=0.016}\\
\vspace{1pt}\\
\color{black!85} Skaters need to check their skate helmets every so often and ask yourself, "Is it time to replace this helmet?" Well, that depends. Did you crash in it? For starters, most people are aware that you must replace a helmet after any crash where your head hit. The foam part of a helmet is made for one-time use, and after crushing once it is n\dots}}
\end{minipage}\hfill
\begin{minipage}[t]{0.49\linewidth}
\fcolorbox{black!25}{gray!6}{\parbox{\dimexpr\linewidth-2\fboxsep-2\fboxrule\relax}{\scriptsize
\textbf{Sample 32}\hfill \textcolor{black!45}{\textbf{Not selected}}\\
\textbf{Candidate \#1}\hfill \texttt{score=0.000579}\\
\vspace{1pt}\\
\color{black!85} Wedding \& Party Venues - Sort By: Edgartown : (508) 627-9510 A 19th century gothic revival home transformed into the island's premier eco-boutique hotel. Guests either stay in the 17-room Hob Knob hotel or in the privacy of their own Hob Knob House. Guests can expect individualized Hob Knob hospitality and modern luxury amenities in a rel\dots}}
\end{minipage}
\vspace{3pt}

\clearpage
\noindent\phantomsection
\label{app:qual_fasttext}
\noindent \textbf{DCLM-FastText} \\[6pt]
\noindent
\begin{minipage}[t]{0.49\linewidth}
\fcolorbox{black!25}{gray!6}{\parbox{\dimexpr\linewidth-2\fboxsep-2\fboxrule\relax}{\scriptsize
\textbf{Sample 1}\hfill \textcolor{green!45!black}{\textbf{Selected}}\\
\textbf{Candidate \#28}\hfill \texttt{score=0.902}\\
\vspace{1pt}\\
\color{black!85} You really have to be alert when studying science. Galaxies were created after matter. The stars in those galaxies were supposed to move slowly because there was more mass in the center of the galaxy. However, after dark matter was added, the stars appeared to move faster; however, this is not the case in our galaxy, suggesting that there\dots}}
\end{minipage}\hfill
\begin{minipage}[t]{0.49\linewidth}
\fcolorbox{black!25}{gray!6}{\parbox{\dimexpr\linewidth-2\fboxsep-2\fboxrule\relax}{\scriptsize
\textbf{Sample 2}\hfill \textcolor{green!45!black}{\textbf{Selected}}\\
\textbf{Candidate \#29}\hfill \texttt{score=0.761}\\
\vspace{1pt}\\
\color{black!85} Earthquakes are the result of sudden movement along faults within the Earth. The movement releases stored-up ‘elastic strain’ energy in the form of seismic waves, which propagate through the Earth and cause the ground surface to shake. Such movement on the faults is generally a response to long-term deformation and the buildup of stress.\dots}}
\end{minipage}
\vspace{3pt}
\noindent
\begin{minipage}[t]{0.49\linewidth}
\fcolorbox{black!25}{gray!6}{\parbox{\dimexpr\linewidth-2\fboxsep-2\fboxrule\relax}{\scriptsize
\textbf{Sample 3}\hfill \textcolor{green!45!black}{\textbf{Selected}}\\
\textbf{Candidate \#15}\hfill \texttt{score=0.632}\\
\vspace{1pt}\\
\color{black!85} Origami is an art form that combines precision, creativity, and patience. While basic origami is obtainable to every one, mastering complex origami designs can be quite a rewarding and impressive achievement. In this article, we’ll show you with the procedure for creating intricate origami while highlighting essential techniques for achie\dots}}
\end{minipage}\hfill
\begin{minipage}[t]{0.49\linewidth}
\fcolorbox{black!25}{gray!6}{\parbox{\dimexpr\linewidth-2\fboxsep-2\fboxrule\relax}{\scriptsize
\textbf{Sample 4}\hfill \textcolor{green!45!black}{\textbf{Selected}}\\
\textbf{Candidate \#26}\hfill \texttt{score=0.612}\\
\vspace{1pt}\\
\color{black!85} Unveiling the Power: Key Provisions of the Civil Rights Act of 1864 What were the Civil Rights Act of 1864's key provisions? The Civil Rights Act of 1864 was a pivotal moment in American history, establishing crucial legal protections for African Americans in the face of rampant discrimination. Editor Note: The Civil Rights Act of 1864 la\dots}}
\end{minipage}
\vspace{3pt}
\noindent
\begin{minipage}[t]{0.49\linewidth}
\fcolorbox{black!25}{gray!6}{\parbox{\dimexpr\linewidth-2\fboxsep-2\fboxrule\relax}{\scriptsize
\textbf{Sample 5}\hfill \textcolor{green!45!black}{\textbf{Selected}}\\
\textbf{Candidate \#31}\hfill \texttt{score=0.564}\\
\vspace{1pt}\\
\color{black!85} One of the challenges of working with ancient DNA samples is that damage accumulates over time, breaking the double helix structure into ever-smaller fragments. In the samples we worked with, these fragments were scattered and mixed with contaminants, making genome reconstruction a major technical challenge. But a shocking paper published\dots}}
\end{minipage}\hfill
\begin{minipage}[t]{0.49\linewidth}
\fcolorbox{black!25}{gray!6}{\parbox{\dimexpr\linewidth-2\fboxsep-2\fboxrule\relax}{\scriptsize
\textbf{Sample 6}\hfill \textcolor{green!45!black}{\textbf{Selected}}\\
\textbf{Candidate \#27}\hfill \texttt{score=0.483}\\
\vspace{1pt}\\
\color{black!85} 24/7 writing help on your phone Save to my list Remove from my list In the tumultuous 19th century, both Italy and Germany found themselves fragmented into numerous separate ruling states. The impetus for change came in the form of rising nationalism and liberalism, paving the way for the unification of these disparate entities. However,\dots}}
\end{minipage}
\vspace{3pt}
\noindent
\begin{minipage}[t]{0.49\linewidth}
\fcolorbox{black!25}{gray!6}{\parbox{\dimexpr\linewidth-2\fboxsep-2\fboxrule\relax}{\scriptsize
\textbf{Sample 7}\hfill \textcolor{green!45!black}{\textbf{Selected}}\\
\textbf{Candidate \#3}\hfill \texttt{score=0.367}\\
\vspace{1pt}\\
\color{black!85} starring John Travolta and Sam Jackson The first thing to understand about Basic --the basic thing, let's say-- is that although the commercials make it look like a war movie, it is not, for which we can all be grateful. No, Basic is a plot-twisty whodunnit. If The Usual Suspects died, and its body turned to cheese, and then that cheese-b\dots}}
\end{minipage}\hfill
\begin{minipage}[t]{0.49\linewidth}
\fcolorbox{black!25}{gray!6}{\parbox{\dimexpr\linewidth-2\fboxsep-2\fboxrule\relax}{\scriptsize
\textbf{Sample 8}\hfill \textcolor{green!45!black}{\textbf{Selected}}\\
\textbf{Candidate \#30}\hfill \texttt{score=0.294}\\
\vspace{1pt}\\
\color{black!85} Over 1.8 million professionals use CFI to learn accounting, financial analysis, modeling and more. Start with a free account to explore 20+ always-free courses and hundreds of finance templates and cheat sheets. What is the Central Limit Theorem (CLT)? The Central Limit Theorem (CLT) is a statistical concept that states that the sample me\dots}}
\end{minipage}
\vspace{3pt}
\noindent
\begin{minipage}[t]{0.49\linewidth}
\fcolorbox{black!25}{gray!6}{\parbox{\dimexpr\linewidth-2\fboxsep-2\fboxrule\relax}{\scriptsize
\textbf{Sample 9}\hfill \textcolor{green!45!black}{\textbf{Selected}}\\
\textbf{Candidate \#20}\hfill \texttt{score=0.242}\\
\vspace{1pt}\\
\color{black!85} Conduct Disorder (CD) is a complex and serious behavioural and emotional disorder that can occur in children and adolescents. It’s characterised by a repetitive and persistent pattern of behaviour where the basic rights of others or major age-appropriate societal norms or rules are violated. Here’s an outline of Conduct Disorder in line w\dots}}
\end{minipage}\hfill
\begin{minipage}[t]{0.49\linewidth}
\fcolorbox{black!25}{gray!6}{\parbox{\dimexpr\linewidth-2\fboxsep-2\fboxrule\relax}{\scriptsize
\textbf{Sample 10}\hfill \textcolor{green!45!black}{\textbf{Selected}}\\
\textbf{Candidate \#16}\hfill \texttt{score=0.168}\\
\vspace{1pt}\\
\color{black!85} What is rotavirus and why does my baby need to be immunised? Rotavirus is a very infectious virus that causes the majority of serious cases of gastroenteritis in babies. It causes diarrhoea, vomiting and abdominal pain, usually lasting around a week. Most children will be infected by rotavirus once by the age of five. Gastroenteritis (cau\dots}}
\end{minipage}
\vspace{3pt}
\noindent
\begin{minipage}[t]{0.49\linewidth}
\fcolorbox{black!25}{gray!6}{\parbox{\dimexpr\linewidth-2\fboxsep-2\fboxrule\relax}{\scriptsize
\textbf{Sample 11}\hfill \textcolor{green!45!black}{\textbf{Selected}}\\
\textbf{Candidate \#21}\hfill \texttt{score=0.126}\\
\vspace{1pt}\\
\color{black!85} Nestled in the leafy suburbs of western Berlin, the Wannsee Conference House stands as a poignant reminder of a dark chapter in human history. The Wannsee Conference: A Pivotal Moment The Wannsee Conference, held on January 20, 1942, marked a pivotal moment in the implementation of Nazi Germany's genocidal plans. Organized by SS-Obergrupp\dots}}
\end{minipage}\hfill
\begin{minipage}[t]{0.49\linewidth}
\fcolorbox{black!25}{gray!6}{\parbox{\dimexpr\linewidth-2\fboxsep-2\fboxrule\relax}{\scriptsize
\textbf{Sample 12}\hfill \textcolor{green!45!black}{\textbf{Selected}}\\
\textbf{Candidate \#24}\hfill \texttt{score=0.108}\\
\vspace{1pt}\\
\color{black!85} Next we will talk about solar radiation, that is, the forms of solar radiation that we receive on earth. Solar radiation is generated by a series of nuclear fusion reactions that occur in the Sun and, as a consequence, emit electromagnetic radiation that reaches the earth. This radiation received by the earth’s surface is measured in W /\dots}}
\end{minipage}
\vspace{3pt}
\noindent
\begin{minipage}[t]{0.49\linewidth}
\fcolorbox{black!25}{gray!6}{\parbox{\dimexpr\linewidth-2\fboxsep-2\fboxrule\relax}{\scriptsize
\textbf{Sample 13}\hfill \textcolor{green!45!black}{\textbf{Selected}}\\
\textbf{Candidate \#8}\hfill \texttt{score=0.107}\\
\vspace{1pt}\\
\color{black!85} The Unsung Heroes of Your HVAC System: Understanding the Importance of Filters When it comes to your HVAC (Heating, Ventilation, and Air Conditioning) system, you might be quick to think about the thermostat, air ducts, or even the unit itself. However, there’s an unsung hero in your HVAC system that plays a pivotal role in maintaining in\dots}}
\end{minipage}\hfill
\begin{minipage}[t]{0.49\linewidth}
\fcolorbox{black!25}{gray!6}{\parbox{\dimexpr\linewidth-2\fboxsep-2\fboxrule\relax}{\scriptsize
\textbf{Sample 14}\hfill \textcolor{green!45!black}{\textbf{Selected}}\\
\textbf{Candidate \#17}\hfill \texttt{score=0.080}\\
\vspace{1pt}\\
\color{black!85} Political Parties and Elections Political parties are an established part of modern mass democracy, and the conduct of elections in India is largely dependent on the behaviour of political parties. Although many candidates for Indian elections are independent, the winning candidates for Lok Sabha and Vidhan Sabha elections usually stand a\dots}}
\end{minipage}
\vspace{3pt}
\noindent
\begin{minipage}[t]{0.49\linewidth}
\fcolorbox{black!25}{gray!6}{\parbox{\dimexpr\linewidth-2\fboxsep-2\fboxrule\relax}{\scriptsize
\textbf{Sample 15}\hfill \textcolor{green!45!black}{\textbf{Selected}}\\
\textbf{Candidate \#12}\hfill \texttt{score=0.067}\\
\vspace{1pt}\\
\color{black!85} Can you please give us a little short bio? (education, professional experiences, select publications, academic specialty, awards won) Public school teacher for 5 years BA art (UC Irvine) PhD. (UCLA) educational psychology Professor of Child Development, (25 years) CSUS Senior Research Scientist (Oregon Research Institute with Institute of\dots}}
\end{minipage}\hfill
\begin{minipage}[t]{0.49\linewidth}
\fcolorbox{black!25}{gray!6}{\parbox{\dimexpr\linewidth-2\fboxsep-2\fboxrule\relax}{\scriptsize
\textbf{Sample 16}\hfill \textcolor{green!45!black}{\textbf{Selected}}\\
\textbf{Candidate \#7}\hfill \texttt{score=0.041}\\
\vspace{1pt}\\
\color{black!85} Elizabeth Hurley played as Dalila Release: Dec 8, 1996 Mara and her husband Manoa are both upstanding and religious Israelites living under the harsh and unjust rule of the Philistines. Much to their regret, they have not been able to have children. One day, a mysterious stranger appears to Mara and promises her that she will bear a son w\dots}}
\end{minipage}
\vspace{3pt}
\clearpage
\noindent\phantomsection
\label{app:qual_fasttext_cont}
% \noindent \textbf{DCLM-FastText} \\[6pt]
\noindent
\begin{minipage}[t]{0.49\linewidth}
\fcolorbox{black!25}{gray!6}{\parbox{\dimexpr\linewidth-2\fboxsep-2\fboxrule\relax}{\scriptsize
\textbf{Sample 17}\hfill \textcolor{black!45}{\textbf{Not selected}}\\
\textbf{Candidate \#9}\hfill \texttt{score=0.030}\\
\vspace{1pt}\\
\color{black!85} “Last night three cargoes of Bohea Tea were emptied into the sea. This is the most magnificent movement of all. There is a dignity, a majesty, a sublimity, in this last effort of the Patriots that I greatly admire.” - John Adams, diary entry, December 17, 1773 - John Adams, diary entry, December 17, 1773 A Novel Idea Is something so new a\dots}}
\end{minipage}\hfill
\begin{minipage}[t]{0.49\linewidth}
\fcolorbox{black!25}{gray!6}{\parbox{\dimexpr\linewidth-2\fboxsep-2\fboxrule\relax}{\scriptsize
\textbf{Sample 18}\hfill \textcolor{black!45}{\textbf{Not selected}}\\
\textbf{Candidate \#5}\hfill \texttt{score=0.027}\\
\vspace{1pt}\\
\color{black!85} Well this is the big one. So big apparently, that I had to take it there and raise the number from 10 to 15. There’s just that many fails in the world of female rap. Some slight missteps, some EPIC. Nevertheless, they are all worth mentioning. You can probably think of a bunch more, but this is what I have gathered picking up from my prev\dots}}
\end{minipage}
\vspace{3pt}
\noindent
\begin{minipage}[t]{0.49\linewidth}
\fcolorbox{black!25}{gray!6}{\parbox{\dimexpr\linewidth-2\fboxsep-2\fboxrule\relax}{\scriptsize
\textbf{Sample 19}\hfill \textcolor{black!45}{\textbf{Not selected}}\\
\textbf{Candidate \#10}\hfill \texttt{score=0.026}\\
\vspace{1pt}\\
\color{black!85} Deforestation isn't just happening in well-known global hotspots like Indonesia and Brazil's rainforest. A new analysis says forests are also shrinking on state and private land in Oregon, where an estimated 522,000 acres of forest cover have disappeared since 2000. That's an area six times larger than the city of Portland, equal to more\dots}}
\end{minipage}\hfill
\begin{minipage}[t]{0.49\linewidth}
\fcolorbox{black!25}{gray!6}{\parbox{\dimexpr\linewidth-2\fboxsep-2\fboxrule\relax}{\scriptsize
\textbf{Sample 20}\hfill \textcolor{black!45}{\textbf{Not selected}}\\
\textbf{Candidate \#4}\hfill \texttt{score=0.024}\\
\vspace{1pt}\\
\color{black!85} 5 Types of Women’s Underwear That Men Love Underwear can say a lot about a woman. It’s something that men are obsessed with, to the point that, a mere glimpse of a thong waistband causes us to go into shock. On the surface we find them sexy, revealing. We’re able to see who a woman actually is—or maybe some guys are just plain horny. Howe\dots}}
\end{minipage}
\vspace{3pt}
\noindent
\begin{minipage}[t]{0.49\linewidth}
\fcolorbox{black!25}{gray!6}{\parbox{\dimexpr\linewidth-2\fboxsep-2\fboxrule\relax}{\scriptsize
\textbf{Sample 21}\hfill \textcolor{black!45}{\textbf{Not selected}}\\
\textbf{Candidate \#6}\hfill \texttt{score=0.019}\\
\vspace{1pt}\\
\color{black!85} Skaters need to check their skate helmets every so often and ask yourself, "Is it time to replace this helmet?" Well, that depends. Did you crash in it? For starters, most people are aware that you must replace a helmet after any crash where your head hit. The foam part of a helmet is made for one-time use, and after crushing once it is n\dots}}
\end{minipage}\hfill
\begin{minipage}[t]{0.49\linewidth}
\fcolorbox{black!25}{gray!6}{\parbox{\dimexpr\linewidth-2\fboxsep-2\fboxrule\relax}{\scriptsize
\textbf{Sample 22}\hfill \textcolor{black!45}{\textbf{Not selected}}\\
\textbf{Candidate \#23}\hfill \texttt{score=0.012}\\
\vspace{1pt}\\
\color{black!85} The St. James kindergarteners have been working up to Project Week over the past month. We started slowly by taking walks in our neighborhood while Ms. Meghan and I noted what caught the children’s interest. It became apparent that the class was very interested in the L trains that they saw on our walks. It started with a simple question,\dots}}
\end{minipage}
\vspace{3pt}
\noindent
\begin{minipage}[t]{0.49\linewidth}
\fcolorbox{black!25}{gray!6}{\parbox{\dimexpr\linewidth-2\fboxsep-2\fboxrule\relax}{\scriptsize
\textbf{Sample 23}\hfill \textcolor{black!45}{\textbf{Not selected}}\\
\textbf{Candidate \#19}\hfill \texttt{score=0.012}\\
\vspace{1pt}\\
\color{black!85} Dividing Fractions Using Models Worksheet. This worksheet has six division with fractions issues to be solved — three must be solved with fashions and three with algorithms — options are on the second page. Answer key divide the unit fractions by whole numbers using th e fashions given. Use these resources to help reinforce the following\dots}}
\end{minipage}\hfill
\begin{minipage}[t]{0.49\linewidth}
\fcolorbox{black!25}{gray!6}{\parbox{\dimexpr\linewidth-2\fboxsep-2\fboxrule\relax}{\scriptsize
\textbf{Sample 24}\hfill \textcolor{black!45}{\textbf{Not selected}}\\
\textbf{Candidate \#13}\hfill \texttt{score=0.00832}\\
\vspace{1pt}\\
\color{black!85} In Heart of Darkness it is the white invaders for instance, who are, almost without exception, embodiments of blindness, selfishness, and cruelty; and even in the cognitive domain, where such positive phrases as “to enlighten,” for instance, are conventionally opposed to negative ones such as “to be in the dark,” the traditional expectati\dots}}
\end{minipage}
\vspace{3pt}
\noindent
\begin{minipage}[t]{0.49\linewidth}
\fcolorbox{black!25}{gray!6}{\parbox{\dimexpr\linewidth-2\fboxsep-2\fboxrule\relax}{\scriptsize
\textbf{Sample 25}\hfill \textcolor{black!45}{\textbf{Not selected}}\\
\textbf{Candidate \#11}\hfill \texttt{score=0.00455}\\
\vspace{1pt}\\
\color{black!85} In decades past, classroom design was often an afterthought and followed a standardised layout. Plain boxed shaped classrooms, with identical chairs and tables throughout were commonplace in many schools. Read the latest issue of School News HERE Recently, though, there has been a shift away from this one-size-fits all approach to classro\dots}}
\end{minipage}\hfill
\begin{minipage}[t]{0.49\linewidth}
\fcolorbox{black!25}{gray!6}{\parbox{\dimexpr\linewidth-2\fboxsep-2\fboxrule\relax}{\scriptsize
\textbf{Sample 26}\hfill \textcolor{black!45}{\textbf{Not selected}}\\
\textbf{Candidate \#22}\hfill \texttt{score=0.00335}\\
\vspace{1pt}\\
\color{black!85} How To Choose Decodable Readers for First Grade To decode or not to decode: really, there is no question. To help rising first graders become successful and enthusiastic readers this summer, decodable readers are essential reading resources. Although “decodable text” might sound like yet another form of educational lingo, parents and educ\dots}}
\end{minipage}
\vspace{3pt}
\noindent
\begin{minipage}[t]{0.49\linewidth}
\fcolorbox{black!25}{gray!6}{\parbox{\dimexpr\linewidth-2\fboxsep-2\fboxrule\relax}{\scriptsize
\textbf{Sample 27}\hfill \textcolor{black!45}{\textbf{Not selected}}\\
\textbf{Candidate \#2}\hfill \texttt{score=0.00324}\\
\vspace{1pt}\\
\color{black!85} With the advent of new technologies for sneakers such as Vac Tech, Hyperfuse and Flyknit, the mid 90s and early 2000s methods of production and designing are becoming obsolete in this sneaker world. Nike Running is the future for Nike, generating billions of dollars per year, and we see Nike also not afraid to experiment with technology s\dots}}
\end{minipage}\hfill
\begin{minipage}[t]{0.49\linewidth}
\fcolorbox{black!25}{gray!6}{\parbox{\dimexpr\linewidth-2\fboxsep-2\fboxrule\relax}{\scriptsize
\textbf{Sample 28}\hfill \textcolor{black!45}{\textbf{Not selected}}\\
\textbf{Candidate \#18}\hfill \texttt{score=0.00124}\\
\vspace{1pt}\\
\color{black!85} This article originally appeared in the December 2015 issue of Resource Recycling. Subscribe today for access to all print content. Since the 1990s, curbside and drop-off recycling has grown substantially – nearly 90 percent of households now have access, according to recent surveys from Moore Recycling Associates, the American Forest and\dots}}
\end{minipage}
\vspace{3pt}
\noindent
\begin{minipage}[t]{0.49\linewidth}
\fcolorbox{black!25}{gray!6}{\parbox{\dimexpr\linewidth-2\fboxsep-2\fboxrule\relax}{\scriptsize
\textbf{Sample 29}\hfill \textcolor{black!45}{\textbf{Not selected}}\\
\textbf{Candidate \#0}\hfill \texttt{score=0.00109}\\
\vspace{1pt}\\
\color{black!85} As it turns out, the exercises synonymous with strong, attractive abs may not be the best way to train your core—and may be doing damage to your back. Read more If you are worried about the excess holiday pounds many of us are still carrying around. There are a few easy, natural things you can do to shed them, and none of them requires an\dots}}
\end{minipage}\hfill
\begin{minipage}[t]{0.49\linewidth}
\fcolorbox{black!25}{gray!6}{\parbox{\dimexpr\linewidth-2\fboxsep-2\fboxrule\relax}{\scriptsize
\textbf{Sample 30}\hfill \textcolor{black!45}{\textbf{Not selected}}\\
\textbf{Candidate \#14}\hfill \texttt{score=0.000783}\\
\vspace{1pt}\\
\color{black!85} Is your major sustainable enough? Whether you’re pursuing a sustainability degree and want to further your knowledge, or are interested in supplementing your major in another area with sustainability education, plenty of independent learning resources are available. A wide range of credit and noncredit courses—including university- and or\dots}}
\end{minipage}
\vspace{3pt}
\noindent
\begin{minipage}[t]{0.49\linewidth}
\fcolorbox{black!25}{gray!6}{\parbox{\dimexpr\linewidth-2\fboxsep-2\fboxrule\relax}{\scriptsize
\textbf{Sample 31}\hfill \textcolor{black!45}{\textbf{Not selected}}\\
\textbf{Candidate \#1}\hfill \texttt{score=0.000569}\\
\vspace{1pt}\\
\color{black!85} Wedding \& Party Venues - Sort By: Edgartown : (508) 627-9510 A 19th century gothic revival home transformed into the island's premier eco-boutique hotel. Guests either stay in the 17-room Hob Knob hotel or in the privacy of their own Hob Knob House. Guests can expect individualized Hob Knob hospitality and modern luxury amenities in a rel\dots}}
\end{minipage}\hfill
\begin{minipage}[t]{0.49\linewidth}
\fcolorbox{black!25}{gray!6}{\parbox{\dimexpr\linewidth-2\fboxsep-2\fboxrule\relax}{\scriptsize
\textbf{Sample 32}\hfill \textcolor{black!45}{\textbf{Not selected}}\\
\textbf{Candidate \#25}\hfill \texttt{score=0.00016}\\
\vspace{1pt}\\
\color{black!85} KS2 Maths is an important core subject in the National Curriculum and this area of the website covers all the major aspects of the curriculum including numbers, calculations, problems and measures. Each subject area is designed to help children develop their knowledge, whether they are learning in a classroom or home schooling environment\dots}}
\end{minipage}
\vspace{3pt}

\clearpage
\noindent\phantomsection
\label{app:qual_dsir}
\noindent \textbf{DSIR} \\[6pt]
\noindent
\begin{minipage}[t]{0.49\linewidth}
\fcolorbox{black!25}{gray!6}{\parbox{\dimexpr\linewidth-2\fboxsep-2\fboxrule\relax}{\scriptsize
\textbf{Sample 1}\hfill \textcolor{green!45!black}{\textbf{Selected}}\\
\textbf{Candidate \#9}\hfill \texttt{score=11.70}\\
\vspace{1pt}\\
\color{black!85} “Last night three cargoes of Bohea Tea were emptied into the sea. This is the most magnificent movement of all. There is a dignity, a majesty, a sublimity, in this last effort of the Patriots that I greatly admire.” - John Adams, diary entry, December 17, 1773 - John Adams, diary entry, December 17, 1773 A Novel Idea Is something so new a\dots}}
\end{minipage}\hfill
\begin{minipage}[t]{0.49\linewidth}
\fcolorbox{black!25}{gray!6}{\parbox{\dimexpr\linewidth-2\fboxsep-2\fboxrule\relax}{\scriptsize
\textbf{Sample 2}\hfill \textcolor{green!45!black}{\textbf{Selected}}\\
\textbf{Candidate \#26}\hfill \texttt{score=8.06}\\
\vspace{1pt}\\
\color{black!85} Unveiling the Power: Key Provisions of the Civil Rights Act of 1864 What were the Civil Rights Act of 1864's key provisions? The Civil Rights Act of 1864 was a pivotal moment in American history, establishing crucial legal protections for African Americans in the face of rampant discrimination. Editor Note: The Civil Rights Act of 1864 la\dots}}
\end{minipage}
\vspace{3pt}
\noindent
\begin{minipage}[t]{0.49\linewidth}
\fcolorbox{black!25}{gray!6}{\parbox{\dimexpr\linewidth-2\fboxsep-2\fboxrule\relax}{\scriptsize
\textbf{Sample 3}\hfill \textcolor{green!45!black}{\textbf{Selected}}\\
\textbf{Candidate \#5}\hfill \texttt{score=4.71}\\
\vspace{1pt}\\
\color{black!85} Well this is the big one. So big apparently, that I had to take it there and raise the number from 10 to 15. There’s just that many fails in the world of female rap. Some slight missteps, some EPIC. Nevertheless, they are all worth mentioning. You can probably think of a bunch more, but this is what I have gathered picking up from my prev\dots}}
\end{minipage}\hfill
\begin{minipage}[t]{0.49\linewidth}
\fcolorbox{black!25}{gray!6}{\parbox{\dimexpr\linewidth-2\fboxsep-2\fboxrule\relax}{\scriptsize
\textbf{Sample 4}\hfill \textcolor{green!45!black}{\textbf{Selected}}\\
\textbf{Candidate \#4}\hfill \texttt{score=4.62}\\
\vspace{1pt}\\
\color{black!85} 5 Types of Women’s Underwear That Men Love Underwear can say a lot about a woman. It’s something that men are obsessed with, to the point that, a mere glimpse of a thong waistband causes us to go into shock. On the surface we find them sexy, revealing. We’re able to see who a woman actually is—or maybe some guys are just plain horny. Howe\dots}}
\end{minipage}
\vspace{3pt}
\noindent
\begin{minipage}[t]{0.49\linewidth}
\fcolorbox{black!25}{gray!6}{\parbox{\dimexpr\linewidth-2\fboxsep-2\fboxrule\relax}{\scriptsize
\textbf{Sample 5}\hfill \textcolor{green!45!black}{\textbf{Selected}}\\
\textbf{Candidate \#28}\hfill \texttt{score=3.89}\\
\vspace{1pt}\\
\color{black!85} You really have to be alert when studying science. Galaxies were created after matter. The stars in those galaxies were supposed to move slowly because there was more mass in the center of the galaxy. However, after dark matter was added, the stars appeared to move faster; however, this is not the case in our galaxy, suggesting that there\dots}}
\end{minipage}\hfill
\begin{minipage}[t]{0.49\linewidth}
\fcolorbox{black!25}{gray!6}{\parbox{\dimexpr\linewidth-2\fboxsep-2\fboxrule\relax}{\scriptsize
\textbf{Sample 6}\hfill \textcolor{green!45!black}{\textbf{Selected}}\\
\textbf{Candidate \#0}\hfill \texttt{score=3.42}\\
\vspace{1pt}\\
\color{black!85} As it turns out, the exercises synonymous with strong, attractive abs may not be the best way to train your core—and may be doing damage to your back. Read more If you are worried about the excess holiday pounds many of us are still carrying around. There are a few easy, natural things you can do to shed them, and none of them requires an\dots}}
\end{minipage}
\vspace{3pt}
\noindent
\begin{minipage}[t]{0.49\linewidth}
\fcolorbox{black!25}{gray!6}{\parbox{\dimexpr\linewidth-2\fboxsep-2\fboxrule\relax}{\scriptsize
\textbf{Sample 7}\hfill \textcolor{green!45!black}{\textbf{Selected}}\\
\textbf{Candidate \#3}\hfill \texttt{score=3.39}\\
\vspace{1pt}\\
\color{black!85} starring John Travolta and Sam Jackson The first thing to understand about Basic --the basic thing, let's say-- is that although the commercials make it look like a war movie, it is not, for which we can all be grateful. No, Basic is a plot-twisty whodunnit. If The Usual Suspects died, and its body turned to cheese, and then that cheese-b\dots}}
\end{minipage}\hfill
\begin{minipage}[t]{0.49\linewidth}
\fcolorbox{black!25}{gray!6}{\parbox{\dimexpr\linewidth-2\fboxsep-2\fboxrule\relax}{\scriptsize
\textbf{Sample 8}\hfill \textcolor{green!45!black}{\textbf{Selected}}\\
\textbf{Candidate \#12}\hfill \texttt{score=3.30}\\
\vspace{1pt}\\
\color{black!85} Can you please give us a little short bio? (education, professional experiences, select publications, academic specialty, awards won) Public school teacher for 5 years BA art (UC Irvine) PhD. (UCLA) educational psychology Professor of Child Development, (25 years) CSUS Senior Research Scientist (Oregon Research Institute with Institute of\dots}}
\end{minipage}
\vspace{3pt}
\noindent
\begin{minipage}[t]{0.49\linewidth}
\fcolorbox{black!25}{gray!6}{\parbox{\dimexpr\linewidth-2\fboxsep-2\fboxrule\relax}{\scriptsize
\textbf{Sample 9}\hfill \textcolor{green!45!black}{\textbf{Selected}}\\
\textbf{Candidate \#2}\hfill \texttt{score=2.96}\\
\vspace{1pt}\\
\color{black!85} With the advent of new technologies for sneakers such as Vac Tech, Hyperfuse and Flyknit, the mid 90s and early 2000s methods of production and designing are becoming obsolete in this sneaker world. Nike Running is the future for Nike, generating billions of dollars per year, and we see Nike also not afraid to experiment with technology s\dots}}
\end{minipage}\hfill
\begin{minipage}[t]{0.49\linewidth}
\fcolorbox{black!25}{gray!6}{\parbox{\dimexpr\linewidth-2\fboxsep-2\fboxrule\relax}{\scriptsize
\textbf{Sample 10}\hfill \textcolor{green!45!black}{\textbf{Selected}}\\
\textbf{Candidate \#18}\hfill \texttt{score=2.51}\\
\vspace{1pt}\\
\color{black!85} This article originally appeared in the December 2015 issue of Resource Recycling. Subscribe today for access to all print content. Since the 1990s, curbside and drop-off recycling has grown substantially – nearly 90 percent of households now have access, according to recent surveys from Moore Recycling Associates, the American Forest and\dots}}
\end{minipage}
\vspace{3pt}
\noindent
\begin{minipage}[t]{0.49\linewidth}
\fcolorbox{black!25}{gray!6}{\parbox{\dimexpr\linewidth-2\fboxsep-2\fboxrule\relax}{\scriptsize
\textbf{Sample 11}\hfill \textcolor{green!45!black}{\textbf{Selected}}\\
\textbf{Candidate \#15}\hfill \texttt{score=2.07}\\
\vspace{1pt}\\
\color{black!85} Origami is an art form that combines precision, creativity, and patience. While basic origami is obtainable to every one, mastering complex origami designs can be quite a rewarding and impressive achievement. In this article, we’ll show you with the procedure for creating intricate origami while highlighting essential techniques for achie\dots}}
\end{minipage}\hfill
\begin{minipage}[t]{0.49\linewidth}
\fcolorbox{black!25}{gray!6}{\parbox{\dimexpr\linewidth-2\fboxsep-2\fboxrule\relax}{\scriptsize
\textbf{Sample 12}\hfill \textcolor{green!45!black}{\textbf{Selected}}\\
\textbf{Candidate \#11}\hfill \texttt{score=2.06}\\
\vspace{1pt}\\
\color{black!85} In decades past, classroom design was often an afterthought and followed a standardised layout. Plain boxed shaped classrooms, with identical chairs and tables throughout were commonplace in many schools. Read the latest issue of School News HERE Recently, though, there has been a shift away from this one-size-fits all approach to classro\dots}}
\end{minipage}
\vspace{3pt}
\noindent
\begin{minipage}[t]{0.49\linewidth}
\fcolorbox{black!25}{gray!6}{\parbox{\dimexpr\linewidth-2\fboxsep-2\fboxrule\relax}{\scriptsize
\textbf{Sample 13}\hfill \textcolor{green!45!black}{\textbf{Selected}}\\
\textbf{Candidate \#31}\hfill \texttt{score=1.04}\\
\vspace{1pt}\\
\color{black!85} One of the challenges of working with ancient DNA samples is that damage accumulates over time, breaking the double helix structure into ever-smaller fragments. In the samples we worked with, these fragments were scattered and mixed with contaminants, making genome reconstruction a major technical challenge. But a shocking paper published\dots}}
\end{minipage}\hfill
\begin{minipage}[t]{0.49\linewidth}
\fcolorbox{black!25}{gray!6}{\parbox{\dimexpr\linewidth-2\fboxsep-2\fboxrule\relax}{\scriptsize
\textbf{Sample 14}\hfill \textcolor{green!45!black}{\textbf{Selected}}\\
\textbf{Candidate \#24}\hfill \texttt{score=0.518}\\
\vspace{1pt}\\
\color{black!85} Next we will talk about solar radiation, that is, the forms of solar radiation that we receive on earth. Solar radiation is generated by a series of nuclear fusion reactions that occur in the Sun and, as a consequence, emit electromagnetic radiation that reaches the earth. This radiation received by the earth’s surface is measured in W /\dots}}
\end{minipage}
\vspace{3pt}
\noindent
\begin{minipage}[t]{0.49\linewidth}
\fcolorbox{black!25}{gray!6}{\parbox{\dimexpr\linewidth-2\fboxsep-2\fboxrule\relax}{\scriptsize
\textbf{Sample 15}\hfill \textcolor{green!45!black}{\textbf{Selected}}\\
\textbf{Candidate \#22}\hfill \texttt{score=0.487}\\
\vspace{1pt}\\
\color{black!85} How To Choose Decodable Readers for First Grade To decode or not to decode: really, there is no question. To help rising first graders become successful and enthusiastic readers this summer, decodable readers are essential reading resources. Although “decodable text” might sound like yet another form of educational lingo, parents and educ\dots}}
\end{minipage}\hfill
\begin{minipage}[t]{0.49\linewidth}
\fcolorbox{black!25}{gray!6}{\parbox{\dimexpr\linewidth-2\fboxsep-2\fboxrule\relax}{\scriptsize
\textbf{Sample 16}\hfill \textcolor{green!45!black}{\textbf{Selected}}\\
\textbf{Candidate \#10}\hfill \texttt{score=0.378}\\
\vspace{1pt}\\
\color{black!85} Deforestation isn't just happening in well-known global hotspots like Indonesia and Brazil's rainforest. A new analysis says forests are also shrinking on state and private land in Oregon, where an estimated 522,000 acres of forest cover have disappeared since 2000. That's an area six times larger than the city of Portland, equal to more\dots}}
\end{minipage}
\vspace{3pt}
\clearpage
\noindent\phantomsection
\label{app:qual_dsir_cont}
% \noindent \textbf{DSIR} \\[6pt]
\noindent
\begin{minipage}[t]{0.49\linewidth}
\fcolorbox{black!25}{gray!6}{\parbox{\dimexpr\linewidth-2\fboxsep-2\fboxrule\relax}{\scriptsize
\textbf{Sample 17}\hfill \textcolor{black!45}{\textbf{Not selected}}\\
\textbf{Candidate \#30}\hfill \texttt{score=0.272}\\
\vspace{1pt}\\
\color{black!85} Over 1.8 million professionals use CFI to learn accounting, financial analysis, modeling and more. Start with a free account to explore 20+ always-free courses and hundreds of finance templates and cheat sheets. What is the Central Limit Theorem (CLT)? The Central Limit Theorem (CLT) is a statistical concept that states that the sample me\dots}}
\end{minipage}\hfill
\begin{minipage}[t]{0.49\linewidth}
\fcolorbox{black!25}{gray!6}{\parbox{\dimexpr\linewidth-2\fboxsep-2\fboxrule\relax}{\scriptsize
\textbf{Sample 18}\hfill \textcolor{black!45}{\textbf{Not selected}}\\
\textbf{Candidate \#29}\hfill \texttt{score=-0.299}\\
\vspace{1pt}\\
\color{black!85} Earthquakes are the result of sudden movement along faults within the Earth. The movement releases stored-up ‘elastic strain’ energy in the form of seismic waves, which propagate through the Earth and cause the ground surface to shake. Such movement on the faults is generally a response to long-term deformation and the buildup of stress.\dots}}
\end{minipage}
\vspace{3pt}
\noindent
\begin{minipage}[t]{0.49\linewidth}
\fcolorbox{black!25}{gray!6}{\parbox{\dimexpr\linewidth-2\fboxsep-2\fboxrule\relax}{\scriptsize
\textbf{Sample 19}\hfill \textcolor{black!45}{\textbf{Not selected}}\\
\textbf{Candidate \#6}\hfill \texttt{score=-0.307}\\
\vspace{1pt}\\
\color{black!85} Skaters need to check their skate helmets every so often and ask yourself, "Is it time to replace this helmet?" Well, that depends. Did you crash in it? For starters, most people are aware that you must replace a helmet after any crash where your head hit. The foam part of a helmet is made for one-time use, and after crushing once it is n\dots}}
\end{minipage}\hfill
\begin{minipage}[t]{0.49\linewidth}
\fcolorbox{black!25}{gray!6}{\parbox{\dimexpr\linewidth-2\fboxsep-2\fboxrule\relax}{\scriptsize
\textbf{Sample 20}\hfill \textcolor{black!45}{\textbf{Not selected}}\\
\textbf{Candidate \#23}\hfill \texttt{score=-0.429}\\
\vspace{1pt}\\
\color{black!85} The St. James kindergarteners have been working up to Project Week over the past month. We started slowly by taking walks in our neighborhood while Ms. Meghan and I noted what caught the children’s interest. It became apparent that the class was very interested in the L trains that they saw on our walks. It started with a simple question,\dots}}
\end{minipage}
\vspace{3pt}
\noindent
\begin{minipage}[t]{0.49\linewidth}
\fcolorbox{black!25}{gray!6}{\parbox{\dimexpr\linewidth-2\fboxsep-2\fboxrule\relax}{\scriptsize
\textbf{Sample 21}\hfill \textcolor{black!45}{\textbf{Not selected}}\\
\textbf{Candidate \#8}\hfill \texttt{score=-0.675}\\
\vspace{1pt}\\
\color{black!85} The Unsung Heroes of Your HVAC System: Understanding the Importance of Filters When it comes to your HVAC (Heating, Ventilation, and Air Conditioning) system, you might be quick to think about the thermostat, air ducts, or even the unit itself. However, there’s an unsung hero in your HVAC system that plays a pivotal role in maintaining in\dots}}
\end{minipage}\hfill
\begin{minipage}[t]{0.49\linewidth}
\fcolorbox{black!25}{gray!6}{\parbox{\dimexpr\linewidth-2\fboxsep-2\fboxrule\relax}{\scriptsize
\textbf{Sample 22}\hfill \textcolor{black!45}{\textbf{Not selected}}\\
\textbf{Candidate \#14}\hfill \texttt{score=-0.743}\\
\vspace{1pt}\\
\color{black!85} Is your major sustainable enough? Whether you’re pursuing a sustainability degree and want to further your knowledge, or are interested in supplementing your major in another area with sustainability education, plenty of independent learning resources are available. A wide range of credit and noncredit courses—including university- and or\dots}}
\end{minipage}
\vspace{3pt}
\noindent
\begin{minipage}[t]{0.49\linewidth}
\fcolorbox{black!25}{gray!6}{\parbox{\dimexpr\linewidth-2\fboxsep-2\fboxrule\relax}{\scriptsize
\textbf{Sample 23}\hfill \textcolor{black!45}{\textbf{Not selected}}\\
\textbf{Candidate \#20}\hfill \texttt{score=-0.804}\\
\vspace{1pt}\\
\color{black!85} Conduct Disorder (CD) is a complex and serious behavioural and emotional disorder that can occur in children and adolescents. It’s characterised by a repetitive and persistent pattern of behaviour where the basic rights of others or major age-appropriate societal norms or rules are violated. Here’s an outline of Conduct Disorder in line w\dots}}
\end{minipage}\hfill
\begin{minipage}[t]{0.49\linewidth}
\fcolorbox{black!25}{gray!6}{\parbox{\dimexpr\linewidth-2\fboxsep-2\fboxrule\relax}{\scriptsize
\textbf{Sample 24}\hfill \textcolor{black!45}{\textbf{Not selected}}\\
\textbf{Candidate \#1}\hfill \texttt{score=-0.923}\\
\vspace{1pt}\\
\color{black!85} Wedding \& Party Venues - Sort By: Edgartown : (508) 627-9510 A 19th century gothic revival home transformed into the island's premier eco-boutique hotel. Guests either stay in the 17-room Hob Knob hotel or in the privacy of their own Hob Knob House. Guests can expect individualized Hob Knob hospitality and modern luxury amenities in a rel\dots}}
\end{minipage}
\vspace{3pt}
\noindent
\begin{minipage}[t]{0.49\linewidth}
\fcolorbox{black!25}{gray!6}{\parbox{\dimexpr\linewidth-2\fboxsep-2\fboxrule\relax}{\scriptsize
\textbf{Sample 25}\hfill \textcolor{black!45}{\textbf{Not selected}}\\
\textbf{Candidate \#7}\hfill \texttt{score=-0.971}\\
\vspace{1pt}\\
\color{black!85} Elizabeth Hurley played as Dalila Release: Dec 8, 1996 Mara and her husband Manoa are both upstanding and religious Israelites living under the harsh and unjust rule of the Philistines. Much to their regret, they have not been able to have children. One day, a mysterious stranger appears to Mara and promises her that she will bear a son w\dots}}
\end{minipage}\hfill
\begin{minipage}[t]{0.49\linewidth}
\fcolorbox{black!25}{gray!6}{\parbox{\dimexpr\linewidth-2\fboxsep-2\fboxrule\relax}{\scriptsize
\textbf{Sample 26}\hfill \textcolor{black!45}{\textbf{Not selected}}\\
\textbf{Candidate \#13}\hfill \texttt{score=-1.18}\\
\vspace{1pt}\\
\color{black!85} In Heart of Darkness it is the white invaders for instance, who are, almost without exception, embodiments of blindness, selfishness, and cruelty; and even in the cognitive domain, where such positive phrases as “to enlighten,” for instance, are conventionally opposed to negative ones such as “to be in the dark,” the traditional expectati\dots}}
\end{minipage}
\vspace{3pt}
\noindent
\begin{minipage}[t]{0.49\linewidth}
\fcolorbox{black!25}{gray!6}{\parbox{\dimexpr\linewidth-2\fboxsep-2\fboxrule\relax}{\scriptsize
\textbf{Sample 27}\hfill \textcolor{black!45}{\textbf{Not selected}}\\
\textbf{Candidate \#25}\hfill \texttt{score=-1.26}\\
\vspace{1pt}\\
\color{black!85} KS2 Maths is an important core subject in the National Curriculum and this area of the website covers all the major aspects of the curriculum including numbers, calculations, problems and measures. Each subject area is designed to help children develop their knowledge, whether they are learning in a classroom or home schooling environment\dots}}
\end{minipage}\hfill
\begin{minipage}[t]{0.49\linewidth}
\fcolorbox{black!25}{gray!6}{\parbox{\dimexpr\linewidth-2\fboxsep-2\fboxrule\relax}{\scriptsize
\textbf{Sample 28}\hfill \textcolor{black!45}{\textbf{Not selected}}\\
\textbf{Candidate \#21}\hfill \texttt{score=-1.61}\\
\vspace{1pt}\\
\color{black!85} Nestled in the leafy suburbs of western Berlin, the Wannsee Conference House stands as a poignant reminder of a dark chapter in human history. The Wannsee Conference: A Pivotal Moment The Wannsee Conference, held on January 20, 1942, marked a pivotal moment in the implementation of Nazi Germany's genocidal plans. Organized by SS-Obergrupp\dots}}
\end{minipage}
\vspace{3pt}
\noindent
\begin{minipage}[t]{0.49\linewidth}
\fcolorbox{black!25}{gray!6}{\parbox{\dimexpr\linewidth-2\fboxsep-2\fboxrule\relax}{\scriptsize
\textbf{Sample 29}\hfill \textcolor{black!45}{\textbf{Not selected}}\\
\textbf{Candidate \#16}\hfill \texttt{score=-1.98}\\
\vspace{1pt}\\
\color{black!85} What is rotavirus and why does my baby need to be immunised? Rotavirus is a very infectious virus that causes the majority of serious cases of gastroenteritis in babies. It causes diarrhoea, vomiting and abdominal pain, usually lasting around a week. Most children will be infected by rotavirus once by the age of five. Gastroenteritis (cau\dots}}
\end{minipage}\hfill
\begin{minipage}[t]{0.49\linewidth}
\fcolorbox{black!25}{gray!6}{\parbox{\dimexpr\linewidth-2\fboxsep-2\fboxrule\relax}{\scriptsize
\textbf{Sample 30}\hfill \textcolor{black!45}{\textbf{Not selected}}\\
\textbf{Candidate \#17}\hfill \texttt{score=-2.02}\\
\vspace{1pt}\\
\color{black!85} Political Parties and Elections Political parties are an established part of modern mass democracy, and the conduct of elections in India is largely dependent on the behaviour of political parties. Although many candidates for Indian elections are independent, the winning candidates for Lok Sabha and Vidhan Sabha elections usually stand a\dots}}
\end{minipage}
\vspace{3pt}
\noindent
\begin{minipage}[t]{0.49\linewidth}
\fcolorbox{black!25}{gray!6}{\parbox{\dimexpr\linewidth-2\fboxsep-2\fboxrule\relax}{\scriptsize
\textbf{Sample 31}\hfill \textcolor{black!45}{\textbf{Not selected}}\\
\textbf{Candidate \#27}\hfill \texttt{score=-4.50}\\
\vspace{1pt}\\
\color{black!85} 24/7 writing help on your phone Save to my list Remove from my list In the tumultuous 19th century, both Italy and Germany found themselves fragmented into numerous separate ruling states. The impetus for change came in the form of rising nationalism and liberalism, paving the way for the unification of these disparate entities. However,\dots}}
\end{minipage}\hfill
\begin{minipage}[t]{0.49\linewidth}
\fcolorbox{black!25}{gray!6}{\parbox{\dimexpr\linewidth-2\fboxsep-2\fboxrule\relax}{\scriptsize
\textbf{Sample 32}\hfill \textcolor{black!45}{\textbf{Not selected}}\\
\textbf{Candidate \#19}\hfill \texttt{score=-8.76}\\
\vspace{1pt}\\
\color{black!85} Dividing Fractions Using Models Worksheet. This worksheet has six division with fractions issues to be solved — three must be solved with fashions and three with algorithms — options are on the second page. Answer key divide the unit fractions by whole numbers using th e fashions given. Use these resources to help reinforce the following\dots}}
\end{minipage}
\vspace{3pt}

\end{document}